\documentclass{article}

\usepackage{microtype}
\usepackage{graphicx}
\usepackage{subcaption}
\usepackage{booktabs} 

\usepackage{hyperref}
\usepackage{xurl}

\usepackage[preprint]{icml2026}

\usepackage{amsmath}
\usepackage{amssymb}
\usepackage{mathtools}
\usepackage{amsthm}

\usepackage{algorithm}
\usepackage{algorithmic}
\usepackage{multirow}
\usepackage{placeins}
\usepackage{amsfonts}
\usepackage{xcolor}
\usepackage[most]{tcolorbox}
\usepackage{fvextra}

\definecolor{reppoblue}{HTML}{2A78D6}
\definecolor{indgreen}{HTML}{008300}
\tcbset{solution/.style={breakable, enhanced, boxrule=0.5pt, arc=2pt, left=7pt, right=7pt,
  top=5pt, bottom=5pt, toptitle=3pt, bottomtitle=3pt, fontupper=\small, coltitle=black,
  before skip=7pt, after skip=7pt,
  before upper={\setlength{\parskip}{3.5pt}\setlength{\parindent}{0pt}}}}
\newcommand{\codelabel}[1]{{\color{black!50}\ttfamily\scriptsize #1}\nopagebreak[4]\vspace{-3pt}}
\newtcolorbox{problembox}[1]{solution, colback=black!3, colframe=black!50, colbacktitle=black!12,
  title={\small\textbf{#1}}, before skip=14pt}
\newtcolorbox{reppobox}[3]{solution, colback=white, colframe=reppoblue!75, colbacktitle=reppoblue!10,
  title={\small\textbf{#1}\hfill\footnotesize #2\\[1pt]\scriptsize\color{black!70}#3}}
\newtcolorbox{indbox}[3]{solution, colback=white, colframe=indgreen!70, colbacktitle=indgreen!9,
  title={\small\textbf{#1}\hfill\footnotesize #2\\[1pt]\scriptsize\color{black!70}#3}}
\newcommand{\thinktag}[1]{{\color{black!55}\ttfamily\footnotesize #1}}
\providecommand{\tightlist}{\setlength{\itemsep}{0pt}\setlength{\parskip}{0pt}}

\usepackage[capitalize,noabbrev]{cleveref}

\theoremstyle{plain}
\newtheorem{theorem}{Theorem}[section]
\newtheorem{proposition}[theorem]{Proposition}
\newtheorem{lemma}[theorem]{Lemma}

\theoremstyle{definition}

\theoremstyle{remark}

\usepackage[textsize=tiny]{todonotes}

\icmltitlerunning{Repulsive Population Policy Gradients for Behaviorally Diverse Reinforcement Learning}

\begin{document}

\twocolumn[
  \icmltitle{Repulsive Population Policy Gradients for Behaviorally Diverse Reinforcement Learning}

  \begin{icmlauthorlist}
    \icmlauthor{Saeed Hedayatian}{usc}
    \icmlauthor{Harsh Shah}{usc}
    \icmlauthor{Celine Mneimneh}{ind}
    \icmlauthor{Stefanos Nikolaidis}{usc}
  \end{icmlauthorlist}

  \icmlaffiliation{usc}{Thomas Lord Department of Computer Science, University of Southern California, Los Angeles, USA}
  \icmlaffiliation{ind}{Independent Researcher}

  \icmlcorrespondingauthor{Saeed Hedayatian}{saeedhed@usc.edu}

  \icmlkeywords{Machine Learning, ICML, Quality Diversity, Reinforcement Learning}

  \vskip 0.3in
]



\printAffiliationsAndNotice{}  

\begin{abstract}
The goal of Quality-Diversity Reinforcement Learning (QD-RL) is to find a population of high-performing policies that solve a decision-making problem in diverse ways.
Despite recent progress, QD-RL methods predominantly rely on evolutionary operators--including black-box optimizers, mutations, and crossovers--to facilitate exploration and induce diversity.
Scaling these methods becomes increasingly costly as policies and behavior spaces grow larger.
Building on the Soft QD view of quality-diversity optimization, we introduce \textbf{Re}pulsive \textbf{P}opulation \textbf{P}olicy \textbf{O}ptimization (\textbf{RePPO}), which adapts PPO to jointly optimize a compact population of diverse, high-quality policies using a population-dependent, dynamically shaped reward.
On standard continuous-control tasks, RePPO achieves a strong quality-diversity tradeoff compared with state-of-the-art QD-RL methods.
Ablations further characterize this tradeoff by examining how each component affects individual quality and collective diversity.
To assess RePPO's scalability, we extend our evaluation to learning from pixels and fine-tuning large language models.
In pixel-based tasks, where large policies make most traditional QD-RL methods infeasible, RePPO maintains quality on par with standard PPO while achieving substantially higher diversity.
Its successful application to language model fine-tuning further demonstrates, for the first time, that QD-RL methods can scale to this domain.
Together, these results show that directly optimizing compact populations with RePPO provides a scalable approach to QD-RL across increasingly complex domains.
\end{abstract}
\section{Introduction}
Standard RL algorithms typically seek one policy that maximizes expected return, even though many tasks admit qualitatively different strategies.
A humanoid can move forward by walking, hopping, or dragging itself~\citep{ppga}; a manipulator can reach an object from different angles~\citep{huber2023quality, huber2025qdgset}, and an autonomous vehicle can make progress while expressing different driving tendencies~\citep{shiroshita2020behaviorally, zhang2022trajgen}.
Retaining these strategies allows us to compare how different behaviors achieve high reward and makes the available alternatives explicit.
Quality-Diversity (QD) optimization formalizes this broader objective by seeking solutions that combine high task performance with diversity under a user-defined behavior descriptor~\citep{pugh2016quality}.
The descriptor specifies which aspects of behavior should distinguish the solutions, making the desired diversity an explicit part of the learning problem.

The scalability of QD-RL depends on both how diverse policies are discovered and how the resulting collection is represented.
The dominant approach, exemplified by MAP-Elites~\citep{mapelites}, partitions behavior space into cells and stores the highest-quality solution found in each cell.
This makes behavioral coverage explicit, but ties the representation of diversity to a discretization: finer grids create more cells to fill, and the number of cells grows rapidly with descriptor dimension.
These costs become substantial when each policy is a large neural network.
Recent methods have improved search efficiency by incorporating deep RL updates~\citep{qdpg, dcgme, ppga}, yet diversity remains organized around an archive and often a separate evolutionary search procedure.
We therefore ask whether quality and diversity can instead be optimized directly over a compact population using standard RL updates.

A continuous objective over a finite population offers a way to pursue this goal without organizing the search around archive cells.
We build on Soft Quality-Diversity optimization~\citep{softqd}, which measures the behavioral coverage induced by the population through smooth contributions from individual solutions.
A high-quality solution contributes strongly around its descriptor, while nearby solutions provide overlapping coverage.
Soft QD captures this redundancy through a tractable lower bound that rewards individual quality and penalizes high-quality pairs that are behaviorally close.
The population size can then be chosen directly, with diversity expressed through interactions among its members.
Applying this objective to RL, however, requires optimizing both return and behavior without differentiating through the environment.

We show that standard policy gradients can optimize this population objective through a reward shaped by the current population.
The key observation is that additive Markovian descriptors~\citep{ppga} express behavior as the expected cumulative value of per-step signals.
A weighted combination of these signals can therefore serve as a reward for moving the behavior in a chosen direction.
We establish that, at the current population, each policy's gradient of the Soft QD lower bound is exactly a policy gradient under a shaped reward combining task performance with a population-derived behavior term.
This connection yields \textbf{Re}pulsive \textbf{P}opulation \textbf{P}olicy \textbf{O}ptimization (\textbf{RePPO}), which adapts PPO~\citep{ppo} to joint population training.
RePPO uses this formulation as a starting point, adapting its task and diversity terms for stable PPO training while preserving a consistent task-improvement signal.

Our evaluation examines both the quality-diversity tradeoff under a fixed interaction budget and the scalability of direct population training.
Standard continuous-control tasks provide a comparison with established archive-based QD-RL methods.
Atari games extend this evaluation to learning directly from pixels in a standard discrete-action domain.
Lastly, zero-RL fine-tuning of LLMs on mathematical reasoning tasks further tests the approach in large language model reinforcement learning.
Component ablations complement these experiments by examining how the algorithm's design choices affect training and the balance between individual quality and collective diversity.

Our contributions are:
\begin{enumerate}
\item We derive a population-dependent shaped reward whose policy gradient matches each policy's gradient of the Soft QD lower bound for additive Markovian behavior descriptors.
This connects a continuous QD objective to standard policy optimization in RL.
\item We introduce RePPO, a practical extension of PPO that jointly trains a compact policy population without behavior-space discretization or evolutionary operators.
We develop controls for stable population training and ablate their role in balancing quality and diversity.
\item We show competitive quality-diversity performance on continuous-control benchmarks, and demonstrate scalability to pixel-based Atari policies and RL fine-tuning of LLMs.
The language model experiments provide the first demonstration of QD-RL for large language model fine-tuning.
\end{enumerate}

\section{Related Work}

\subsection{Quality-Diversity objectives and search}

QD research has developed both representations of behavioral coverage and search procedures for discovering high-quality solutions.
MAP-Elites~\citep{mapelites} represents coverage through a partition of behavior space and retains the highest-quality solution found in each region.
This separates the task objective from the behavioral dimensions along which diversity is sought~\citep{pugh2016quality}.
Alternative archive structures address limitations of regular grids; for example, centroidal Voronoi tessellations allow the number of archive regions to be chosen independently of descriptor dimension~\citep{vassiliades2017using, mouret2023fast}.

Adaptive search and gradient information have substantially improved the efficiency of this archive-based framework.
CMA-ME combines covariance-adaptive search with archive improvement~\citep{cmame}, while CMA-MAE introduces adaptive acceptance thresholds to improve exploration~\citep{cmamae}.
Scalable variants reduce the cost of covariance adaptation for neural controllers~\citep{sepcmamae}.
Differentiable QD uses gradients of both quality and behavior descriptors to guide search in the joint objective--behavior space~\citep{dqd}.
These advances show how the structure of a QD problem can inform its search operators.

Soft QD provides a continuous population objective that supports direct gradient-based optimization~\citep{softqd}.
Its SQUAD algorithm optimizes a tractable lower bound through quality-weighted interactions between population members.
We inherit this objective and develop policy-gradient estimators and a practical RL training procedure.

\subsection{Quality-Diversity in reinforcement learning}

QD-RL has advanced through both scalable evolutionary search and the incorporation of deep RL updates.
ME-ES scales archive-based search to neural policies using evolution strategies~\citep{colas2020scaling}.
PGA-ME combines genetic variation with off-policy actor gradients~\citep{pgame}.
DCG-ME extends this approach with a descriptor-conditioned critic and distills archived behaviors into a single descriptor-conditioned policy~\citep{dcgme}.
DCRL-ME further advances this line of methods by injecting specialized versions of the descriptor-conditioned policy into the offspring population~\citep{dcrlme}.
QD-PG combines quality and diversity policy gradients to improve exploration and sample efficiency~\citep{qdpg}.
These methods demonstrate different ways to use experience from policy evaluation to guide the QD search.

On-policy optimization and parallel execution provide further routes to improving QD-RL.
Work on approximating objective and descriptor gradients brought differentiable QD to RL settings~\citep{tjanaka2022approximating}.
PPGA uses PPO to estimate these gradients through Markovian measure proxies, and an evolutionary search over gradient coefficients guides archive improvement~\citep{ppga}.
RePPO builds on the use of RL updates for quality and behavior, with population interactions determined by a continuous QD objective.

QD-RL also includes approaches that represent diverse behaviors through a conditioned policy.
QDAC, for instance, jointly learns high-performing skills using successor-feature critics under behavioral constraints~\citep{qdac}.
Its skill-conditioned actor provides a shared representation of many behaviors.
RePPO instead treats a finite population of policies as the optimization variable with interactions derived from the Soft QD lower bound.

\subsection{Diverse policies and behavior representations}

Direct policy-population optimization has also been studied through diversity objectives outside the QD formulation.
DOMiNO optimizes occupancy-based diversity while maintaining per-policy near-optimality constraints, and expresses diversity gradients through intrinsic rewards~\citep{domino}.
DvD uses a determinant-based population diversity criterion~\citep{dvd}, while DIPG encourages differences between policy-induced trajectory distributions using maximum mean discrepancy~\citep{dipg}.
RePPO's distinction lies in deriving quality-weighted interactions from a QD objective over given behavior descriptors.

Skill discovery and descriptor learning address complementary aspects of behavioral diversity.
DIAYN learns distinguishable skills through an information-theoretic objective~\citep{diayn}, APS combines exploration with successor-feature learning~\citep{liu2021aps}, and METRA develops a metric-aware objective for scalable unsupervised skill discovery~\citep{park2024metra}.
Moreover, connections between skill discovery and goal-conditioned RL can provide further approaches to learning behavioral alternatives~\citep{choi2021variational}.
Lastly, unsupervised QD approaches address the representation of diversity by constructing behavior descriptors from sampled trajectories~\citep{aurora, autoqd}.
Our focus is the population optimization problem given a descriptor set.
The policy-gradient construction applies when that descriptor admits the additive representation specified below, independent of how they are obtained.
\section{Background}

\subsection{The Soft QD population objective}

We adopt the tractable population objective derived in Soft QD~\citep{softqd}.
Let $\Theta$ be a solution space, with quality function $f:\Theta\to\mathbb{R}^{+}$ and behavior descriptor $\mathbf{b}:\Theta\to\mathbb{R}^{d}$.
For a population $\boldsymbol{\theta}=\{\theta_1,\ldots,\theta_N\}$, write $f_n=f(\theta_n)$ and $\mathbf{b}_n=\mathbf{b}(\theta_n)$.
In our setting, each $\theta_n$ parameterizes a policy.

With a positive multiplicative constant omitted and the interaction scale denoted by $\ell>0$, the Soft QD lower bound gives the following maximization objective
\begin{align}
S(\boldsymbol{\theta})
&=
\sum_{n=1}^{N} f_n
-
\sum_{1\leq i<j\leq N}
\sqrt{f_i f_j}
\exp\!\left(
-\frac{\|\mathbf{b}_i-\mathbf{b}_j\|_2^2}{\ell^2}
\right).
\label{main:eq:squad-objective}
\end{align}
The first term sums individual qualities while the second penalizes behavioral overlap between pairs, with the penalty increasing when both policies have high quality.
The scale $\ell$ determines how far these interactions extend in behavior space.
Our goal is to optimize this inherited objective using RL updates.

\subsection{RL with additive behavior descriptors}
\label{main:sec:background-rl}
In RL, both quality and behavior can be represented through expected cumulative signals~\citep{silver2021reward,ppga}.
Consider an MDP~\citep{puterman2014markov}
$\mathcal{M}=(\mathcal{S},\mathcal{A},P,r,\rho_0,\gamma)$
and a stochastic policy $\pi_\theta(a\mid s)$.
A trajectory $\tau=(s_0,a_0,s_1,a_1,\ldots)$ is generated from $\rho_0$, the policy, and the transition kernel $P$.
We assume that the following expectations are finite, with sums ending at termination in episodic tasks.

The policy's quality is its expected discounted task return,
\begin{align}
f(\theta) &= \mathbb{E}_{\tau\sim\pi_\theta} \left[\sum_{t}\gamma^t r(s_t,a_t)\right].
\label{main:eq:quality-return}
\end{align}
We consider additive Markovian behavior descriptors of the form
\begin{align}
\mathbf{b}(\theta) &= \mathbb{E}_{\tau\sim\pi_\theta} \left[\sum_{t}\gamma^t\mathbf{m}(s_t,a_t)\right],
\label{main:eq:additive-descriptor}
\end{align}
where $\mathbf{m}:\mathcal{S}\times\mathcal{A}\to\mathbb{R}^{d}$ is a vector of per-step behavior signals (cumulants).
Each coordinate describes an aspect of behavior through its expected cumulative value.
The gradient computation presented next concerns such descriptors.

\subsection{Policy gradients for quality and behavior}

The same policy-gradient identity provides gradients for task return and for directional changes in behavior.
For a scalar cumulant $c(s,a)$, define
\begin{align}
J_c(\theta) &= \mathbb{E}_{\tau\sim\pi_\theta} \left[\sum_{t}\gamma^t c(s_t,a_t)\right].
\end{align}
Let $Q_c^{\pi_\theta}(s,a)$ denote its expected future discounted sum after taking action $a$ in state $s$.
Under the usual regularity assumptions, the policy-gradient theorem~\citep{reinforce} gives
\begin{align}
\nabla_\theta J_c(\theta) &= \mathbb{E}_{\tau\sim\pi_\theta} \left[ \sum_{t}\gamma^t \nabla_\theta\log\pi_\theta(a_t\mid s_t) Q_c^{\pi_\theta}(s_t,a_t) \right].
\label{main:eq:pgt}
\end{align}
Consequently, for any fixed vector of coefficients $\mathbf{v}\in\mathbb{R}^{d}$,
\begin{align}
\big(\nabla_\theta\mathbf{b}(\theta)\big)^\top\mathbf{v} &= \nabla_\theta J_{\mathbf{v}^\top\mathbf{m}}(\theta).
\label{main:eq:directional-descriptor-pg}
\end{align}
A directional descriptor gradient is thus an ordinary policy gradient under the scalar reward $\mathbf{v}^\top\mathbf{m}(s,a)$.
This identity estimates behavioral changes from sampled experience without differentiating through the environment.
Section~\ref{main:sec:method-derivation} uses it to determine reward coefficients from the current population.
\section{Repulsive Population Policy Optimization}
\label{main:sec:method}

RePPO jointly trains a population by using the policies' relative qualities and behaviors to shape their PPO updates.
Each iteration collects experience, summarizes the population, and determines how each policy should change its behavior to complement the others.
These directions become weights on behavior advantages, which are combined with the task advantage in the actor update.
We first derive this connection from the Soft QD objective, then explain how RePPO constructs the population interaction and incorporates it into PPO.

\subsection{Population gradients as shaped rewards}
\label{main:sec:method-derivation}

The Soft QD gradient balances individual task improvement against behavioral overlap with other high-quality policies.
For policy $k$, let $W_k$ denote the objective's marginal value of increasing its quality and let $\mathbf{c}_k$ denote its direction of improvement in behavior space.
With independently parameterized policies, differentiating Eq.~\eqref{main:eq:squad-objective} and applying the chain rule separates these contributions:
\begin{align}
\nabla_{\theta_k}S(\boldsymbol{\theta}) &= W_k\nabla_{\theta_k}f_k + \big(\nabla_{\theta_k}\mathbf{b}_k\big)^\top\mathbf{c}_k,
\label{main:eq:population-gradient}
\end{align}
with the coefficients defined as
\begin{align}
W_k &= 1 - \frac{1}{2}\sum_{i\neq k}\sqrt{\frac{f_i}{f_k}} \exp\!\left(-\frac{\|\mathbf{b}_k - \mathbf{b}_i\|_2^2}{\ell^2}\right),
\label{main:eq:wk}\\
\mathbf{c}_k &= \sum_{i\neq k}\frac{2\sqrt{f_kf_i}}{\ell^2} \exp\!\left(-\frac{\|\mathbf{b}_k - \mathbf{b}_i\|_2^2}{\ell^2}\right) (\mathbf{b}_k - \mathbf{b}_i).
\label{main:eq:repulsion_direction}
\end{align}
The vector $\mathbf{c}_k$ combines directions away from other policies, weighted by both policies' qualities and their behavioral overlap.
It therefore encourages behavioral separation where high-quality policies overlap.
Each pairwise contribution vanishes at exact coincidence and decays at large separation.

Additive behavior descriptors turn this population interaction into an ordinary RL reward.
\begin{proposition}
\label{main:prop:pg}
Assume the additive representation of Section~\ref{main:sec:background-rl}, independently parameterized policies with strictly positive qualities, and the usual policy-gradient regularity conditions.
At the current population, the gradient of $S$ (Equation~\ref{main:eq:squad-objective}) with respect to $\theta_k$ is the policy gradient under
\begin{align}
\widetilde{r}_k(s,a)
&=W_k r(s,a)+\mathbf{c}_k^\top\mathbf{m}(s,a),
\label{main:eq:shaped-reward}
\end{align}
with $W_k$ and $\mathbf{c}_k$ held fixed when differentiating the return under this reward.
\end{proposition}
The result follows by applying the policy-gradient theorem to both terms in~\eqref{main:eq:population-gradient}.
Appendix~\ref{app:proof} provides the proof.
Each policy thus receives its own scalar reward, whose behavior component depends on the current population.
The resulting update can be estimated from sampled experience without differentiating through the environment.
As the population evolves, recomputing the coefficients changes how quality and diversity enter each policy's update.

The exact quality coefficient can nevertheless conflict with task learning.
When $W_k < 0$, the objective encourages a policy to become \textit{less efficient} in order to change its behavior and occupy a different behavioral niche.
In RL, we found such reversal of the task-learning signal to cause extreme instabilities that leads policies towards degenerate behaviors from which they can't easily recover.
This phenomenon is more thoroughly studied in Appendix~\ref{app:structural-ablations}.
As such, RePPO fixes the task coefficient to one and retains the derived repulsion structure, with practical controls described next.

\subsection{Constructing diversity weights from the population}
\label{main:sec:method-practical}

RePPO converts population repulsion into a diversity-weight vector $\mathbf{p}_k$ for each policy.
Its components will weight the advantages of the original behavior signals: positive and negative weights add pressure to increase and decrease the corresponding features.
Constructing this vector requires estimating the population's qualities and behaviors, computing repulsion in a chosen behavior coordinates, and expressing repulsion in terms of the signals used for learning.

Episode-level summaries provide a smoothed description of the population.
For each policy, we maintain exponential moving averages $\widehat{F}_k$ of completed-episode returns $\sum_t r_t$ and $\widehat{\mathbf{d}}_k$ of episode-average descriptors $T_\tau^{-1}\sum_t\mathbf{m}_t$, where $T_\tau$ is the episode length.
The returns measure individual quality, while the descriptors locate policies in behavior space.
Smoothing reduces fluctuations in both the interaction strengths and directions as policies learn.
We obtain positive quality weights using
$f_k^+=\widehat{F}_k-\min_j\widehat{F}_j+\epsilon_f$, with $\epsilon_f>0$.

Behavioral distances are measured in transformed coordinates.
Following SQUAD~\citep{softqd}, we map descriptors normalized to $[-1,1]^d$ into unbounded coordinates using the componentwise logit $g(d)=\log((1+d)/(1-d))$.
This defines the interaction geometry and stretches differences near the descriptor bounds.
We compute local repulsion $\mathbf{c}_k^{\mathrm{local}}$ using~\eqref{main:eq:repulsion_direction}, replacing $(f_i,\mathbf{b}_i)$ with $(f_i^+,g(\widehat{\mathbf{d}}_i))$ and restricting the sum to the $K$ nearest other policies in transformed space.
In the experiments, we use $K\!=\!8$ unless otherwise stated.
The neighborhood selects which policies interact, while the bandwidth $\ell$ determines how their influence decays with distance.
This gives us a direction in the (transformed) behavior space that the policy should follow to maximize behavioral coverage (i.e., diversity).

The resulting direction must be converted into weights on the original behavior signals before it can enter PPO.
Repulsion is computed in logit coordinates, but the behavior advantages describe changes in the original features $\mathbf{m}_t$.
The logit Jacobian $J_g$ connects these coordinates through the chain rule.
After this conversion, we cap the vector's norm to control the overall diversity pressure:
\begin{align}
\mathbf{p}_k
&=\operatorname{cap}_{C_{\max}}\!\left(
J_g(\widehat{\mathbf{d}}_k)^\top\mathbf{c}_k^{\mathrm{local}}
\right),
\label{main:eq:practical-repulsion}
\end{align}
where the cap preserves direction, rescales vectors exceeding norm $C_{\max}$ to that norm, and leaves all others unchanged.

\subsection{PPO updates with population-shaped advantages}
\label{main:sec:method-ppo}

Separate task and behavior critics allow the population interaction to change while keeping each critic tied to a fixed signal.
For each policy, a task critic and a behavior critic with $d$ outputs estimate discounted generalized advantages~\citep{gae} for $r_t$ and each $m_{t,j}$.
The population affects how the channels are combined, without entering the critics' prediction targets.

RePPO combines these learning signals after putting their advantage channels on comparable scales.
Within each policy's PPO minibatch, we standardize each channel separately, denoting the resulting advantages by $\overline{A}$.
This prevents differences in the numerical scales of the channels from determining their relative influence.
The actor then uses
\begin{align}
\widetilde{A}_{k,t}
&=\overline{A}^{r}_{k,t}
+\sum_{j=1}^{d}p_{k,j}\overline{A}^{m_j}_{k,t}.
\label{main:eq:composite-advantage}
\end{align}
The task channel retains unit weight, while $\mathbf{p}_k$ determines the contribution of each behavior channel, collectively capped by $C_{\max}$.
The combined advantage is then used with the clipped PPO actor update~\citep{ppo}.
The critics are fitted to return targets for their respective signals.
Note that setting $\mathbf{p}_k=\mathbf{0}$ recovers the task-only PPO actor update.

Updating population interaction coefficients and policy learning alternate throughout training.
After collecting rollouts from all policies, RePPO updates the population summaries and computes the diversity weights.
These weights remain fixed throughout the PPO epochs on that rollout data, so each policy learns consistently during its update.
The updated policies then generate new experience, refreshing the summaries and the interaction for the next iteration.
Algorithm~\ref{main:alg:reppo} summarizes this loop.
The episode statistics, unit task coefficient, local and capped interactions, and channel normalization modify the exact gradient in Proposition~\ref{main:prop:pg} to enable stable and efficient optimization.
We examine their effects, together with population size and other hyperparameters, in Appendix~\ref{app:ablations}.

\begin{algorithm}[th!]
\caption{Repulsive Population Policy Optimization (RePPO)}
\label{main:alg:reppo}
\begin{algorithmic}[1]
\REQUIRE Population size $N$; interaction parameters $K,\ell^2,C_{\max},\epsilon_f$; EMA rates; PPO settings
\STATE Initialize policies, task critics, and behavior critics
\STATE Initialize $\widehat{F}_k\gets\epsilon_f$ and $\widehat{\mathbf{d}}_k\gets\mathbf{0}$ for all $k$
\FOR{each training iteration}
    \FOR{each policy $k$}
        \STATE Collect on-policy data $\mathcal{D}_k$, including task and behavior signals
        \STATE Update $\widehat{F}_k,\widehat{\mathbf{d}}_k$ from completed episodes
    \ENDFOR
    \STATE $f_k^+\gets\widehat{F}_k-\min_j\widehat{F}_j+\epsilon_f$ for all $k$
    \STATE Transform smoothed descriptors using the logit and boundary clipping
    \FOR{each policy $k$}
        \STATE Compute local repulsion using shifted qualities and nearest neighbors in transformed space
        \STATE Convert and cap this repulsion to obtain $\mathbf{p}_k$ using~\eqref{main:eq:practical-repulsion}
        \STATE Compute task and behavior GAE channels on $\mathcal{D}_k$
    \ENDFOR
    \FOR{each policy $k$ and each PPO epoch}
        \FOR{each minibatch from $\mathcal{D}_k$}
            \STATE Standardize each advantage channel within the minibatch
            \STATE Combine channels using $\mathbf{p}_k$ as in~\eqref{main:eq:composite-advantage}
            \label{main:alg:line-combine}
            \STATE Apply the PPO actor update and fit the critics to their return targets
        \ENDFOR
    \ENDFOR
\ENDFOR
\STATE \textbf{return} $\{\pi_{\theta_k}\}_{k=1}^N$
\end{algorithmic}
\end{algorithm}
\section{Experiments}
\label{main:sec:experiments}

Our experiments test whether directly optimizing a compact population yields a favorable balance of quality and diversity, and whether this approach scales to domains where archive-based QD-RL becomes impractical.
We compare RePPO with established QD-RL methods on continuous-control benchmarks, then turn to two settings where archives of policies become costly or impractical: pixel-based Atari policies and LLM fine-tuning for mathematical reasoning.
Appendices~\ref{app:pixels} and~\ref{app:lunarlander} add results on pixel-based locomotion and on LunarLander.

\paragraph{Benchmarks.}
The continuous-control benchmark consists of four Brax locomotion tasks~\citep{brax} commonly used in QD-RL: Ant, HalfCheetah, Humanoid, and Walker2D~\citep{ppga}.
Fitness, the quality $f$ of Section~\ref{main:sec:background-rl}, is the episode return, and the descriptor records the fraction of time each foot touches the ground, giving four dimensions for Ant and two for the other tasks.
For Atari, we use four games from the Arcade Learning Environment~\citep{bellmare2015arcade, machado2018revisiting}: Boxing, Asterix, Seaquest, and Ms.\ Pac-Man.
For each task, fitness is the game score, and two-dimensional descriptors record where the agent plays, its distant to opponent (in Boxing), and the order in which it clears the maze (in Ms.\ Pac-Man).
For LLMs, we fine-tune two base models, Qwen3-1.7B~\citep{qwen3} and Granite-4.1-3B~\citep{granite}, on MATH problems~\citep{hendrycks2021math} with a binary correctness reward, and evaluate them on MATH-500~\citep{lightman2024verify}, Minerva~\citep{lewkowycz2022minerva}, and OlympiadBench~\citep{he2024olympiadbench}.
Four descriptors characterize how a correct solution is written: the share of Python \emph{code}, the rate of explicit self-\emph{verification}, the share of \LaTeX{} \emph{symbols}, and the use of \emph{markdown} structure.
In this domain, all descriptors and the reward are sparse as only the last token may get non-zero values.
Appendices~\ref{app:cc},~\ref{app:atari}, and~\ref{app:llm} define them and list all hyperparameters.

\paragraph{Baselines.}
In every domain, we compare against Independent PPO (Ind-PPO), which uses RePPO's implementation, population size, and hyperparameters with the diversity weights $\mathbf{p}_k$ set to zero, so that the comparison isolates the effect of repulsion.
It is equivalent to training $N$ PPO policies each with a different seed.
In continuous control, we additionally compare against archive-based methods representing the main families of QD-RL: PPGA~\citep{ppga}, which uses PPO for differentiable QD; PGA-ME~\citep{pgame} and DCRL-ME~\citep{dcrlme}, which add TD3-based variation~\citep{td3} to MAP-Elites; and the black-box Sep-CMA-MAE~\citep{sepcmamae}.
In Atari, since the aforementioned methods become prohibitively inefficient and costly, we compare against RefQD~\citep{wang2024quality}, which shares one representation network across its archive.
For LLMs, we compare against the base model and two prompting baselines in which each one of the members is the base model with a different style instruction, either combining extreme values of several descriptors or emphasizing a single one.

\paragraph{Evaluation.}
After training, we evaluate every policy anew to estimate its fitness and mean descriptor, using $256$ episodes in continuous control, $32$ full games in Atari, and $8$ samples per test problem for LLMs.
Archive-based methods return thousands of policies, many of which fail to solve the task.
Because of this, we discard policies whose fitness is below $10\%$ of the best policy on that task and evaluate the remaining policies.
On Ant, we use a $25\%$ threshold, since stationary policies with varied contact patterns collect enough survival bonus to clear the lower bar.
We measure quality by the mean fitness of the retained policies and diversity by the Vendi Score of their descriptors~\citep{vendi}, which can be read as the effective number of distinct behaviors in the population.
Their product, the quality-weighted Vendi Score (qVS)~\citep{qvs}, serves as a combined measure of quality and diversity.
We also report each population's best fitness before filtering, and vary the threshold in Appendix~\ref{app:cc-filter}.

\subsection{Continuous control}
\label{main:sec:cc}

\begin{figure*}[t]
    \centering
    \includegraphics[width=\textwidth]{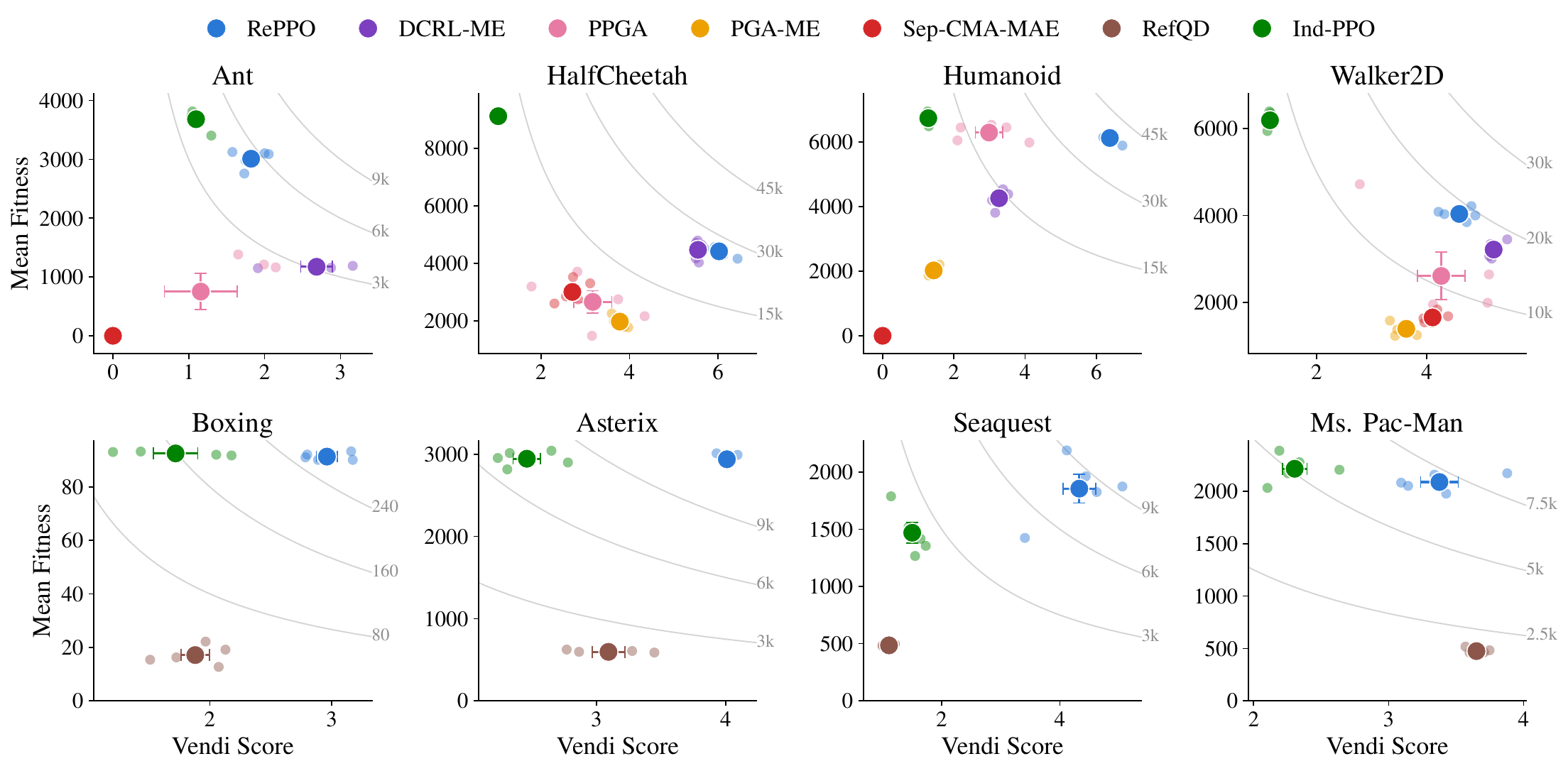}
    \caption{
    Quality and diversity of the final populations in continuous control (top) and Atari (bottom).
    Large markers show the mean over five seeds with one standard error, and faint dots show individual runs.
    Gray curves are levels of constant qVS, signifying the combined quality and diversity of a population.
    Up and to the right is better.
    }
    \label{main:fig:qd}
\end{figure*}

RePPO trains $N\!=\!32$ policies for $2.05$~B environment steps in total, the same budget as Ind-PPO, PGA-ME, DCRL-ME, and Sep-CMA-MAE, while PPGA uses $2.74$~B steps.
PPGA's budget can only be controlled indirectly through its number of iterations which is why it has a higher budget.

RePPO attains the highest qVS on all four tasks by combining the quality of PPO training with diversity that rivals archive-based search (Figure~\ref{main:fig:qd}, top).
The baselines fall on either side of this balance.
Ind-PPO trains strong policies, but without repulsion they converge to the same behavior.
Archive-based methods cover behavior space broadly, but with their budget spread over thousands of cells, most of their policies remain weaker.
RePPO's population balances the two.
On Humanoid, it is about twice as diverse as any archive at a mean fitness close to that of the fittest one.
Ant proves to be the hardest task for QD baselines, with many of them struggling to find high-performing policies and instead diversifying low-quality behaviors.
This is particularly evident when we look at the highest fitness among the populations that each method finds, where these methods trail RePPO and Ind-PPO substantially (Table~\ref{app:tab:cc-full}).
Still, RePPO is able to find high-quality policies while diversifying more than Ind-PPO.
Overall, DCRL-ME comes closest and even matches RePPO on HalfCheetah.

The results show that repulsion costs part of the population's mean fitness but leaves its best policies largely intact.
Policies pushed toward less common contact patterns adopt gaits that are often slower, which lowers the average, most visibly on HalfCheetah.
The best policy of each population, however, stays close to the best one trained by Ind-PPO and outperforms those of nearly all archive-based baselines (Table~\ref{app:tab:cc-full}).
Appendices~\ref{app:cc-results} and~\ref{app:ablations} visualize the resulting populations and examine how each component of RePPO shapes this tradeoff.

\subsection{Pixel-based policies in Atari}
\label{main:sec:atari}

\begin{figure}[t]
    \centering
    \includegraphics[width=\columnwidth]{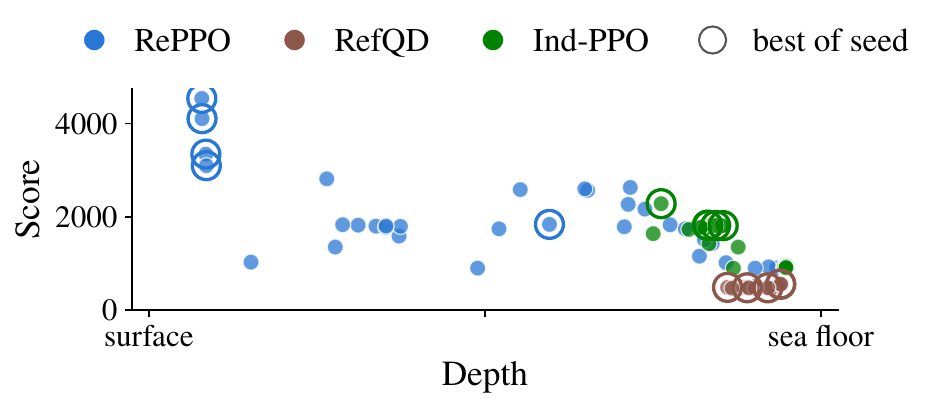}
    \caption{
    Score and mean depth of every Seaquest policy over five seeds that passes the $10\%$ quality filter, with the best policy of each seed circled.
    The policies of Ind-PPO and RefQD stay near the sea floor, whereas each RePPO population spreads across depths, and its best policies operate just below the surface.
    }
    \label{main:fig:seaquest}
\end{figure}

RePPO and Ind-PPO train $N\!=\!8$ policies for $10$~M steps each, following the standard PPO configuration for Atari~\citep{huang2022cleanrl}.
RefQD uses a comparable budget of $86.4$~M steps and maintains an archive of $1000$ cells.
Its training configuration and hyperparameters largely follow the published code.

RePPO retains the quality of independent PPO training while making its populations substantially more diverse (Figure~\ref{main:fig:qd}, bottom).
Its mean scores are close to or above Ind-PPO's in every game and its diversity is higher than both baselines, culminating in substantial gains in qVS over both baselines.
RefQD's archives cover behavior space broadly, but their policies score far below those trained with PPO.

Diversity can also uncover strategies that independent training misses.
In Seaquest, every policy trained by Ind-PPO, and every competent policy in RefQD's archives, patrols near the sea floor (Figure~\ref{main:fig:seaquest}).
Repulsion pushes some of RePPO's policies toward shallower water, and in four of the five seeds, the best policy of the population operates just below the surface, where it outscores every Ind-PPO policy.
Here, diversity improves not only the population as a whole but also its best member.
Appendix~\ref{app:atari-results} visualizes the behaviors in all four games.

\subsection{Fine-tuning large language models}
\label{main:sec:llm}

\begin{table}[t]
\centering
\caption{
LLM populations on the test problems.
Vendi is the Vendi Score of the members' mean descriptors, and ID is the accuracy of a classifier that identifies a solution's author (chance is $20\%$).
}
\label{main:tab:llm}
\small
\setlength{\tabcolsep}{5pt}
\begin{tabular}{@{}llccc@{}}
\toprule
Model & Method & pass@1 & Vendi & ID \\
\midrule
\multirow{5}{*}{Qwen3}
& Base model & $0.320$ & -- & -- \\
& Prompt (corners) & $0.274$ & $1.11$ & $51\%$ \\
& Prompt (emphasis) & $0.275$ & $1.08$ & $44\%$ \\
& Ind-PPO & $\mathbf{0.423}$ & $1.04$ & $41\%$ \\
& RePPO & $0.413$ & $\mathbf{1.85}$ & $\mathbf{82\%}$ \\
\midrule
\multirow{5}{*}{Granite}
& Base model & $0.233$ & -- & -- \\
& Prompt (corners) & $0.163$ & $1.01$ & $31\%$ \\
& Prompt (emphasis) & $0.194$ & $1.03$ & $37\%$ \\
& Ind-PPO & $\mathbf{0.560}$ & $1.03$ & $53\%$ \\
& RePPO & $0.550$ & $\mathbf{2.57}$ & $\mathbf{81\%}$ \\
\bottomrule
\end{tabular}
\end{table}

\begin{figure}[t]
    \centering
    \includegraphics[width=\columnwidth]{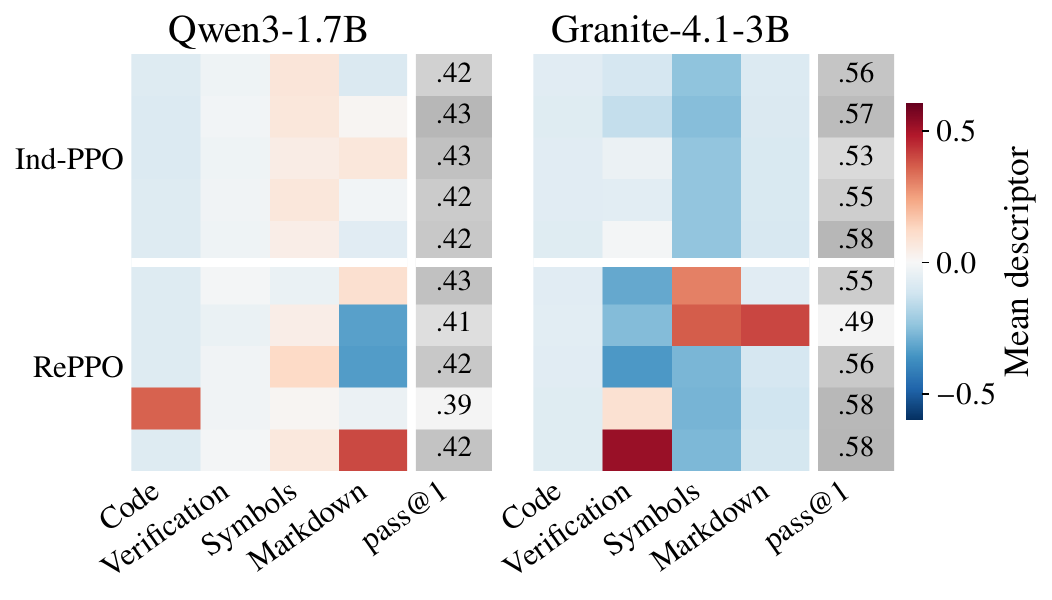}
    \caption{
    Mean descriptor of each member over its test solutions, alongside its pass@1. Descriptors are centered on the base model's style, and incorrect solutions count as zero.
    RePPO's members adopt distinct styles while matching the accuracy of Ind-PPO's near-identical members.
    }
    \label{main:fig:llm}
\end{figure}

In fine-tuning large language models, each population consists of $N\!=\!5$ members, each a low-rank adapter~\citep{hu2022lora} on a shared frozen backbone with its own adapted critic.
Members are trained with PPO in the zero-RL setting~\citep{hu2025openreasonerzero} for $100$ iterations of $1024$ solutions each.
We select checkpoints by validation accuracy, and report behavior and performance over test problems (Appendix~\ref{app:llm}).

RePPO's members acquire distinct and consistent writing styles while nearly matching the accuracy of independent training (Table~\ref{main:tab:llm}).
In both models, repulsion makes the members' descriptors far more diverse than those of Ind-PPO or of either prompting baseline.
The members are also distinguishable from their text alone: a simple classifier identifies which RePPO member wrote a solution much more reliably than for any other population.
The price is a small but consistent drop of about one point in per-member accuracy, while the population as a whole solves as many problems as Ind-PPO's (Appendix~\ref{app:llm-results}).
Prompting, in contrast, changes the members' style only slightly and at a much higher cost of accuracy.

The styles differ across members and base models (Figure~\ref{main:fig:llm}).
On Qwen3, one RePPO member appends Python code to its solutions, another writes in itemized markdown, and two write in plain prose.
On Granite, the members separate along verification and notation: some re-check their intermediate results far more often than others, and some typeset their reasoning in \LaTeX{} instead of plain text.
Appendix~\ref{app:llm-examples} provides examples of the different models in RePPO's population generating stylistically diverse responses to the same problems.

\section{Conclusion and Outlook}
\label{main:sec:conclusion}

We introduced RePPO, a Quality-Diversity algorithm that directly trains a compact population of reinforcement learning policies.
Starting from the Soft QD objective, we showed that additive Markovian behavior descriptors turn each policy's population gradient into an ordinary policy gradient under a shaped reward, which RePPO optimizes with PPO.
On continuous-control benchmarks, RePPO attains a better balance of quality and diversity than established QD-RL methods.
Because its cost scales with the population size rather than with the resolution of an archive, it extends to pixel-based Atari policies and to LLM fine-tuning, where it makes populations substantially more diverse at nearly the quality of independent training.

RePPO is designed for optimizing compact populations of useful policies rather than exhaustive coverage of behavior space, for which archive-based methods remain better suited.
It also requires descriptors that admit additive forms, and it diversifies whatever these descriptors measure: in our LLM experiments, hand-designed descriptors produced differences mostly in presentation rather than in strategy.
Addressing these limitations calls for learned or preference-derived descriptors, adaptive bandwidths and population sizes, and hybrids with external archives.

More broadly, shaping each policy's reward by the population is useful whenever one optimum is not enough, from adaptation to changing conditions to generative models whose adapters express different preferences or styles.
In all these settings, diversity can be encoded as a reward induced by the geometry of a population.

\bibliography{src/references}

@article{silver2021reward,
  title={Reward is enough},
  author={Silver, David and Singh, Satinder and Precup, Doina and Sutton, Richard S},
  journal={Artificial intelligence},
  volume={299},
  pages={103535},
  year={2021},
  publisher={Elsevier}
}

@article{reinforce,
  title={Simple statistical gradient-following algorithms for connectionist reinforcement learning},
  author={Williams, Ronald J},
  journal={Machine learning},
  volume={8},
  number={3},
  pages={229--256},
  year={1992},
  publisher={Springer}
}

@inproceedings{
softqd,
title={Soft Quality-Diversity Optimization},
author={Saeed Hedayatian and Stefanos Nikolaidis},
booktitle={The Fourteenth International Conference on Learning Representations},
year={2026},
url={https://openreview.net/forum?id=LCKixyFShR}
}

@article{mapelites,
  author       = {Antoine Cully and
                  Jeff Clune and
                  Danesh Tarapore and
                  Jean{-}Baptiste Mouret},
  title        = {Robots that can adapt like animals},
  journal      = {Nat.},
  volume       = {521},
  number       = {7553},
  pages        = {503--507},
  year         = {2015},
  url          = {https://doi.org/10.1038/nature14422},
  doi          = {10.1038/NATURE14422},
}

@inproceedings{ppga,
  title={Proximal policy gradient arborescence for quality diversity reinforcement learning},
  author={Batra, Sumeet and Tjanaka, Bryon and Fontaine, Matthew and Petrenko, Aleksei and Nikolaidis, Stefanos and Sukhatme, Gaurav},
  booktitle={International Conference on Learning Representations},
  volume={2024},
  pages={35884--35904},
  year={2024}
}

@article{huber2023quality,
  title={Quality diversity under sparse reward and sparse interaction: Application to grasping in robotics},
  author={Huber, Johann and H{\'e}l{\'e}non, Fran{\c{c}}ois and Coninx, Miranda and Amar, F Ben and Doncieux, St{\'e}phane},
  journal={arXiv preprint arXiv:2308.05483},
  year={2023}
}

@inproceedings{huber2025qdgset,
  title={Qdgset: A large scale grasping dataset generated with quality-diversity},
  author={Huber, Johann and H{\'e}l{\'e}non, Fran{\c{c}}ois and Kappel, Mathilde and de Loyola P{\'a}ez-Ubieta, Ignacio and Puente, Santiago T and Gil, Pablo and Amar, Fa{\"\i}z Ben and Doncieux, St{\'e}phane},
  booktitle={2025 IEEE International Conference on Robotics and Automation (ICRA)},
  pages={01--08},
  year={2025},
  organization={IEEE}
}

@inproceedings{shiroshita2020behaviorally,
  title={Behaviorally diverse traffic simulation via reinforcement learning},
  author={Shiroshita, Shinya and Maruyama, Shirou and Nishiyama, Daisuke and Castro, Mario Ynocente and Hamzaoui, Karim and Rosman, Guy and DeCastro, Jonathan and Lee, Kuan-Hui and Gaidon, Adrien},
  booktitle={2020 IEEE/RSJ International Conference on Intelligent Robots and Systems (IROS)},
  pages={2103--2110},
  year={2020},
  organization={IEEE}
}

@article{zhang2022trajgen,
  title={TrajGen: Generating realistic and diverse trajectories with reactive and feasible agent behaviors for autonomous driving},
  author={Zhang, Qichao and Gao, Yinfeng and Zhang, Yikang and Guo, Youtian and Ding, Dawei and Wang, Yunpeng and Sun, Peng and Zhao, Dongbin},
  journal={IEEE Transactions on Intelligent Transportation Systems},
  volume={23},
  number={12},
  pages={24474--24487},
  year={2022},
  publisher={IEEE}
}

@article{pugh2016quality,
  title={Quality diversity: A new frontier for evolutionary computation},
  author={Pugh, Justin K and Soros, Lisa B and Stanley, Kenneth O},
  journal={Frontiers in Robotics and AI},
  volume={3},
  pages={40},
  year={2016},
  publisher={Frontiers Media SA}
}

@inproceedings{mouret2023fast,
  title={Fast generation of centroids for MAP-Elites},
  author={Mouret, Jean-Baptiste},
  booktitle={Proceedings of the Companion Conference on Genetic and Evolutionary Computation},
  pages={155--158},
  year={2023}
}

@article{vassiliades2017using,
  title={Using centroidal voronoi tessellations to scale up the multidimensional archive of phenotypic elites algorithm},
  author={Vassiliades, Vassilis and Chatzilygeroudis, Konstantinos and Mouret, Jean-Baptiste},
  journal={IEEE Transactions on Evolutionary Computation},
  volume={22},
  number={4},
  pages={623--630},
  year={2017},
  publisher={IEEE}
}

@inproceedings{pgame,
  title={Policy gradient assisted map-elites},
  author={Nilsson, Olle and Cully, Antoine},
  booktitle={Proceedings of the Genetic and Evolutionary Computation Conference},
  pages={866--875},
  year={2021}
}

@inproceedings{colas2020scaling,
  title={Scaling map-elites to deep neuroevolution},
  author={Colas, C{\'e}dric and Madhavan, Vashisht and Huizinga, Joost and Clune, Jeff},
  booktitle={Proceedings of the 2020 Genetic and Evolutionary Computation Conference},
  pages={67--75},
  year={2020}
}

@inproceedings{dcgme,
  title={Map-elites with descriptor-conditioned gradients and archive distillation into a single policy},
  author={Faldor, Maxence and Chalumeau, F{\'e}lix and Flageat, Manon and Cully, Antoine},
  booktitle={Proceedings of the Genetic and Evolutionary Computation Conference},
  pages={138--146},
  year={2023}
}

@inproceedings{qdpg,
  title={Diversity policy gradient for sample efficient quality-diversity optimization},
  author={Pierrot, Thomas and Mac{\'e}, Valentin and Chalumeau, Felix and Flajolet, Arthur and Cideron, Geoffrey and Beguir, Karim and Cully, Antoine and Sigaud, Olivier and Perrin-Gilbert, Nicolas},
  booktitle={Proceedings of the Genetic and Evolutionary Computation Conference},
  pages={1075--1083},
  year={2022}
}

@inproceedings{tjanaka2022approximating,
  title={Approximating gradients for differentiable quality diversity in reinforcement learning},
  author={Tjanaka, Bryon and Fontaine, Matthew C and Togelius, Julian and Nikolaidis, Stefanos},
  booktitle={Proceedings of the Genetic and Evolutionary Computation Conference},
  pages={1102--1111},
  year={2022}
}

@article{dqd,
  title={Differentiable quality diversity},
  author={Fontaine, Matthew and Nikolaidis, Stefanos},
  journal={Advances in Neural Information Processing Systems},
  volume={34},
  pages={10040--10052},
  year={2021}
}

@article{ppo,
  title={Proximal policy optimization algorithms},
  author={Schulman, John and Wolski, Filip and Dhariwal, Prafulla and Radford, Alec and Klimov, Oleg},
  journal={arXiv preprint arXiv:1707.06347},
  year={2017}
}

@inproceedings{cmamae,
  title={Covariance matrix adaptation map-annealing},
  author={Fontaine, Matthew and Nikolaidis, Stefanos},
  booktitle={Proceedings of the genetic and evolutionary computation conference},
  pages={456--465},
  year={2023}
}

@inproceedings{cmame,
  title={Covariance matrix adaptation for the rapid illumination of behavior space},
  author={Fontaine, Matthew C and Togelius, Julian and Nikolaidis, Stefanos and Hoover, Amy K},
  booktitle={Proceedings of the 2020 genetic and evolutionary computation conference},
  pages={94--102},
  year={2020}
}

@article{sepcmamae,
  title={Training diverse high-dimensional controllers by scaling covariance matrix adaptation map-annealing},
  author={Tjanaka, Bryon and Fontaine, Matthew C and Lee, David H and Kalkar, Aniruddha and Nikolaidis, Stefanos},
  journal={IEEE Robotics and Automation Letters},
  volume={8},
  number={10},
  pages={6771--6778},
  year={2023},
  publisher={IEEE}
}

@article{domino,
  title={Discovering policies with domino: Diversity optimization maintaining near optimality},
  author={Zahavy, Tom and Schroecker, Yannick and Behbahani, Feryal and Baumli, Kate and Flennerhag, Sebastian and Hou, Shaobo and Singh, Satinder},
  journal={arXiv preprint arXiv:2205.13521},
  year={2022}
}

@article{dvd,
  title={Effective diversity in population based reinforcement learning},
  author={Parker-Holder, Jack and Pacchiano, Aldo and Choromanski, Krzysztof M and Roberts, Stephen J},
  journal={Advances in Neural Information Processing Systems},
  volume={33},
  pages={18050--18062},
  year={2020}
}

@inproceedings{liu2021aps,
  title={Aps: Active pretraining with successor features},
  author={Liu, Hao and Abbeel, Pieter},
  booktitle={International Conference on Machine Learning},
  pages={6736--6747},
  year={2021},
  organization={PMLR}
}

@article{diayn,
  title={Diversity is all you need: Learning skills without a reward function},
  author={Eysenbach, Benjamin and Gupta, Abhishek and Ibarz, Julian and Levine, Sergey},
  journal={arXiv preprint arXiv:1802.06070},
  year={2018}
}

@inproceedings{park2024metra,
  title={Metra: Scalable unsupervised rl with metric-aware abstraction},
  author={Park, Seohong and Rybkin, Oleh and Levine, Sergey},
  booktitle={International Conference on Learning Representations},
  volume={2024},
  pages={18579--18603},
  year={2024}
}

@inproceedings{choi2021variational,
  title={Variational empowerment as representation learning for goal-conditioned reinforcement learning},
  author={Choi, Jongwook and Sharma, Archit and Lee, Honglak and Levine, Sergey and Gu, Shixiang Shane},
  booktitle={International conference on machine learning},
  pages={1953--1963},
  year={2021},
  organization={PMLR}
}

@article{dipg,
  title={Diversity-inducing policy gradient: Using maximum mean discrepancy to find a set of diverse policies},
  author={Masood, Muhammad A and Doshi-Velez, Finale},
  journal={arXiv preprint arXiv:1906.00088},
  year={2019}
}

@inproceedings{
autoqd,
title={Auto{QD}: Automatic Discovery of Diverse Behaviors with Quality-Diversity Optimization},
author={Saeed Hedayatian and Stefanos Nikolaidis},
booktitle={The Fourteenth International Conference on Learning Representations},
year={2026},
url={https://openreview.net/forum?id=FNnJIf4ymV}
}

@book{puterman2014markov,
  title={Markov decision processes: discrete stochastic dynamic programming},
  author={Puterman, Martin L},
  year={2014},
  publisher={John Wiley \& Sons}
}

@inproceedings{gae,
  author       = {John Schulman and
                  Philipp Moritz and
                  Sergey Levine and
                  Michael I. Jordan and
                  Pieter Abbeel},
  title        = {High-Dimensional Continuous Control Using Generalized Advantage Estimation},
  booktitle    = {4th International Conference on Learning Representations, {ICLR} 2016,
                  San Juan, Puerto Rico, May 2-4, 2016, Conference Track Proceedings},
  year         = {2016},
  url          = {http://arxiv.org/abs/1506.02438},
}

@inproceedings{brax,
  author       = {C. Daniel Freeman and
                  Erik Frey and
                  Anton Raichuk and
                  Sertan Girgin and
                  Igor Mordatch and
                  Olivier Bachem},
  title        = {Brax - {A} Differentiable Physics Engine for Large Scale Rigid Body
                  Simulation},
  booktitle    = {Proceedings of the Neural Information Processing Systems Track on
                  Datasets and Benchmarks 1, NeurIPS Datasets and Benchmarks 2021, December
                  2021, virtual},
  year         = {2021},
  url          = {https://datasets-benchmarks-proceedings.neurips.cc/paper/2021/hash/d1f491a404d6854880943e5c3cd9ca25-Abstract-round1.html}
}

@inproceedings{pyribs,
  author       = {Bryon Tjanaka and
                  Matthew C. Fontaine and
                  David H. Lee and
                  Yulun Zhang and
                  Nivedit Reddy Balam and
                  Nathaniel Dennler and
                  Sujay S. Garlanka and
                  Nikitas Dimitri Klapsis and
                  Stefanos Nikolaidis},
  title        = {pyribs: {A} Bare-Bones Python Library for Quality Diversity Optimization},
  booktitle    = {Proceedings of the Genetic and Evolutionary Computation Conference,
                  {GECCO} 2023, Lisbon, Portugal, July 15-19, 2023},
  pages        = {220--229},
  publisher    = {{ACM}},
  year         = {2023},
  url          = {https://doi.org/10.1145/3583131.3590374},
}

@article{vendi,
  title={The Vendi Score: A Diversity Evaluation Metric for Machine Learning},
  author={Friedman, Dan and Dieng, Adji Bousso},
  journal={Transactions on Machine Learning Research},
  year={2023}
}

@inproceedings{qvs,
  title={Quality-Weighted Vendi Scores And Their Application To Diverse Experimental Design},
  author={Nguyen, Quan and Dieng, Adji Bousso},
  booktitle={Forty-first International Conference on Machine Learning},
  year={2024}
}

@inproceedings{wang2024quality,
  title={Quality-diversity with limited resources},
  author={Wang, Ren-Jian and Xue, Ke and Guan, Cong and Qian, Chao},
  booktitle={Proceedings of the 41st International Conference on Machine Learning},
  pages={51984--52001},
  year={2024}
}

@inproceedings{td3,
  title={Addressing function approximation error in actor-critic methods},
  author={Fujimoto, Scott and Hoof, Herke and Meger, David},
  booktitle={International conference on machine learning},
  pages={1587--1596},
  year={2018},
  organization={PMLR}
}

@article{chalumeau2024qdax,
    title={Qdax: A library for quality-diversity and population-based algorithms with hardware acceleration},
    author={Chalumeau, Felix and Lim, Bryan and Boige, Raphael and Allard, Maxime and Grillotti, Luca and Flageat, Manon and Mac{\'e}, Valentin and Richard, Guillaume and Flajolet, Arthur and Pierrot, Thomas and others},
    journal={Journal of Machine Learning Research},
    volume={25},
    number={108},
    pages={1--16},
    year={2024}
}

@inproceedings{
gym,
title={Gymnasium: A Standard Interface for Reinforcement Learning Environments},
author={Mark Towers and Ariel Kwiatkowski and John U. Balis and Gianluca De Cola and Tristan Deleu and Manuel Goul{\~a}o and Kallinteris Andreas and Markus Krimmel and Arjun KG and Rodrigo De Lazcano Perez-Vicente and J K Terry and Andrea Pierr{\'e} and Sander V Schulhoff and Jun Jet Tai and Hannah Tan and Omar G. Younis},
booktitle={The Thirty-ninth Annual Conference on Neural Information Processing Systems Datasets and Benchmarks Track},
year={2026},
url={https://openreview.net/forum?id=qPMLvJxtPK}
}

@article{mnih2015human,
  title={Human-level control through deep reinforcement learning},
  author={Mnih, Volodymyr and Kavukcuoglu, Koray and Silver, David and Rusu, Andrei A and Veness, Joel and Bellemare, Marc G and Graves, Alex and Riedmiller, Martin and Fidjeland, Andreas K and Ostrovski, Georg and others},
  journal={nature},
  volume={518},
  number={7540},
  pages={529--533},
  year={2015},
  publisher={Nature Publishing Group}
}

@article{sb3,
  author  = {Antonin Raffin and Ashley Hill and Adam Gleave and Anssi Kanervisto and Maximilian Ernestus and Noah Dormann},
  title   = {Stable-Baselines3: Reliable Reinforcement Learning Implementations},
  journal = {Journal of Machine Learning Research},
  year    = {2021},
  volume  = {22},
  number  = {268},
  pages   = {1--8},
  url     = {http://jmlr.org/papers/v22/20-1364.html}
}

@article{dcrlme,
  title={Synergizing quality-diversity with descriptor-conditioned reinforcement learning},
  author={Faldor, Maxence and Chalumeau, F{\'e}lix and Flageat, Manon and Cully, Antoine},
  journal={ACM Transactions on Evolutionary Learning},
  volume={5},
  number={1},
  pages={1--35},
  year={2025},
  publisher={ACM New York, NY}
}

@inproceedings{qdac,
  author       = {Luca Grillotti and
                  Maxence Faldor and
                  Borja G. Le{\'{o}}n and
                  Antoine Cully},
  title        = {Quality-Diversity Actor-Critic: Learning High-Performing and Diverse
                  Behaviors via Value and Successor Features Critics},
  booktitle    = {Forty-first International Conference on Machine Learning, {ICML} 2024,
                  Vienna, Austria, July 21-27, 2024},
  pages        = {16416--16459},
  publisher    = {{PMLR} / OpenReview.net},
  year         = {2024},
  url          = {https://proceedings.mlr.press/v235/grillotti24a.html}
}

@inproceedings{bellmare2015arcade,
  author       = {Marc G. Bellemare and
                  Yavar Naddaf and
                  Joel Veness and
                  Michael Bowling},
  title        = {The Arcade Learning Environment: An Evaluation Platform for General
                  Agents (Extended Abstract)},
  booktitle    = {Proceedings of the Twenty-Fourth International Joint Conference on
                  Artificial Intelligence, {IJCAI} 2015, Buenos Aires, Argentina, July
                  25-31, 2015},
  pages        = {4148--4152},
  publisher    = {{AAAI} Press},
  year         = {2015}
}

@inproceedings{machado2018revisiting,
  author       = {Marlos C. Machado and
                  Marc G. Bellemare and
                  Erik Talvitie and
                  Joel Veness and
                  Matthew J. Hausknecht and
                  Michael Bowling},
  title        = {Revisiting the Arcade Learning Environment: Evaluation Protocols and
                  Open Problems for General Agents (Extended Abstract)},
  booktitle    = {Proceedings of the Twenty-Seventh International Joint Conference on
                  Artificial Intelligence, {IJCAI} 2018, July 13-19, 2018, Stockholm,
                  Sweden},
  pages        = {5573--5577},
  publisher    = {ijcai.org},
  year         = {2018}
}

@article{qwen3,
  author       = {Qwen Team},
  title        = {Qwen3 Technical Report},
  journal      = {CoRR},
  volume       = {abs/2505.09388},
  year         = {2025},
  url          = {https://doi.org/10.48550/arXiv.2505.09388},
}

@misc{granite,
  title={Granite 4.1 {LLMs}: How They're Built},
  author={{IBM Granite Team}},
  howpublished={\url{https://huggingface.co/blog/ibm-granite/granite-4-1}},
  year={2026}
}

@inproceedings{hendrycks2021math,
  author       = {Dan Hendrycks and
                  Collin Burns and
                  Saurav Kadavath and
                  Akul Arora and
                  Steven Basart and
                  Eric Tang and
                  Dawn Song and
                  Jacob Steinhardt},
  title        = {Measuring Mathematical Problem Solving With the {MATH} Dataset},
  booktitle    = {Proceedings of the Neural Information Processing Systems Track on
                  Datasets and Benchmarks 1, NeurIPS Datasets and Benchmarks 2021, December
                  2021, virtual},
  year         = {2021},
  url          = {https://datasets-benchmarks-proceedings.neurips.cc/paper/2021/hash/be83ab3ecd0db773eb2dc1b0a17836a1-Abstract-round2.html}
}

@inproceedings{lightman2024verify,
  author       = {Hunter Lightman and
                  Vineet Kosaraju and
                  Yuri Burda and
                  Harrison Edwards and
                  Bowen Baker and
                  Teddy Lee and
                  Jan Leike and
                  John Schulman and
                  Ilya Sutskever and
                  Karl Cobbe},
  title        = {Let's Verify Step by Step},
  booktitle    = {The Twelfth International Conference on Learning Representations,
                  {ICLR} 2024, Vienna, Austria, May 7-11, 2024},
  publisher    = {OpenReview.net},
  year         = {2024},
  url          = {https://openreview.net/forum?id=v8L0pN6EOi}
}

@inproceedings{lewkowycz2022minerva,
  author       = {Aitor Lewkowycz and
                  Anders Andreassen and
                  David Dohan and
                  Ethan Dyer and
                  Henryk Michalewski and
                  Vinay V. Ramasesh and
                  Ambrose Slone and
                  Cem Anil and
                  Imanol Schlag and
                  Theo Gutman{-}Solo and
                  Yuhuai Wu and
                  Behnam Neyshabur and
                  Guy Gur{-}Ari and
                  Vedant Misra},
  title        = {Solving Quantitative Reasoning Problems with Language Models},
  booktitle    = {Advances in Neural Information Processing Systems 35: Annual Conference
                  on Neural Information Processing Systems 2022, NeurIPS 2022, New Orleans,
                  LA, USA, November 28 - December 9, 2022},
  year         = {2022},
  url          = {http://papers.nips.cc/paper\_files/paper/2022/hash/18abbeef8cfe9203fdf9053c9c4fe191-Abstract-Conference.html}
}

@inproceedings{he2024olympiadbench,
  author       = {Chaoqun He and
                  Renjie Luo and
                  Yuzhuo Bai and
                  Shengding Hu and
                  Zhen Leng Thai and
                  Junhao Shen and
                  Jinyi Hu and
                  Xu Han and
                  Yujie Huang and
                  Yuxiang Zhang and
                  Jie Liu and
                  Lei Qi and
                  Zhiyuan Liu and
                  Maosong Sun},
  title        = {OlympiadBench: {A} Challenging Benchmark for Promoting {AGI} with
                  Olympiad-Level Bilingual Multimodal Scientific Problems},
  booktitle    = {Proceedings of the 62nd Annual Meeting of the Association for Computational
                  Linguistics (Volume 1: Long Papers), {ACL} 2024, Bangkok, Thailand,
                  August 11-16, 2024},
  pages        = {3828--3850},
  publisher    = {Association for Computational Linguistics},
  year         = {2024},
  url          = {https://doi.org/10.18653/v1/2024.acl-long.211}
}

@article{huang2022cleanrl,
  author       = {Shengyi Huang and
                  Rousslan Fernand Julien Dossa and
                  Chang Ye and
                  Jeff Braga and
                  Dipam Chakraborty and
                  Kinal Mehta and
                  Jo{\~{a}}o G. M. Ara{\'{u}}jo},
  title        = {CleanRL: High-quality Single-file Implementations of Deep Reinforcement
                  Learning Algorithms},
  journal      = {J. Mach. Learn. Res.},
  volume       = {23},
  pages        = {274:1--274:18},
  year         = {2022},
  url          = {https://jmlr.org/papers/v23/21-1342.html}
}

@inproceedings{hu2022lora,
  author       = {Edward J. Hu and
                  Yelong Shen and
                  Phillip Wallis and
                  Zeyuan Allen{-}Zhu and
                  Yuanzhi Li and
                  Shean Wang and
                  Lu Wang and
                  Weizhu Chen},
  title        = {LoRA: Low-Rank Adaptation of Large Language Models},
  booktitle    = {The Tenth International Conference on Learning Representations, {ICLR}
                  2022, Virtual Event, April 25-29, 2022},
  publisher    = {OpenReview.net},
  year         = {2022},
  url          = {https://openreview.net/forum?id=nZeVKeeFYf9}
}

@inproceedings{hu2025openreasonerzero,
  author       = {Jingcheng Hu and
                  Yinmin Zhang and
                  Qi Han and
                  Daxin Jiang and
                  Xiangyu Zhang and
                  Heung{-}Yeung Shum},
  title        = {Open-Reasoner-Zero: An Open Source Approach to Scaling Up Reinforcement
                  Learning on the Base Model},
  booktitle    = {Advances in Neural Information Processing Systems 38: Annual Conference
                  on Neural Information Processing Systems 2025, NeurIPS 2025, San Diego,
                  CA, USA, December 2-7, 2025 / Mexico City, Mexico, November 30 - December
                  5, 2025},
  year         = {2025},
  url          = {http://papers.nips.cc/paper\_files/paper/2025/hash/ed873d79e7c268c020c4b4db13a2812a-Abstract-Conference.html}
}

@inproceedings{weng2022envpool,
  author       = {Jiayi Weng and
                  Min Lin and
                  Shengyi Huang and
                  Bo Liu and
                  Denys Makoviichuk and
                  Viktor Makoviychuk and
                  Zichen Liu and
                  Yufan Song and
                  Ting Luo and
                  Yukun Jiang and
                  Zhongwen Xu and
                  Shuicheng Yan},
  title        = {EnvPool: {A} Highly Parallel Reinforcement Learning Environment Execution
                  Engine},
  booktitle    = {Advances in Neural Information Processing Systems 35: Annual Conference
                  on Neural Information Processing Systems 2022, NeurIPS 2022, New Orleans,
                  LA, USA, November 28 - December 9, 2022},
  year         = {2022},
  url          = {http://papers.nips.cc/paper\_files/paper/2022/hash/8caaf08e49ddbad6694fae067442ee21-Abstract-Datasets\_and\_Benchmarks.html}
}

@inproceedings{vanhasselt2016deep,
  author       = {Hado van Hasselt and
                  Arthur Guez and
                  David Silver},
  title        = {Deep Reinforcement Learning with Double Q-Learning},
  booktitle    = {Proceedings of the Thirtieth {AAAI} Conference on Artificial Intelligence,
                  February 12-17, 2016, Phoenix, Arizona, {USA}},
  pages        = {2094--2100},
  publisher    = {{AAAI} Press},
  year         = {2016},
  url          = {https://doi.org/10.1609/aaai.v30i1.10295}
}

@article{zeng2025simplerlzoo,
  author       = {Weihao Zeng and
                  Yuzhen Huang and
                  Qian Liu and
                  Wei Liu and
                  Keqing He and
                  Zejun Ma and
                  Junxian He},
  title        = {SimpleRL-Zoo: Investigating and Taming Zero Reinforcement Learning
                  for Open Base Models in the Wild},
  journal      = {CoRR},
  volume       = {abs/2503.18892},
  year         = {2025},
  url          = {https://doi.org/10.48550/arXiv.2503.18892},
}

@article{chen2021evaluating,
  author       = {Mark Chen and
                  Jerry Tworek and
                  Heewoo Jun and
                  Qiming Yuan and
                  Henrique Pond{\'{e}} de Oliveira Pinto and
                  Jared Kaplan and
                  Harri Edwards and
                  Yuri Burda and
                  Nicholas Joseph and
                  Greg Brockman and
                  Alex Ray and
                  Raul Puri and
                  Gretchen Krueger and
                  Michael Petrov and
                  Heidy Khlaaf and
                  Girish Sastry and
                  Pamela Mishkin and
                  Brooke Chan and
                  Scott Gray and
                  Nick Ryder and
                  Mikhail Pavlov and
                  Alethea Power and
                  Lukasz Kaiser and
                  Mohammad Bavarian and
                  Clemens Winter and
                  Philippe Tillet and
                  Felipe Petroski Such and
                  Dave Cummings and
                  Matthias Plappert and
                  Fotios Chantzis and
                  Elizabeth Barnes and
                  Ariel Herbert{-}Voss and
                  William Hebgen Guss and
                  Alex Nichol and
                  Alex Paino and
                  Nikolas Tezak and
                  Jie Tang and
                  Igor Babuschkin and
                  Suchir Balaji and
                  Shantanu Jain and
                  William Saunders and
                  Christopher Hesse and
                  Andrew N. Carr and
                  Jan Leike and
                  Joshua Achiam and
                  Vedant Misra and
                  Evan Morikawa and
                  Alec Radford and
                  Matthew Knight and
                  Miles Brundage and
                  Mira Murati and
                  Katie Mayer and
                  Peter Welinder and
                  Bob McGrew and
                  Dario Amodei and
                  Sam McCandlish and
                  Ilya Sutskever and
                  Wojciech Zaremba},
  title        = {Evaluating Large Language Models Trained on Code},
  journal      = {CoRR},
  volume       = {abs/2107.03374},
  year         = {2021},
  url          = {https://doi.org/10.48550/arXiv.2107.03374}
}

@inproceedings{kwon2023efficient,
  author       = {Woosuk Kwon and
                  Zhuohan Li and
                  Siyuan Zhuang and
                  Ying Sheng and
                  Lianmin Zheng and
                  Cody Hao Yu and
                  Joseph Gonzalez and
                  Hao Zhang and
                  Ion Stoica},
  title        = {Efficient Memory Management for Large Language Model Serving with
                  PagedAttention},
  booktitle    = {Proceedings of the 29th Symposium on Operating Systems Principles,
                  {SOSP} 2023, Koblenz, Germany, October 23-26, 2023},
  pages        = {611--626},
  publisher    = {{ACM}},
  year         = {2023},
  url          = {https://doi.org/10.1145/3600006.3613165}
}

@article{aurora,
  author       = {Luca Grillotti and
                  Antoine Cully},
  title        = {Unsupervised Behavior Discovery With Quality-Diversity Optimization},
  journal      = {{IEEE} Trans. Evol. Comput.},
  volume       = {26},
  number       = {6},
  pages        = {1539--1552},
  year         = {2022},
  url          = {https://doi.org/10.1109/TEVC.2022.3159855}
}
\bibliographystyle{icml2026}

\newpage
\appendix
\onecolumn
\section{Derivations}
\label{app:proof}

This section proves Proposition~\ref{main:prop:pg} and then traces how RePPO's update follows from it.
We use the notation of Sections~\ref{main:sec:background-rl} and~\ref{main:sec:method}: $f_k$ and $\mathbf{b}_k$ are the quality and descriptor of policy $k$, $\mathbf{m}$ is the vector of per-step behavior signals, and $\ell$ is the interaction scale.

\FloatBarrier
\subsection{Proof of Proposition~\ref{main:prop:pg}}
\label{app:proof-prop}

We restate the policy-gradient identity of Section~\ref{main:sec:background-rl} in the form used by the proof.

\begin{lemma}
\label{app:lem:cumulant-pg}
For a per-step signal $c:\mathcal{S}\times\mathcal{A}\to\mathbb{R}$, let $J_c(\theta)=\mathbb{E}_{\tau\sim\pi_\theta}\big[\sum_{t\geq0}\gamma^t c(s_t,a_t)\big]$.
Under the usual regularity conditions,
\begin{align*}
\nabla_\theta J_c(\theta)
=\mathbb{E}_{\tau\sim\pi_\theta}\Big[\sum_{t\geq0}\gamma^t\nabla_\theta\log\pi_\theta(a_t\mid s_t)\,Q_c^{\pi_\theta}(s_t,a_t)\Big],
\end{align*}
and both $Q_c^{\pi_\theta}$ and $\nabla_\theta J_c$ are linear in $c$.
\end{lemma}

The identity is the policy-gradient theorem~\citep{reinforce} with $c$ in place of the task reward, and linearity follows from linearity of expectation.
By the definitions in~\eqref{main:eq:quality-return} and~\eqref{main:eq:additive-descriptor}, $f(\theta)=J_r(\theta)$ and $\mathbf{v}^\top\mathbf{b}(\theta)=J_{\mathbf{v}^\top\mathbf{m}}(\theta)$ for any fixed $\mathbf{v}\in\mathbb{R}^d$.

\begin{proof}[Proof of Proposition~\ref{main:prop:pg}]
Write $e_{ki}=\exp(-\|\mathbf{b}_k-\mathbf{b}_i\|_2^2/\ell^2)$.
Only the terms of~\eqref{main:eq:squad-objective} that contain index $k$ depend on $\theta_k$, so
\begin{align*}
\nabla_{\theta_k}S(\boldsymbol{\theta})
=\nabla_{\theta_k}f_k-\sum_{i\neq k}\nabla_{\theta_k}\big[\sqrt{f_kf_i}\,e_{ki}\big].
\end{align*}
Since $f_k>0$ and the policies are parameterized independently, the product and chain rules give
\begin{align*}
\nabla_{\theta_k}\big[\sqrt{f_kf_i}\,e_{ki}\big]
=\frac{1}{2}\sqrt{\frac{f_i}{f_k}}\,e_{ki}\,\nabla_{\theta_k}f_k
-\frac{2\sqrt{f_kf_i}}{\ell^2}\,e_{ki}\,\big(\nabla_{\theta_k}\mathbf{b}_k\big)^{\top}(\mathbf{b}_k-\mathbf{b}_i),
\end{align*}
where $\nabla_{\theta_k}\mathbf{b}_k$ is the Jacobian of $\mathbf{b}(\theta_k)$.
Summing over $i\neq k$ and collecting terms yields~\eqref{main:eq:population-gradient},
\begin{align*}
\nabla_{\theta_k}S(\boldsymbol{\theta})=W_k\nabla_{\theta_k}f_k+\big(\nabla_{\theta_k}\mathbf{b}_k\big)^{\top}\mathbf{c}_k,
\end{align*}
with $W_k$ and $\mathbf{c}_k$ given by~\eqref{main:eq:wk} and~\eqref{main:eq:repulsion_direction}.
Both are evaluated at the current population, so they are fixed numbers in this expression.

It remains to express both terms as policy gradients.
The first is $W_k\nabla_{\theta_k}J_r(\theta_k)=\nabla_{\theta_k}J_{W_kr}(\theta_k)$.
For the second, $\big(\nabla_{\theta_k}\mathbf{b}_k\big)^{\top}\mathbf{c}_k=\nabla_{\theta_k}\big[\mathbf{v}^\top\mathbf{b}(\theta_k)\big]_{\mathbf{v}=\mathbf{c}_k}=\nabla_{\theta_k}J_{\mathbf{c}_k^\top\mathbf{m}}(\theta_k)$ with $\mathbf{c}_k$ held fixed.
By the linearity in Lemma~\ref{app:lem:cumulant-pg}, the sum of the two gradients is the policy gradient of $J_{\widetilde{r}_k}$ for $\widetilde{r}_k=W_kr+\mathbf{c}_k^\top\mathbf{m}$, which is~\eqref{main:eq:shaped-reward}.
\end{proof}

Note that holding $W_k$ and $\mathbf{c}_k$ fixed is not a stop-gradient approximation.
The chain rule has already accounted for their dependence on $\theta_k$, which is what produces the second term of $W_k$ and the direction $\mathbf{c}_k$.

\FloatBarrier
\subsection{From the objective to the RePPO update}
\label{app:derivation}

\paragraph{Interactions in transformed coordinates.}
Descriptors in our domains are normalized to $[-1,1]^d$, while the Gaussian kernel treats behavior space as unbounded.
Following SQUAD~\citep{softqd}, we therefore define the interaction on the logit coordinates $g(\mathbf{b})$ of Section~\ref{main:sec:method-practical}, replacing $e_{ki}$ with $\exp(-\|g(\mathbf{b}_k)-g(\mathbf{b}_i)\|_2^2/\ell^2)$.
Repeating the proof with this kernel leaves $W_k$ unchanged in form and replaces the behavior term by $\big(\nabla_{\theta_k}\mathbf{b}_k\big)^{\top}J_g(\mathbf{b}_k)^{\top}\mathbf{c}^{g}_k$, where $\mathbf{c}^{g}_k$ is~\eqref{main:eq:repulsion_direction} evaluated at the transformed descriptors and
\begin{align}
J_g(\mathbf{b})=\operatorname{diag}\!\left(\frac{2}{1-b_1^2},\ldots,\frac{2}{1-b_d^2}\right)
\label{app:eq:jacobian}
\end{align}
is the Jacobian of the componentwise logit $g(b)=\log\big((1+b)/(1-b)\big)$.
The shaped reward thus weights the original signals $\mathbf{m}$ by $J_g(\mathbf{b}_k)^{\top}\mathbf{c}^{g}_k$, which is the conversion in~\eqref{main:eq:practical-repulsion}.
Because the Jacobian grows near the bounds, a descriptor that is close to saturation receives a larger weight, matching the stretching of the kernel geometry there.
For numerical stability, the normalized coordinates $(\widehat{\mathbf{d}}_k+\mathbf{1})/2$ are clipped to $[\epsilon_b,1-\epsilon_b]$ before both the logit and its Jacobian are evaluated.

\paragraph{Departures from the exact gradient.}
RePPO modifies this exact update in several ways to obtain a working algorithm. Here, we enumerate them while Appendix~\ref{app:ablations} examines the task coefficient, the cap, and the local interactions empirically.
\begin{enumerate}
\item \emph{Smoothed episode statistics.} The qualities and descriptors are replaced by exponential moving averages of completed-episode returns and episode-level descriptors, which reduces noise in the interaction at the cost of a small lag.
The averages are updated once per iteration from the episodes completed in that iteration, and a policy that completes no episode keeps its previous values.
Following standard practice, the advantages use discounted GAE as a surrogate for the gradient of these undiscounted statistics.
\item \emph{Shifted qualities.} The objective requires $f_k>0$, whereas returns can be negative.
RePPO uses $f_k^+=\widehat{F}_k-\min_j\widehat{F}_j+\epsilon_f$, which is positive and preserves the ordering and differences of the population's qualities.
\item \emph{Local interactions.} The sum in~\eqref{main:eq:repulsion_direction} is restricted to the $K$ nearest neighbors in transformed space.
Under the Gaussian kernel, the omitted terms decay exponentially with distance, so the contribution of farther neighbors is naturally vanishing.
\item \emph{Unit task coefficient and capped repulsion.} RePPO fixes $W_k=1$ and rescales the diversity weights to norm at most $C_{\max}$, preserving their direction.
\item \emph{Channel standardization.} Each advantage channel is standardized before the channels are combined with~\eqref{main:eq:composite-advantage}.
The diversity weights $\mathbf{p}_k$ then set the relative influence of the channels on the update independently of their numerical scales.
\end{enumerate}

\FloatBarrier
\section{Evaluation protocol}
\label{app:evaluation}

\paragraph{Re-evaluation.}
After training, every policy is evaluated anew, and these evaluations define all reported numbers.
In continuous control, each policy runs $256$ episodes with deterministic actions.
In Atari, each policy plays $32$ full games, over all lives and up to the cap of $27000$ agent steps.
Finally, LLM members are evaluated on $8$ samples per test problem.
Fitness is the mean undiscounted return and the behavior descriptor is the mean episode descriptor over these evaluations.

\paragraph{Quality filter.}
For each task, let $f^\star$ be the fitness of the best policy found by any method in any seed.
Each population or archive retains the policies whose fitness is at least $0.1f^\star$ ($0.25f^\star$ on Ant), and all metrics except best fitness are computed on the retained policies.
The filter is applied identically to every method, and archives are otherwise used as they are.
Note that in the LLM experiments, we remove no members.

\paragraph{Metrics.}
For retained policies with mean descriptors $\bar{\mathbf{b}}_1,\ldots,\bar{\mathbf{b}}_n\in[-1,1]^d$, let $G_{ij}=\exp\!\big(-\|\bar{\mathbf{b}}_i-\bar{\mathbf{b}}_j\|_2^2/(2\sigma^2)\big)$ with $\sigma^2=0.5$, and let $\lambda_1,\ldots,\lambda_n$ be the eigenvalues of $G/n$.
The Vendi Score~\citep{vendi} is $\exp\!\big(-\sum_i\lambda_i\log\lambda_i\big)$, which ranges from $1$, when all descriptors coincide, to $n$, when they are mutually far apart.
The quality-weighted Vendi Score~\citep{qvs} is the mean fitness of the retained policies multiplied by their Vendi Score.
In continuous control and Atari, we report means and standard errors over seeds.
In the LLM experiments, fitness is a member's pass@1, the descriptors are averaged over the member's test solutions.

\FloatBarrier
\section{Continuous control}
\label{app:cc}

\FloatBarrier
\subsection{Tasks and descriptors}
\label{app:cc-setup}

We use the Ant, HalfCheetah, Humanoid, and Walker2D tasks of Brax~0.14~\citep{brax} with the MJX physics backend and episodes of at most $1000$ steps.
Fitness is the default task reward, summed over an episode.
The descriptor is the fraction of the episode each foot spends in contact, mapped to $[-1,1]$.
The tracked links are the four lower legs of Ant, the two feet of HalfCheetah and Walker2D, and Humanoid.
During training, the behavior signal of a link is its signed contact indicator $2c_t-1$.

\FloatBarrier
\subsection{RePPO and Ind-PPO}
\label{app:cc-hparams}

Each policy has separate actor, task-critic, and behavior-critic networks, each an MLP with two hidden layers of $128$ $\tanh$ units.
The actor outputs the mean of a diagonal Gaussian with a state-independent standard deviation, and actions are squashed with $\tanh$.
Observations are normalized with per-policy running statistics and clipped to $\pm10$, and training rewards are clipped to $\pm10$ per step.
Table~\ref{app:tab:cc-hparams} lists the hyperparameters.
Ind-PPO uses identical settings with the diversity weights set to zero.

\begin{table}[ht]
\centering
\caption{RePPO hyperparameters in continuous control. The learning rate decays as $\eta_t=\eta_0\max(0.05,\,1-2t/S)$ in the number of environment steps $t$.}
\label{app:tab:cc-hparams}
\small
\begin{tabular}{@{}lcccc@{}}
\toprule
 & Ant & HalfCheetah & Humanoid & Walker2D \\
\midrule
Learning rate $\eta_0$ & $3\times10^{-4}$ & $10^{-3}$ & $10^{-3}$ & $10^{-3}$ \\
Schedule horizon $S$ & $2.05$~B & $4.10$~B & $4.10$~B & $4.10$~B \\
Update epochs & $10$ & $10$ & $10$ & $5$ \\
Entropy coefficient & $0.005$ & $0$ & $0$ & $0$ \\
Value clipping & yes & no & no & no \\
Kernel bandwidth $\ell^2$ & $0.1$ & $0.5$ & $0.1$ & $0.1$ \\
\midrule
\multicolumn{5}{@{}l}{Shared across tasks} \\
\midrule
Population size $N$ & \multicolumn{4}{c}{$32$} \\
Environments per policy / rollout length & \multicolumn{4}{c}{$64$ / $1024$} \\
Total environment steps & \multicolumn{4}{c}{$2.05\times10^{9}$} \\
Minibatches & \multicolumn{4}{c}{$32$} \\
Discount $\gamma$ / GAE $\lambda$ & \multicolumn{4}{c}{$0.99$ / $0.95$} \\
PPO clip / value coefficient / max gradient norm & \multicolumn{4}{c}{$0.2$ / $0.5$ / $0.5$} \\
Neighbors $K$ / repulsion cap $C_{\max}$ & \multicolumn{4}{c}{$8$ / $1.0$} \\
EMA rates for $\widehat{F}_k$ and $\widehat{\mathbf{d}}_k$ & \multicolumn{4}{c}{$0.99$} \\
$\epsilon_f$ / $\epsilon_b$ & \multicolumn{4}{c}{$1$ / $10^{-6}$} \\
\bottomrule
\end{tabular}
\end{table}

\FloatBarrier
\subsection{Baselines}
\label{app:cc-baselines}

All baselines use the same tasks, descriptors, and evaluation as RePPO, and their archives are re-evaluated with the protocol of Appendix~\ref{app:evaluation}.
Table~\ref{app:tab:baselines} summarizes their main settings.

\paragraph{PPGA.}
We port the official implementation~\citep{ppga} to the current version of Brax with the MJX backend.
All hyperparameters follow the original work, except the number of outer iterations, which we set to $200$ so that the budget approaches that of the other methods.
\\
More specifically, PPGA uses one CMA-MAEGA emitter with $300$ branching solutions, $10$ gradient and $10$ mean-update PPO iterations per outer iteration on $3000$ parallel environments with rollouts of $128$ steps, $4$ epochs over $8$ minibatches, value coefficient $2$, no entropy bonus, and a target KL of $0.008$.
It uses archive learning rates of $0.1$ (Ant, Humanoid) and $0.5$ (HalfCheetah, Walker2D).
The initial step sizes are $3.0$, $1.0$, $0.5$, and $1.0$ for Ant, HalfCheetah, Humanoid, and Walker2D, respectively.
PPO's learning rate is $10^{-3}$, except on Humanoid where it is $3\times10^{-4}$.

\paragraph{DCRL-ME and PGA-ME.}
Both build on the QDax implementations~\citep{chalumeau2024qdax} and use the hyperparameters reported by their authors, with the number of iterations matched to the budget.
\\
In DCRL-ME each iteration produces $128$ offspring by isoline variation ($\sigma_{\mathrm{iso}}=0.005$, $\sigma_{\mathrm{line}}=0.05$), $64$ by $150$ descriptor-conditioned policy-gradient steps (learning rate $5\times10^{-3}$), and $64$ by specializing the descriptor-conditioned actor.
The critic has two hidden layers of $256$ units and is trained for $3000$ steps per iteration with TD3 settings (batch $100$, replay buffer $10^6$, discount $0.99$, policy noise $0.2$, noise clip $0.5$, target rate $0.005$, policy delay $2$).
\\
PGA-ME generates $300$ offspring in each iteration, half by isoline variation and half by $100$ policy-gradient steps, with $300$ critic updates per iteration.

\paragraph{Sep-CMA-MAE.}
We use the pyribs implementation~\citep{pyribs} with the hyperparameters of \citet{sepcmamae}, evaluating each candidate on one episode.
In particular, it has five emitters with batches of $40$ using an initial step size of $0.02$, an archive learning rate of $0.001$, improvement ranking, and the basic restart rule.
Its budget is $2.048$~M evaluations of maximum length $1000$.

\begin{table}[ht]
\centering
\caption{Main settings of the archive-based baselines in continuous control. Environment steps count full-length evaluation episodes, except for Sep-CMA-MAE, whose budget is a number of evaluations.}
\label{app:tab:baselines}
\small
\setlength{\tabcolsep}{5pt}
\begin{tabular}{@{}lcccc@{}}
\toprule
 & PPGA & DCRL-ME & PGA-ME & Sep-CMA-MAE \\
\midrule
Archive & grid, $50^2$ ($10^4$ Ant) & CVT, $1024$ & grid, $50^2$ ($10^4$ Ant) & grid, $32^2$ ($6^4$ Ant) \\
Iterations & $200$ & $8000$ & $6827$ & -- \\
Policy network & $128$-$128$ $\tanh$ & $128$-$128$ ReLU & $128$-$128$ ReLU & $128$-$128$ $\tanh$ \\
Gradient estimator & PPO, $3000$ envs & TD3, desc.-conditioned & TD3 & -- \\
Environment steps & $2.74$~B & $2.05$~B & $2.05$~B & $\leq2.05$~B \\
\bottomrule
\end{tabular}
\end{table}

\FloatBarrier
\subsection{Additional results}
\label{app:cc-results}

Table~\ref{app:tab:cc-full} reports the numbers behind the top row of Figure~\ref{main:fig:qd}, together with the best fitness and the number of policies that pass the quality filter.
Despite the larger population sizes of archive-based methods, RePPO has the highest Vendi Score on two of the tasks and the second highest on the other two.

\begin{table}[ht]
\centering
\caption{Final populations in continuous control, as mean $\pm$ standard error over seeds. All other metrics belong to the retained policies}
\label{app:tab:cc-full}
\small
\setlength{\tabcolsep}{4pt}
\begin{tabular}{@{}llccccc@{}}
\toprule
Task & Method & Best fitness & Mean fitness & Vendi Score & qVS & Retained \\
\midrule
\multirow{6}{*}{Ant}
& RePPO & $4090\pm68$ & $3011\pm68$ & $1.82\pm0.09$ & $\mathbf{5492\pm328}$ & $12$ \\
& DCRL-ME & $1335\pm48$ & $1174\pm10$ & $\mathbf{2.69\pm0.21}$ & $3157\pm258$ & $61$ \\
& PPGA & $1183\pm113$ & $752\pm309$ & $1.16\pm0.48$ & $1439\pm588$ & $2$ \\
& PGA-ME & $786\pm32$ & $0\pm0$ & $0.00\pm0.00$ & $0\pm0$ & $0$ \\
& Sep-CMA-MAE & $306\pm175$ & $0\pm0$ & $0.00\pm0.00$ & $0\pm0$ & $0$ \\
& Ind-PPO & $\mathbf{4339\pm31}$ & $\mathbf{3683\pm72}$ & $1.10\pm0.05$ & $4023\pm103$ & $18$ \\
\midrule
\multirow{6}{*}{HalfCheetah}
& RePPO & $9536\pm274$ & $4419\pm82$ & $\mathbf{6.02\pm0.12}$ & $\mathbf{26571\pm263}$ & $30$ \\
& DCRL-ME & $7449\pm234$ & $4471\pm154$ & $5.55\pm0.03$ & $24801\pm847$ & $1024$ \\
& PPGA & $5398\pm906$ & $2659\pm390$ & $3.17\pm0.43$ & $8104\pm1217$ & $696$ \\
& PGA-ME & $5798\pm147$ & $1972\pm90$ & $3.78\pm0.07$ & $7433\pm237$ & $1285$ \\
& Sep-CMA-MAE & $4677\pm275$ & $3005\pm175$ & $2.71\pm0.13$ & $8197\pm778$ & $402$ \\
& Ind-PPO & $\mathbf{10185\pm29}$ & $\mathbf{9125\pm58}$ & $1.04\pm0.00$ & $9454\pm72$ & $32$ \\
\midrule
\multirow{6}{*}{Humanoid}
& RePPO & $8190\pm134$ & $6130\pm64$ & $\mathbf{6.38\pm0.09}$ & $\mathbf{39089\pm286}$ & $32$ \\
& DCRL-ME & $7594\pm456$ & $4263\pm125$ & $3.27\pm0.08$ & $13964\pm672$ & $749$ \\
& PPGA & $\mathbf{9907\pm303}$ & $6298\pm115$ & $2.99\pm0.38$ & $18792\pm2322$ & $598$ \\
& PGA-ME & $4961\pm121$ & $2026\pm60$ & $1.43\pm0.05$ & $2918\pm191$ & $720$ \\
& Sep-CMA-MAE & $576\pm8$ & $0\pm0$ & $0.00\pm0.00$ & $0\pm0$ & $0$ \\
& Ind-PPO & $8601\pm134$ & $\mathbf{6744\pm80}$ & $1.28\pm0.01$ & $8654\pm147$ & $32$ \\
\midrule
\multirow{6}{*}{Walker2D}
& RePPO & $\mathbf{7764\pm274}$ & $4035\pm60$ & $4.59\pm0.14$ & $\mathbf{18511\pm598}$ & $31$ \\
& DCRL-ME & $6512\pm172$ & $3214\pm83$ & $\mathbf{5.21\pm0.06}$ & $16769\pm593$ & $978$ \\
& PPGA & $6866\pm513$ & $2608\pm549$ & $4.26\pm0.43$ & $10426\pm1283$ & $1196$ \\
& PGA-ME & $3255\pm156$ & $1391\pm71$ & $3.63\pm0.14$ & $5047\pm343$ & $1156$ \\
& Sep-CMA-MAE & $3365\pm122$ & $1653\pm51$ & $4.10\pm0.08$ & $6792\pm309$ & $925$ \\
& Ind-PPO & $7624\pm276$ & $\mathbf{6191\pm83}$ & $1.15\pm0.02$ & $7143\pm161$ & $32$ \\
\bottomrule
\end{tabular}
\end{table}

\begin{figure}[ht]
    \centering
    \includegraphics[width=\textwidth]{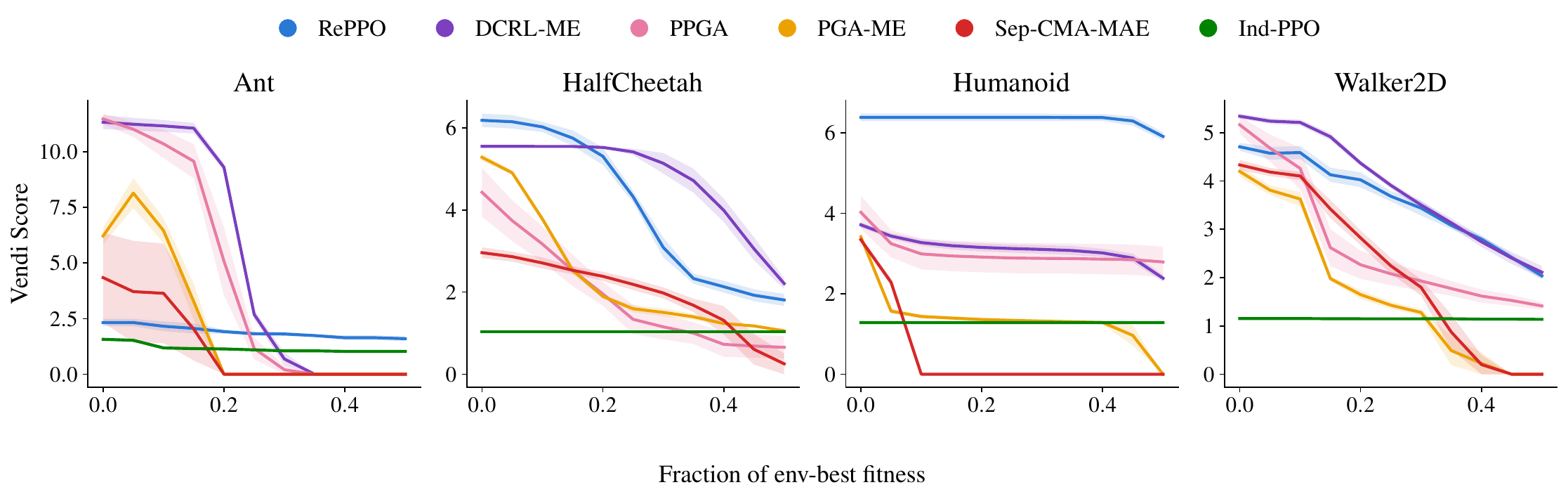}
    \includegraphics[width=\textwidth]{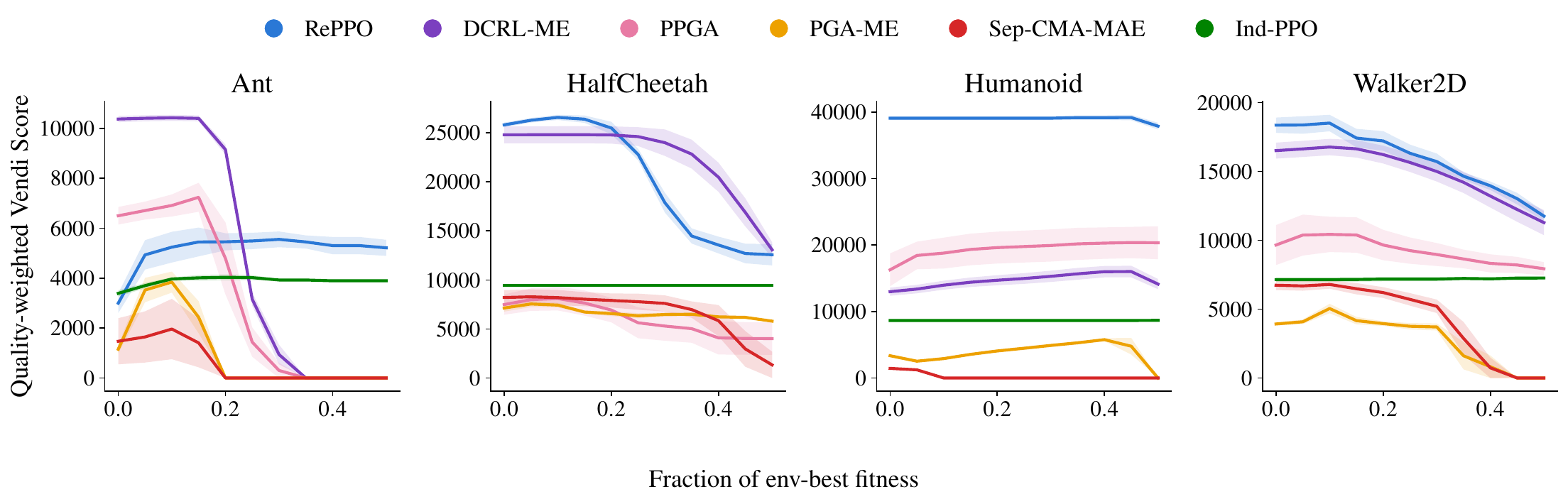}
    \caption{
    Vendi Score (top) and qVS (bottom) of the complete populations as a function of the minimum fitness, given as a fraction of the best policy found on the task.
    Lines show the mean over seeds and shaded regions one standard error.
    }
    \label{app:fig:cc-profiles}
\end{figure}

\paragraph{Effect of the quality filter.}
\label{app:cc-filter}
Diversity metrics can be dominated by policies that behave differently but do not solve the task.
Figure~\ref{app:fig:cc-profiles} therefore recomputes the Vendi Score and qVS of the complete populations and archives while the required fraction of $f^\star$ increases from $0$ to $0.5$.
With no filter, the archives of DCRL-ME and PPGA are far more diverse than RePPO's population on Ant, but their Vendi Scores collapse as the threshold is increased.
Upon manual inspection, we noticed that many of the policies in archive-based methods stand nearly still and occupy different cells by holding different subsets of legs off the ground, which prompted our choice of a $25\%$ threshold for this task.
Beyond this threshold, RePPO attains the highest qVS.

\begin{figure}[ht]
    \centering
    \includegraphics[width=\textwidth]{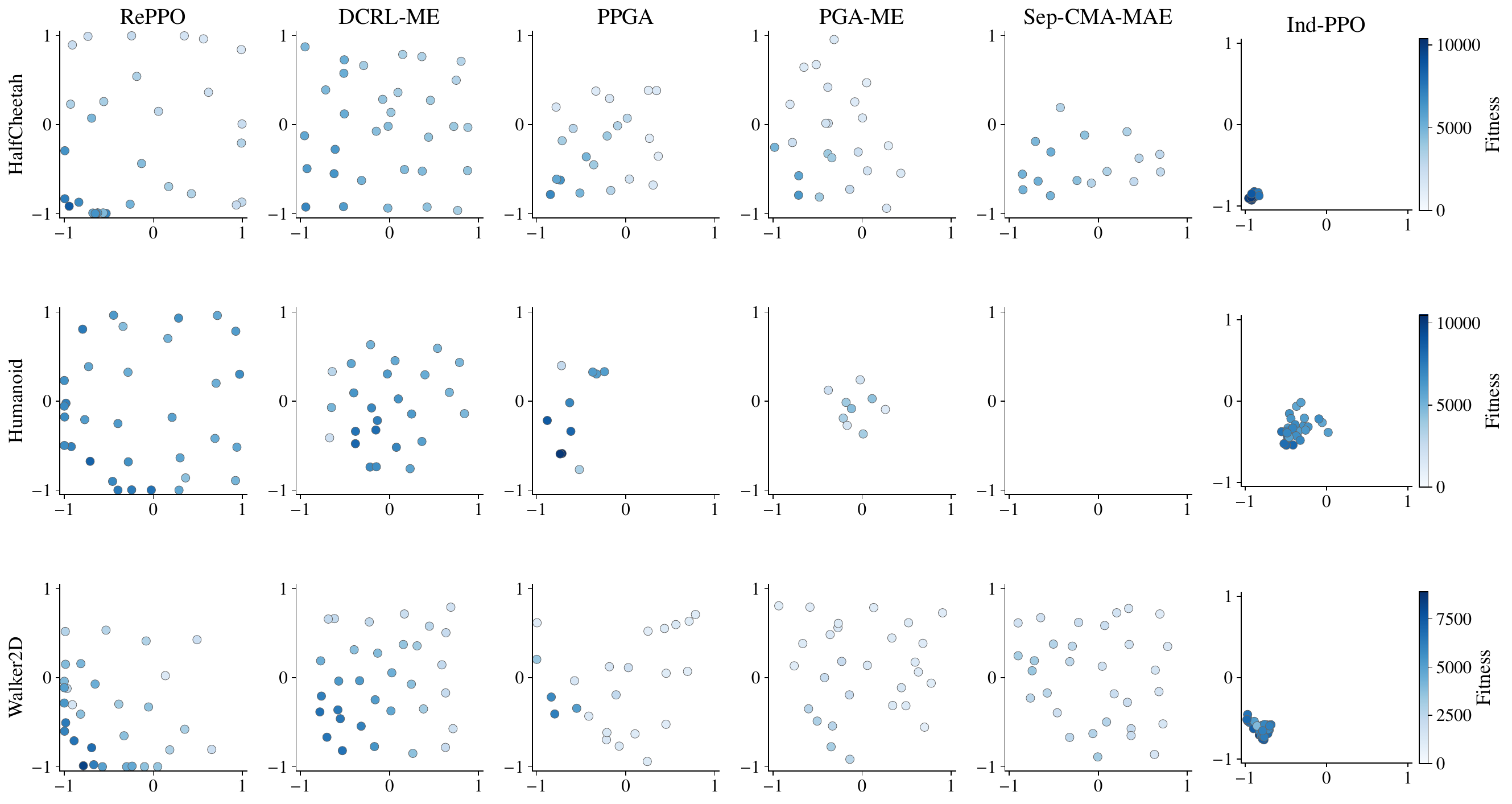}
    \caption{
    Retained policies of one seed per method in behavior space, colored by fitness on a shared scale within each task.
    For readability, archives are thinned to one policy in each of $32$ CVT cells, keeping the fittest.
    }
    \label{app:fig:cc-descriptors}
\end{figure}

\begin{figure}[ht]
    \centering
    \includegraphics[width=\textwidth]{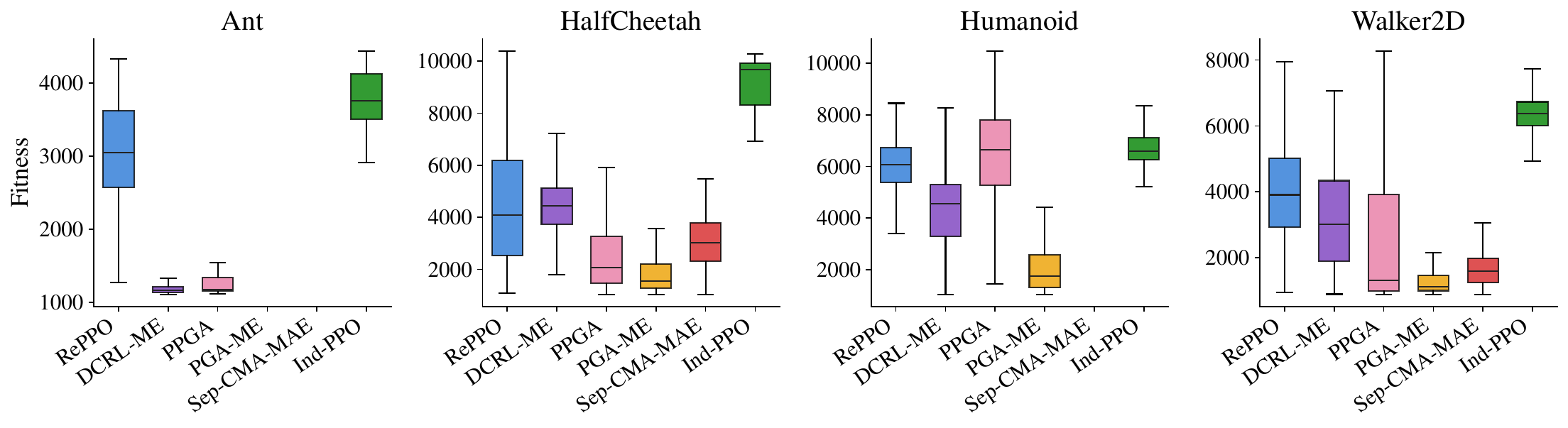}
    \caption{
    Fitness of the retained policies, pooled over seeds.
    Boxes show the interquartile range and medians.
    }
    \label{app:fig:cc-fitness}
\end{figure}

\paragraph{Populations in behavior space.}
Figure~\ref{app:fig:cc-descriptors} shows the retained policies of one seed per method in tasks with two-dimensional behavior spaces, and Figure~\ref{app:fig:cc-fitness} shows the distribution of their fitness.
Archive-based methods cover behavior space more uniformly due to their structure, but much of that coverage comes from low-fitness policies.
RePPO places its policies less uniformly and concentrates them in regions that support fast locomotion, as the Soft QD objective rewards illuminating behavior space with high-quality policies rather than covering it uniformly.
Independent PPO concentrates all the policies in a small, high-quality region.

\begin{figure}[p]
\centering
\includegraphics[width=\textwidth]{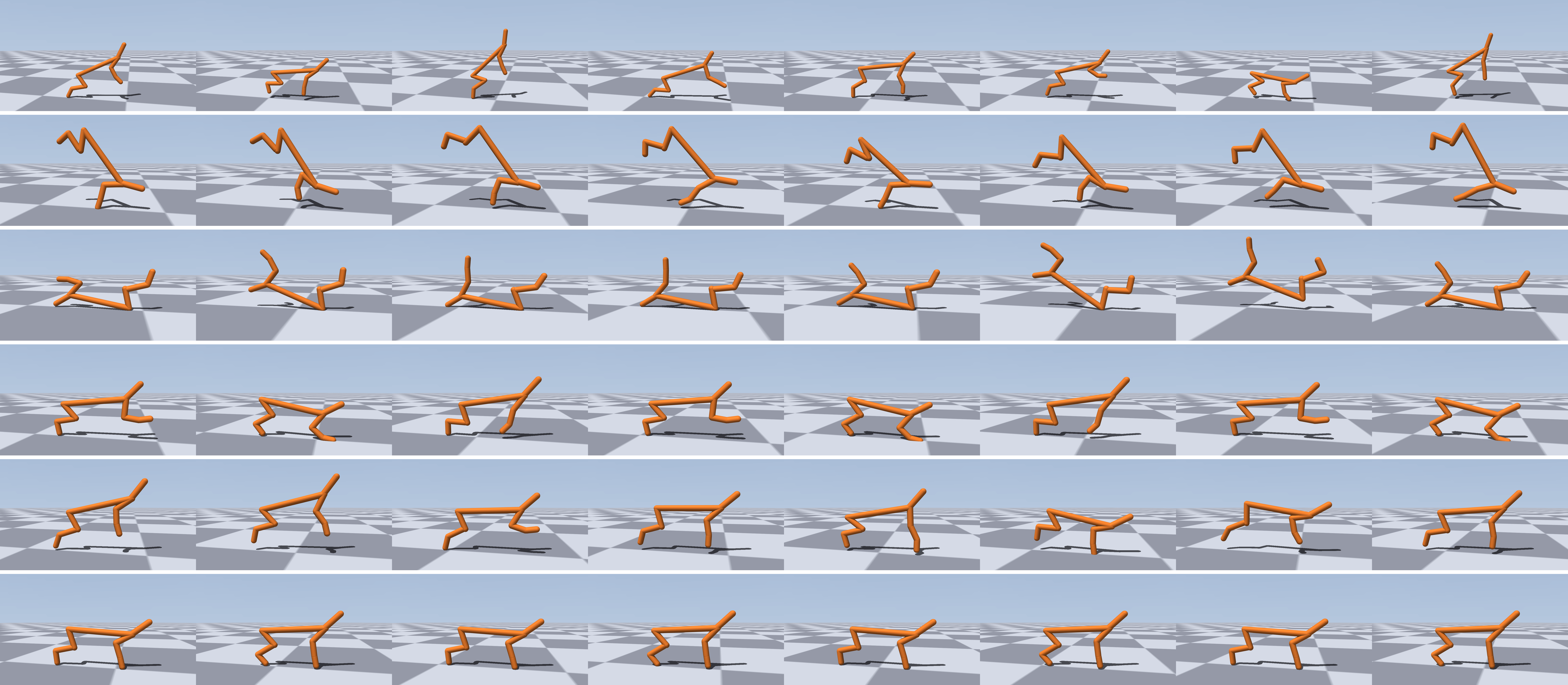}
\caption{Gaits found by one RePPO population in HalfCheetah.}
\label{app:fig:halfcheetah-behavior}
\end{figure}

\begin{figure}[p]
\centering
\includegraphics[width=\textwidth]{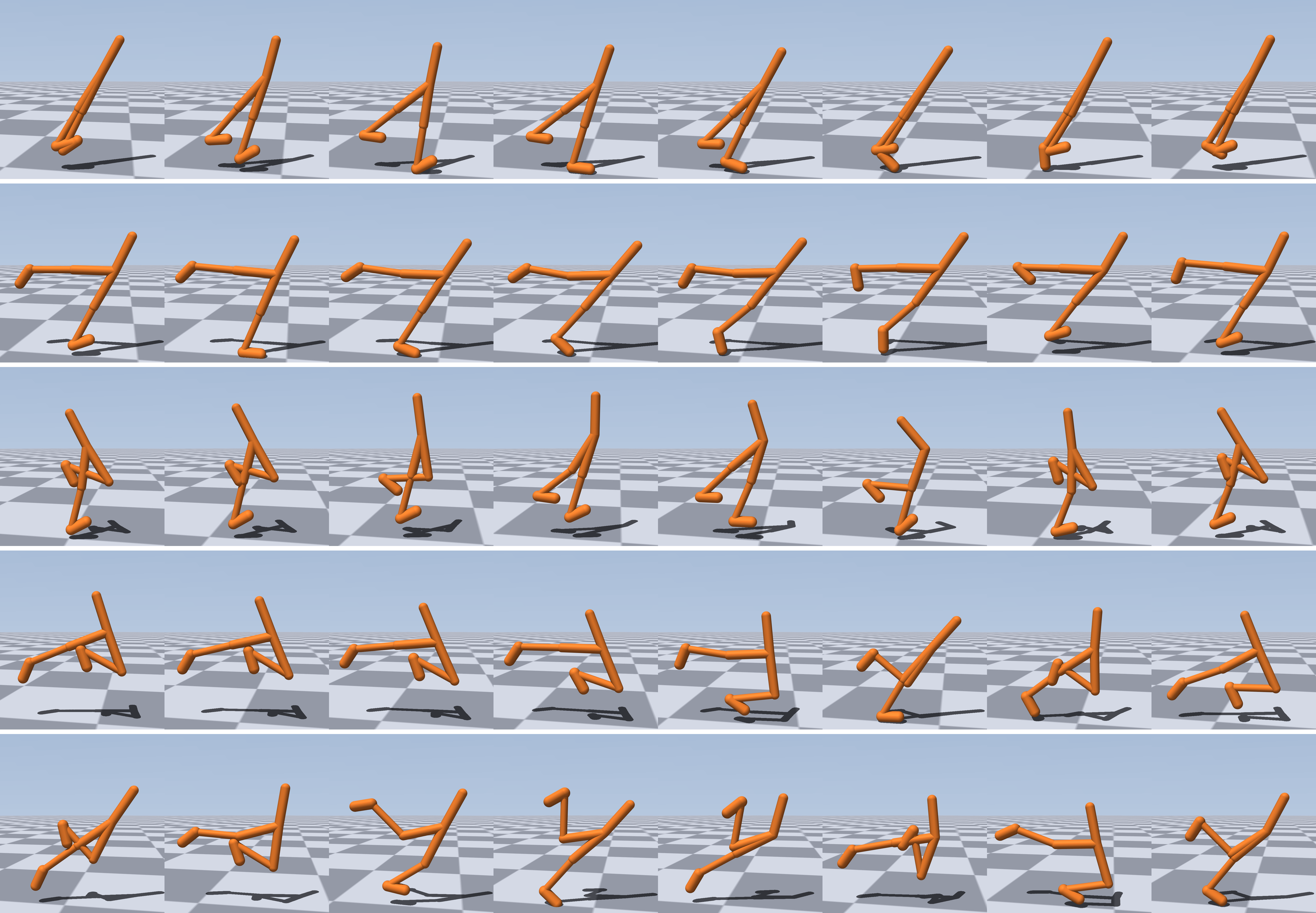}
\caption{Gaits found by one RePPO population in Walker2D.}
\label{app:fig:walker-behavior}
\end{figure}

\begin{figure}[p]
\centering
\includegraphics[width=\textwidth]{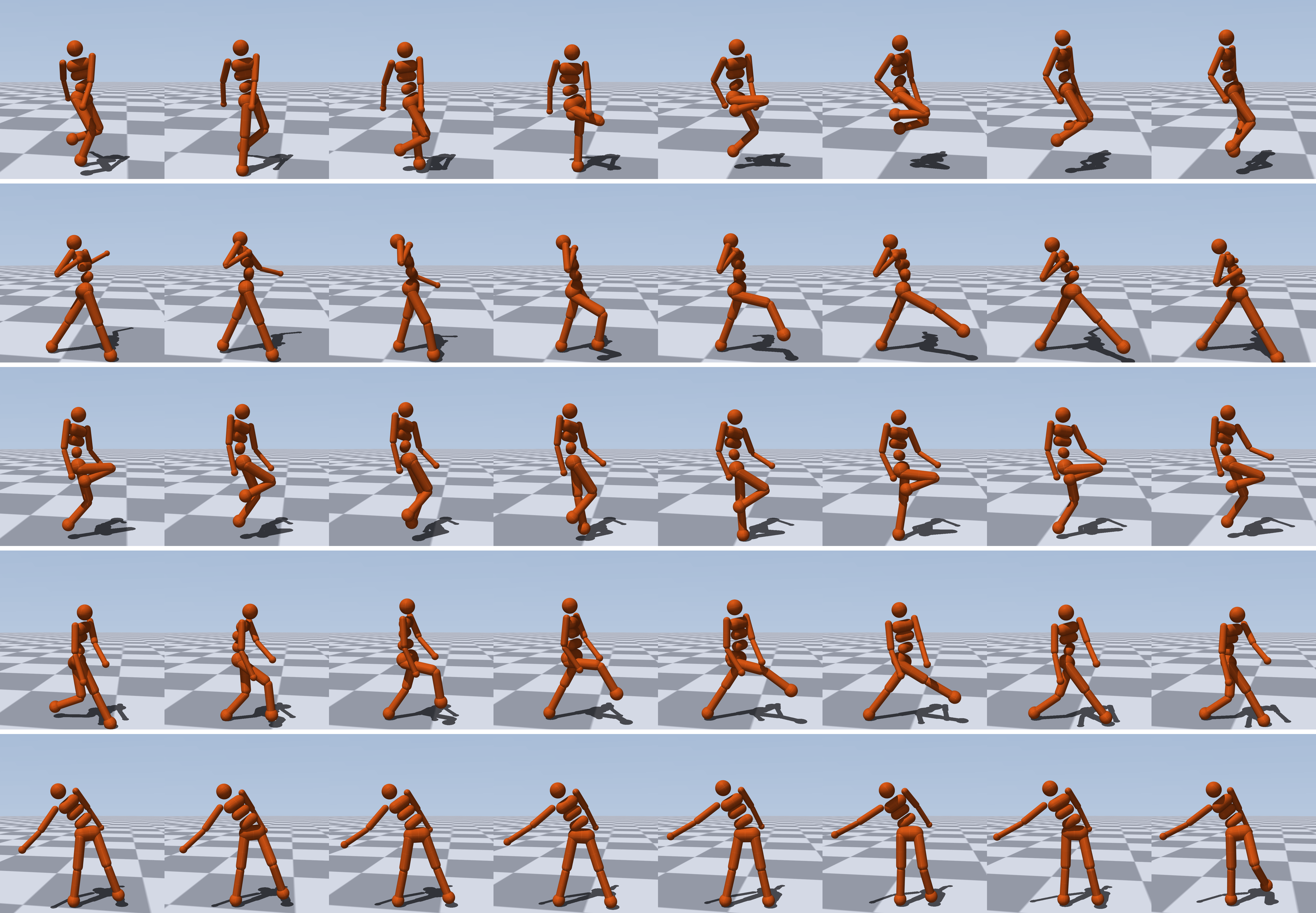}
\caption{Gaits found by one RePPO population in Humanoid.}
\label{app:fig:humanoid-behavior}
\end{figure}

\paragraph{Behaviors.}
Figures~\ref{app:fig:halfcheetah-behavior}--\ref{app:fig:humanoid-behavior} show gaits found by a single RePPO population in each task, which differ in which feet carry the body and how often they touch the ground.

\FloatBarrier
\section{Ablations}
\label{app:ablations}

We ablate RePPO's components and hyperparameters on Humanoid.
Each variant changes one setting of the main Humanoid configuration ($C_{\max}=1$, $\ell^2=0.1$, $K=8$, $N=32$, symmetric repulsion, $W_k=1$) and runs for three seeds.
The default is represented by the first three seeds of the main runs.
All metrics follow Appendix~\ref{app:evaluation}, with the quality threshold of the main comparison.
We additionally report the spread of a population, the mean Euclidean distance of its policies from their centroid in the normalized behavior space $[-1,1]^d$, which characterizes its geometry alongside the Vendi Score.
Table~\ref{app:tab:ablations} lists every variant.

\FloatBarrier
\subsection{Structural choices}
\label{app:structural-ablations}

We first examine two structural choices: the unit task coefficient of Section~\ref{main:sec:method-derivation}, and symmetric interactions, in which every pair of neighbors repels each other regardless of their qualities.

\paragraph{Exact task coefficient.}
Using the exact coefficient $W_k$ of~\eqref{main:eq:wk} causes training to collapse in every seed.
Policies perform significantly worse with the best policy reaching a fitness of only $282\pm230$.
When several high-quality policies share a region of behavior space--which happens often early in training--the second term of $W_k$ outweighs the first and the task coefficient becomes negative.
The policy is then rewarded for lowering its return, and once it reaches low-return behaviors, such as falling immediately, PPO does not recover.
Fixing $W_k=1$ preserves a consistent task-improvement signal and leaves behavioral separation to the descriptor term.

\paragraph{Asymmetric repulsion.}
In this variant, a policy is repelled only by neighbors of higher quality, so that the best policies are never displaced by weaker ones.
The best policy is unaffected ($8409\pm201$ versus $8336\pm68$), but the rest of the population is both less fit and less diverse, lowering qVS from $39212$ to $34086$.
Symmetric interactions let every overlapping pair separate, which spreads the population more evenly.

\FloatBarrier
\subsection{Interaction strength and scale}
\label{app:population-geometry-ablations}

\begin{figure}[ht]
\centering
\includegraphics[width=0.49\textwidth]{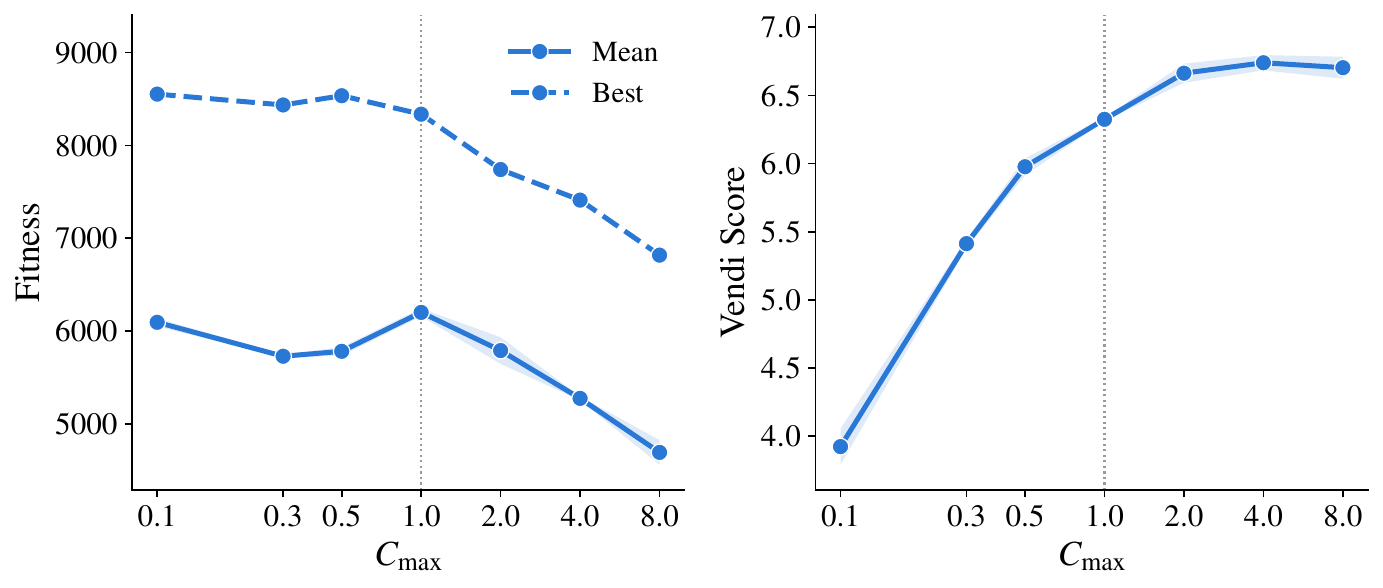}
\includegraphics[width=0.49\textwidth]{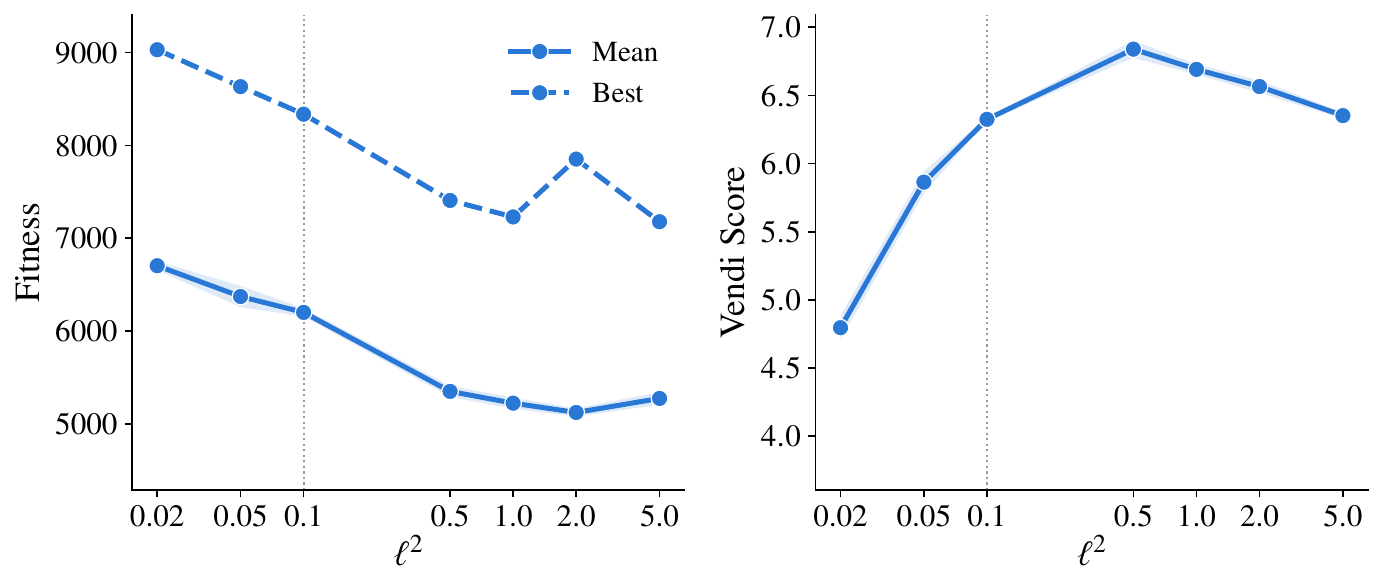}
\caption{Effect of the repulsion cap $C_{\max}$ (left) and the kernel bandwidth $\ell^2$ (right) on Humanoid. Shaded regions show one standard error over seeds.}
\label{app:fig:cmax-gamma}
\end{figure}

\begin{figure}[ht]
\centering
\includegraphics[width=\textwidth]{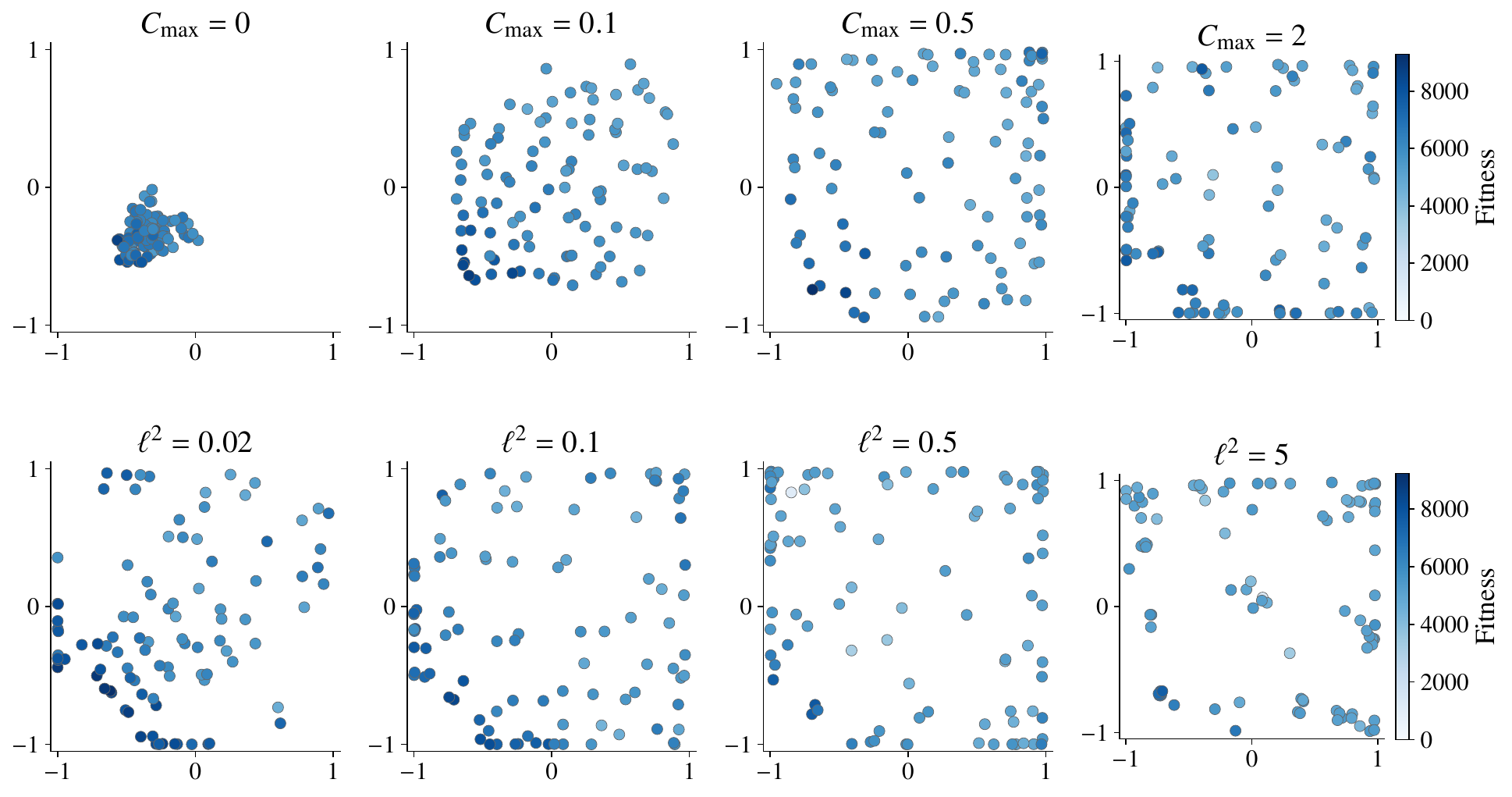}
\caption{Final Humanoid populations for representative values of $C_{\max}$ (top) and $\ell^2$ (bottom), pooled over seeds and colored by fitness. $C_{\max}=0$ corresponds to Ind-PPO.}
\label{app:fig:population-shapes}
\end{figure}

Together, the repulsion cap $C_{\max}$ and the kernel bandwidth $\ell^2$ shape how the population spreads through behavior space, but they do so in different ways (Figures~\ref{app:fig:cmax-gamma} and~\ref{app:fig:population-shapes}).

\paragraph{Repulsion cap.}
The cap controls how strongly diversity enters each update, with $C_{\max}=0$ recovering Ind-PPO.
Raising it from $0.1$ to $1$ increases the Vendi Score from $3.92$ to $6.33$ and the spread from $0.60$ to $0.91$, while mean fitness stays between $5700$ and $6200$.
Beyond $C_{\max}=2$, diversity saturates and the stronger diversity signal begins to compete with task learning: at $C_{\max}=8$, mean and best fitness fall to $4691$ and $6817$.
qVS peaks at the default of $1$, with $C_{\max}=2$ close behind.
The cap therefore mainly controls how far the population expands, and needs to be large enough to spread the policies but not so large that repulsion dominates the update.

\paragraph{Kernel bandwidth.}
The bandwidth sets the scale over which policies interact and thus the configuration the population settles into.
With $\ell^2=0.02$, interactions are highly local, and the population remains compact (spread $0.70$, Vendi Score $4.80$) and fit (mean fitness $6703$).
Increasing the bandwidth pushes policies apart until the spread saturates near $1.0$ at $\ell^2=0.5$, where the Vendi Score peaks at $6.84$.
Larger bandwidths push policies toward the corners of behavior space, leaving its center empty, so that diversity declines slightly while mean fitness stays near $5200$.
qVS peaks at $\ell^2=0.1$, which balances these effects on Humanoid.

\FloatBarrier
\subsection{Local interactions}
\label{app:neighbor-ablation}

Restricting the interactions to $K$ nearest neighbors has little effect over the range we tested (Figure~\ref{app:fig:knn-popsize}, left).
With $K\in\{4,8,16\}$, the Vendi Score lies between $6.33$ and $6.50$ and qVS within $4\%$ of the default.
Interacting with the entire population ($K=31$) produces slightly fitter but less diverse populations.
We therefore conclude that the nearest neighbors carry most of the interaction, and $K=8$ retains its effect at lower cost.

\FloatBarrier
\subsection{Population size at a fixed budget}
\label{app:population-size-ablation}

\begin{figure}[ht]
\centering
\includegraphics[width=0.49\textwidth]{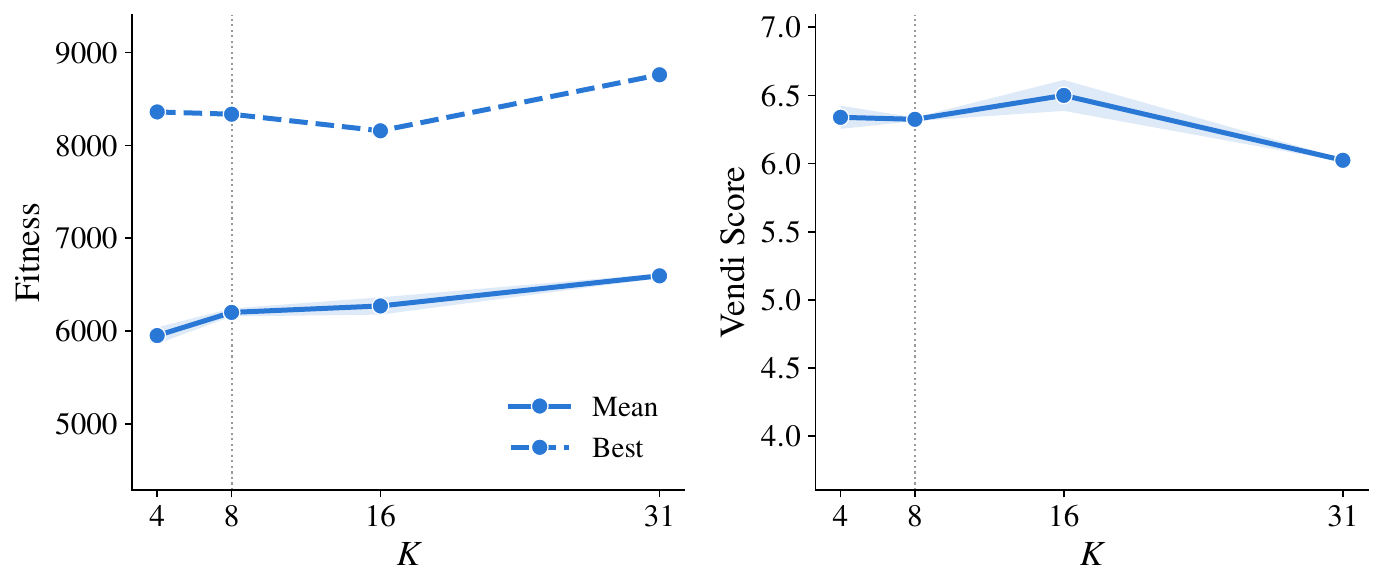}
\includegraphics[width=0.49\textwidth]{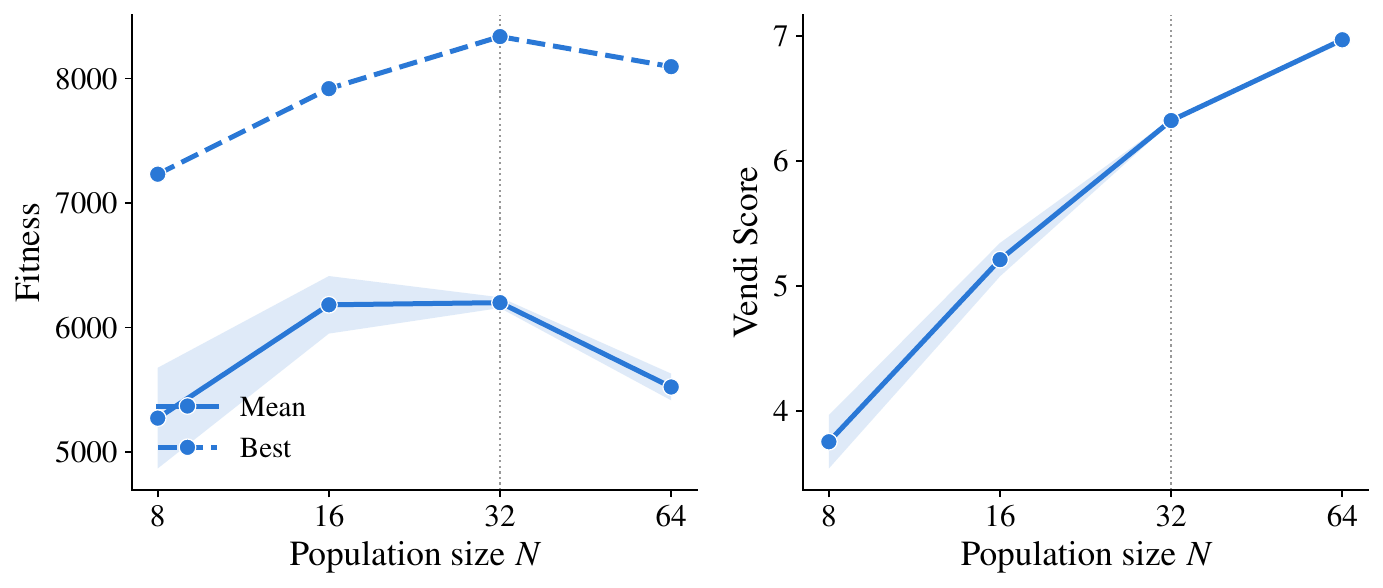}
\caption{Effect of the number of neighbors $K$ (left) and of the population size $N$ at a fixed budget (right) on Humanoid. Shaded regions show one standard error over seeds.}
\label{app:fig:knn-popsize}
\end{figure}

We vary the population size while keeping the total number of environment steps fixed, so that larger populations give each policy less experience (Figure~\ref{app:fig:knn-popsize}, right).
Diversity grows with population size with diminishing returns, from a Vendi Score of $3.76$ at $N=8$ to $6.97$ at $N=64$.
Mean fitness is highest for $N=16$ and $N=32$ and drops at $N=64$, where each policy receives half the experience of the default, whereas the best fitness remains above $7200$ for every size.
A larger population therefore represents behavior space more densely until the budget is spread too thinly, and $N=32$ balances these effects.

\begin{table}[ht]
\centering
\caption{All Humanoid ablations, as mean $\pm$ standard error over three seeds. The default configuration appears in bold.}
\label{app:tab:ablations}
\small
\setlength{\tabcolsep}{5pt}
\begin{tabular}{@{}llccccc@{}}
\toprule
& Setting & Mean fitness & Best fitness & Vendi Score & qVS & Spread \\
\midrule
& \textbf{Default} & $\mathbf{6200\pm43}$ & $\mathbf{8336\pm68}$ & $\mathbf{6.33\pm0.01}$ & $\mathbf{39212\pm231}$ & $\mathbf{0.91}$ \\
\midrule
\multirow{2}{*}{Structure}
& Asymmetric repulsion & $5935\pm144$ & $8409\pm201$ & $5.74\pm0.21$ & $34086\pm1688$ & $0.83$ \\
& Exact $W_k$ & $0\pm0$ & $282\pm230$ & $0.00\pm0.00$ & $0\pm0$ & -- \\
\midrule
\multirow{6}{*}{$C_{\max}$}
& $0.1$ & $6093\pm42$ & $8553\pm138$ & $3.92\pm0.13$ & $23901\pm681$ & $0.60$ \\
& $0.3$ & $5727\pm18$ & $8436\pm195$ & $5.41\pm0.03$ & $31003\pm282$ & $0.79$ \\
& $0.5$ & $5780\pm47$ & $8536\pm443$ & $5.98\pm0.06$ & $34544\pm88$ & $0.87$ \\
& $2.0$ & $5788\pm143$ & $7740\pm67$ & $6.66\pm0.07$ & $38581\pm1225$ & $0.94$ \\
& $4.0$ & $5273\pm22$ & $7410\pm75$ & $6.74\pm0.06$ & $35542\pm444$ & $0.95$ \\
& $8.0$ & $4691\pm132$ & $6817\pm51$ & $6.70\pm0.08$ & $31466\pm1253$ & $0.96$ \\
\midrule
\multirow{6}{*}{$\ell^2$}
& $0.02$ & $6703\pm48$ & $9031\pm100$ & $4.80\pm0.09$ & $32149\pm551$ & $0.70$ \\
& $0.05$ & $6371\pm114$ & $8633\pm276$ & $5.86\pm0.08$ & $37367\pm952$ & $0.83$ \\
& $0.5$ & $5348\pm60$ & $7407\pm338$ & $6.84\pm0.06$ & $36586\pm712$ & $1.03$ \\
& $1.0$ & $5221\pm59$ & $7229\pm388$ & $6.69\pm0.04$ & $34943\pm583$ & $1.00$ \\
& $2.0$ & $5121\pm45$ & $7852\pm200$ & $6.57\pm0.05$ & $33626\pm509$ & $1.01$ \\
& $5.0$ & $5272\pm67$ & $7177\pm363$ & $6.35\pm0.02$ & $33493\pm539$ & $0.97$ \\
\midrule
\multirow{3}{*}{$K$}
& $4$ & $5950\pm89$ & $8361\pm203$ & $6.34\pm0.08$ & $37724\pm791$ & $0.91$ \\
& $16$ & $6269\pm93$ & $8157\pm92$ & $6.50\pm0.11$ & $40734\pm333$ & $0.92$ \\
& $31$ & $6594\pm24$ & $8761\pm196$ & $6.02\pm0.01$ & $39725\pm128$ & $0.90$ \\
\midrule
\multirow{3}{*}{$N$}
& $8$ & $5272\pm404$ & $7231\pm153$ & $3.76\pm0.21$ & $19820\pm1979$ & $0.65$ \\
& $16$ & $6182\pm231$ & $7918\pm512$ & $5.21\pm0.13$ & $32183\pm886$ & $0.79$ \\
& $64$ & $5522\pm107$ & $8095\pm227$ & $6.97\pm0.03$ & $38492\pm619$ & $1.00$ \\
\bottomrule
\end{tabular}
\end{table}

\FloatBarrier
\section{Atari}
\label{app:atari}

\FloatBarrier
\subsection{Games and descriptors}
\label{app:atari-setup}

We use the ALE~v5 versions~\citep{bellmare2015arcade} of Boxing, Asterix, Seaquest, and Ms.\ Pac-Man through EnvPool~\citep{weng2022envpool}, with the standard preprocessing: a frame skip of $4$ with max-pooling over the last two frames, $0$ to $30$ no-op actions at reset, grayscale frames resized to $84\times84$, and a stack of the last $4$ frames.
Following \citet{machado2018revisiting}, the emulator repeats the previous action with probability $0.25$ during both training and evaluation.
A game lasts until all lives are lost or $27000$ agent steps have passed.
Fitness is the raw game score of a full game.
During training, the task reward is clipped to its sign and a lost life ends the episode for the task critic.

\begin{figure}[ht]
    \centering
    \includegraphics[width=\textwidth]{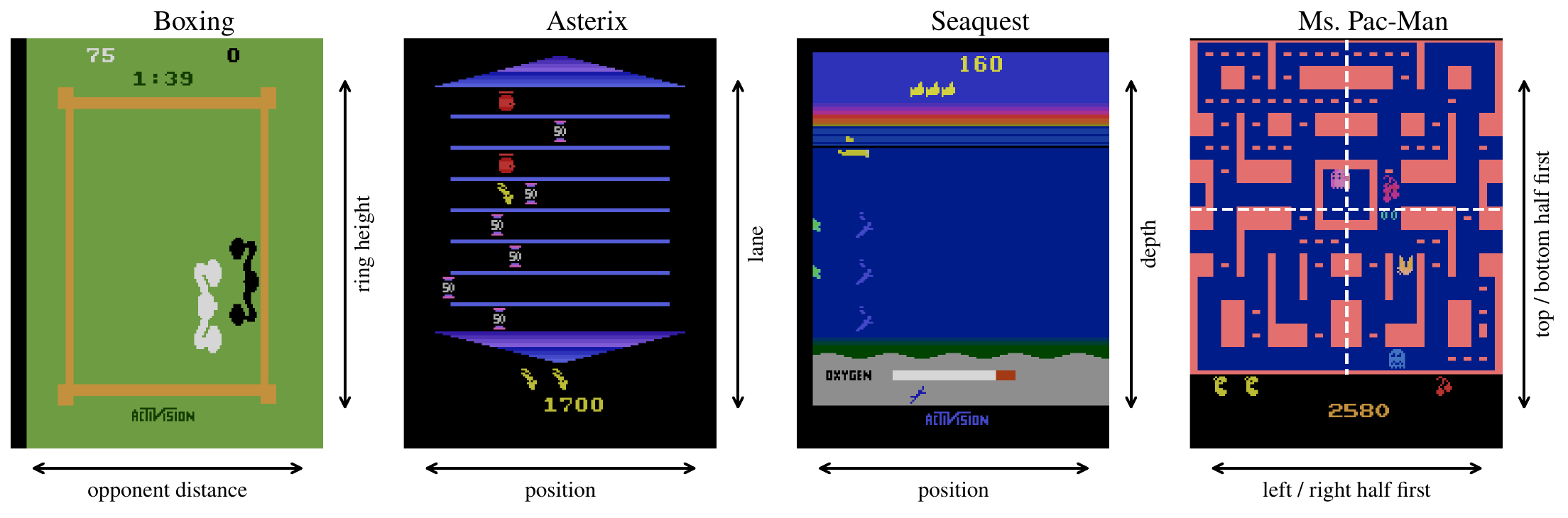}
    \caption{The four Atari games, as played by the best RePPO policy of one population, with the two behavior dimensions of each game indicated by arrows.}
    \label{app:fig:atari-games}
\end{figure}

Each game defines a two-dimensional behavior signal $\mathbf{m}_t$ from the console memory, and the descriptor of a game is its average over all agent steps, clipped to $[-1,1]$.
Table~\ref{app:tab:atari-descriptors} defines the signals and Figure~\ref{app:fig:atari-games} illustrates the games.
For the three positional descriptors, the affine scales map the range that competent play reaches to roughly $[-1,1]$.
In Ms.\ Pac-Man, eating a dot after $k$ of the level's $n$ dots have been eaten contributes $w_k=1-2k/(n-1)$, from $+1$ for the first dot to $-1$ for the last, signed by the half of the maze in which the dot lies, so that the descriptor records which halves are cleared first and is zero when the maze is cleared uniformly.

\begin{table}[ht]
\centering
\caption{Behavior signals in Atari. Positional signals are affine in the memory values and clipped to $[-1,1]$ per step.}
\label{app:tab:atari-descriptors}
\small
\begin{tabular}{@{}lll@{}}
\toprule
Game & First dimension ($-1\to+1$) & Second dimension ($-1\to+1$) \\
\midrule
Boxing & distance to the opponent (close $\to$ far) & height in the ring (top $\to$ bottom) \\
Asterix & horizontal position (left $\to$ right) & lane (top $\to$ bottom) \\
Seaquest & horizontal position (left $\to$ right) & depth (just below the surface $\to$ sea floor) \\
Ms.\ Pac-Man & clearing order (left half first $\to$ right half first) & clearing order (top half first $\to$ bottom half first) \\
\bottomrule
\end{tabular}
\end{table}

\FloatBarrier
\subsection{RePPO and Ind-PPO}
\label{app:atari-hparams}

Each policy is a Nature CNN~\citep{mnih2015human} with linear actor, task-critic, and two-output behavior-critic heads on its $512$-dimensional features, totaling $1.69$~M parameters.
The policies share no parameters.
PPO settings follow the standard Atari configuration~\citep{huang2022cleanrl}, and Table~\ref{app:tab:atari-hparams} lists all hyperparameters.
Two settings adapt the behavior channels to full games of several thousand steps.
First, the behavior channels end only when the game ends rather than at every lost life.
Second, their rewards are scaled by $1-\gamma$ so that the behavior returns stay on the scale of the descriptor itself.
The EMA rates are lower than in continuous control because a game spans many iterations, and the statistics are updated only in iterations in which a game ends.

\begin{table}[ht]
\centering
\caption{RePPO hyperparameters in Atari. Ind-PPO uses the same settings with the diversity weights set to zero.}
\label{app:tab:atari-hparams}
\small
\begin{tabular}{@{}ll@{\hspace{2.5em}}ll@{}}
\toprule
Parameter & Value & Parameter & Value \\
\midrule
Population size $N$ & $8$ & Discount $\gamma$ / GAE $\lambda$ & $0.99$ / $0.95$ \\
Environments per policy & $8$ & PPO clip & $0.1$, with value clipping \\
Rollout length & $128$ & Entropy / value coefficient & $0.01$ / $0.5$ \\
Steps per policy & $10$~M & Max gradient norm & $0.5$ \\
Learning rate & $2.5\times10^{-4}$, linear decay to $0$ & Neighbors $K$ / repulsion cap $C_{\max}$ & $7$ / $1.0$ \\
Update epochs / minibatches & $4$ / $4$ & Kernel bandwidth $\ell^2$ & $0.1$ \\
EMA rates & $0.9$ & $\epsilon_f$ / $\epsilon_b$ & $1$ / $10^{-6}$ \\
\bottomrule
\end{tabular}
\end{table}

\FloatBarrier
\subsection{RefQD}
\label{app:atari-refqd}

RefQD~\citep{wang2024quality} reduces the cost of an archive of neural policies by splitting each network into a representation part, shared by the entire archive, and a decision part, the last layer, which is stored per cell.
We use its Atari variant, Ref-DQN-ME, which produces offspring by isoline variation and by Double DQN updates~\citep{vanhasselt2016deep}, with the authors' default hyperparameters: a CVT archive of $1000$ cells that stores two decision parts per cell, $100$ offspring per generation, $300$ DQN training steps per generation, periodic re-evaluation of archived decision parts with the current representation, and a single grayscale frame as input.
We set the budget to $80$~M steps, which the implementation rounds up to $86.4$~M so that every re-evaluation cycle completes.
For Seaquest, we enlarge the replay buffer from $2\times10^5$ to $3\times10^5$ transitions due to longer episodes.

\FloatBarrier
\subsection{Additional results}
\label{app:atari-results}

Table~\ref{app:tab:atari-full} reports the numbers behind the bottom row of Figure~\ref{main:fig:qd}.
In Boxing, Ind-PPO's policies keep the same distance from the opponent and mostly stay at one of two heights, whereas RePPO's policies spread over the height of the ring and include a second family that fights at closer range.
In Asterix and Ms.\ Pac-Man, Ind-PPO's policies already vary along one dimension, horizontal position and left--right clearing order, which the symmetry of the games makes equally rewarding.
RePPO additionally spreads its policies along the other dimension, covering the outer lanes in Asterix and including policies that clear the top half of the maze first in Ms.\ Pac-Man.

\begin{table}[ht]
\centering
\caption{Final populations in Atari, as mean $\pm$ standard error over five seeds.}
\label{app:tab:atari-full}
\small
\setlength{\tabcolsep}{5pt}
\begin{tabular}{@{}llcccc@{}}
\toprule
Game & Method & Best score & Mean score & Vendi Score & qVS \\
\midrule
\multirow{3}{*}{Boxing}
& RePPO & $\mathbf{98\pm0}$ & $91\pm1$ & $\mathbf{2.96\pm0.09}$ & $\mathbf{271\pm8}$ \\
& RefQD & $33\pm5$ & $17\pm2$ & $1.88\pm0.12$ & $32\pm4$ \\
& Ind-PPO & $96\pm1$ & $\mathbf{93\pm0}$ & $1.72\pm0.18$ & $159\pm16$ \\
\midrule
\multirow{3}{*}{Asterix}
& RePPO & $\mathbf{3426\pm119}$ & $2939\pm28$ & $\mathbf{4.01\pm0.03}$ & $\mathbf{11783\pm129}$ \\
& RefQD & $1385\pm23$ & $596\pm10$ & $3.09\pm0.13$ & $1840\pm68$ \\
& Ind-PPO & $3352\pm102$ & $\mathbf{2945\pm41}$ & $2.46\pm0.11$ & $7252\pm342$ \\
\midrule
\multirow{3}{*}{Seaquest}
& RePPO & $\mathbf{3382\pm465}$ & $\mathbf{1855\pm125}$ & $\mathbf{4.33\pm0.28}$ & $\mathbf{8100\pm830}$ \\
& RefQD & $511\pm19$ & $485\pm5$ & $1.11\pm0.05$ & $538\pm26$ \\
& Ind-PPO & $1907\pm93$ & $1469\pm90$ & $1.50\pm0.10$ & $2173\pm76$ \\
\midrule
\multirow{3}{*}{Ms.\ Pac-Man}
& RePPO & $2588\pm112$ & $2088\pm36$ & $3.38\pm0.14$ & $\mathbf{7061\pm369}$ \\
& RefQD & $1176\pm105$ & $474\pm12$ & $\mathbf{3.65\pm0.03}$ & $1732\pm42$ \\
& Ind-PPO & $\mathbf{2730\pm132}$ & $\mathbf{2215\pm59}$ & $2.30\pm0.09$ & $5107\pm255$ \\
\bottomrule
\end{tabular}
\end{table}

\begin{figure}[p]
    \centering
    \includegraphics[width=\textwidth]{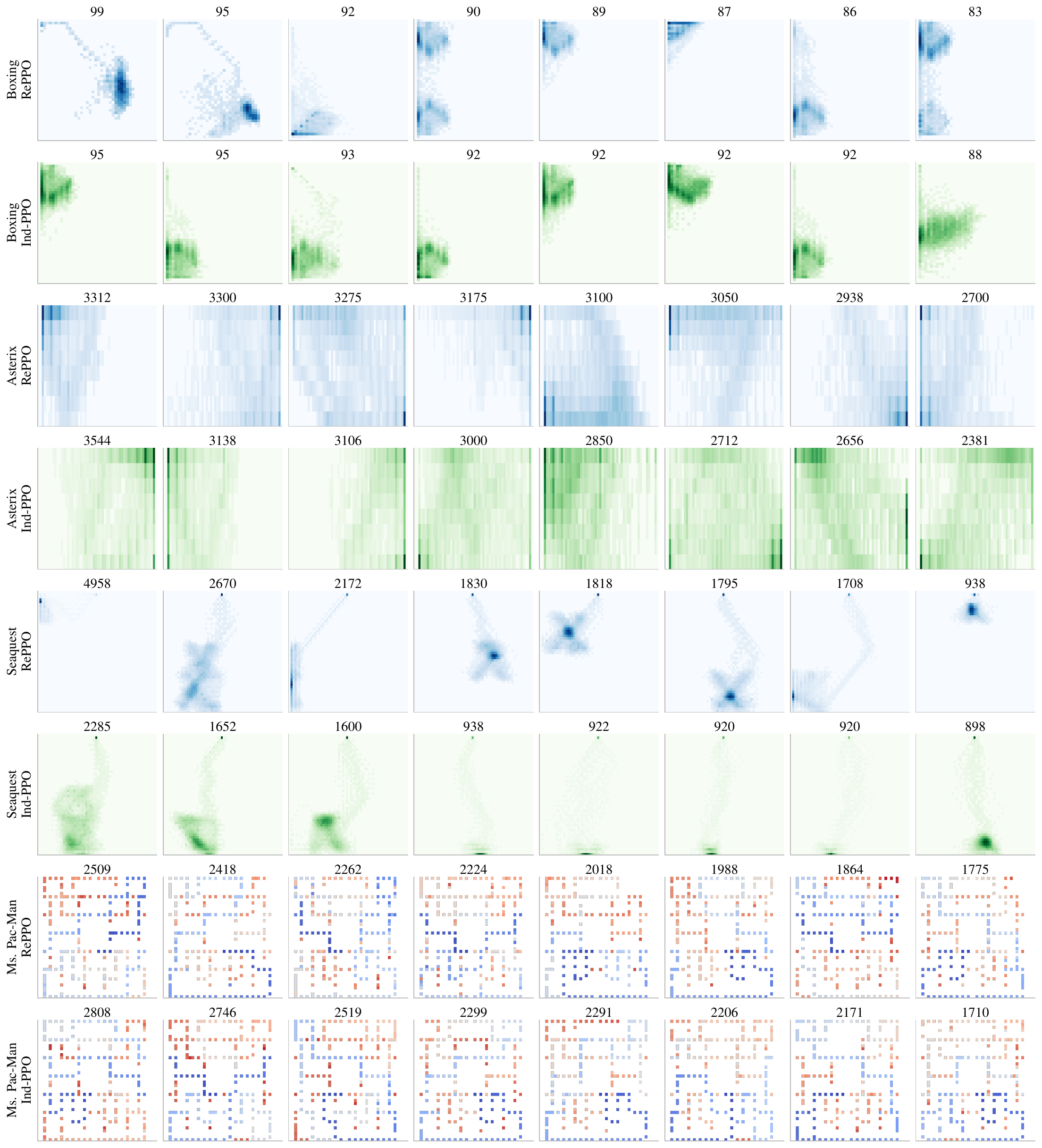}
    \caption{
    Behavior of the eight policies of the first seed of RePPO (blue) and Ind-PPO (green), sorted by score, which is shown above each panel.
    For Boxing, Asterix, and Seaquest, each panel is a heatmap of the player's position on the screen.
    For Ms.\ Pac-Man, each panel colors the dots of the maze by when they are eaten, from early (blue) to late (red), as a fraction of the level.
    }
    \label{app:fig:atari-occupancy}
\end{figure}

Figure~\ref{app:fig:atari-occupancy} shows where each policy of the first seed plays, over eight games per policy.
In Boxing, most Ind-PPO policies fight from the left side of the ring at one of two heights, whereas RePPO's policies also fight from the right side of the ring and along its top and bottom edges.
In Seaquest, Ind-PPO's policies descend to the sea floor and stay there, while RePPO's policies occupy different depths and horizontal positions, and its best policy remains just below the surface.
In Ms.\ Pac-Man, all policies traverse the whole maze, so the figure instead shows when each dot is eaten.
The policies of both methods start from the bottom of the maze, where the game begins, and differ in which regions they leave for last.

\FloatBarrier
\section{Fine-tuning large language models}
\label{app:llm}

\FloatBarrier
\subsection{Task and data}
\label{app:llm-setup}

We train on the $5584$ problems of levels $3$--$5$ of the MATH training set~\citep{hendrycks2021math} whose reference solution contains a boxed answer, as in SimpleRL-Zoo~\citep{zeng2025simplerlzoo}.
For checkpoint selection, we sample $384$ problems of the same levels from the MATH test set, excluding MATH-500.
The test set combines MATH-500~\citep{lightman2024verify} ($500$ problems), Minerva~\citep{lewkowycz2022minerva} ($272$), and the English text-only part of OlympiadBench~\citep{he2024olympiadbench} ($675$).
The base models are Qwen3-1.7B-Base~\citep{qwen3} and Granite-4.1-3B-Base~\citep{granite}.
Every problem is accompanied by the instruction ``Please reason step by step, and put your final answer within \texttt{\textbackslash boxed\{\}}.'', given as the system message of the chat template for Qwen3 and after the problem in a plain question--answer format for Granite.

\paragraph{Fitness.}
A solution scores $r=1$ if it ends within $4096$ tokens and its last boxed expression matches the reference under Math-Verify, and $r=0$ otherwise.
The training reward standardizes this score as $(r-\mu_0)/\sigma_0$, using its mean and standard deviation on the base model's first batch.

\paragraph{Descriptors.}
Four raw features $x_j$ are computed on the solution text up to its last boxed answer (Table~\ref{app:tab:llm-descriptors}).
The descriptor corresponding to a solution is then $d_j=r\tanh\big((x_j-\mu_j)/\sigma_j\big)\in[-1,1]$, where $\mu_j$ is the mean of $x_j$ over the base model's correct solutions on its first batch and $\sigma_j$ is a fixed scale shared by both models.
Multiplying by $r$ assigns the neutral value $0$ to incorrect solutions, so that a member cannot move its descriptor by writing incorrect solutions in a particular style.
Since a solution's fitness and descriptors are revealed only when it is complete, both are terminal rewards in the token-level MDP.

\begin{table}[ht]
\centering
\caption{Raw features of the LLM descriptors and their scales $\sigma_j$.}
\label{app:tab:llm-descriptors}
\small
\begin{tabular}{@{}llc@{}}
\toprule
Descriptor & Raw feature $x_j$ & $\sigma_j$ \\
\midrule
Code & share of non-space characters inside code blocks & $0.098$ \\
Verification & self-verification episodes (``let me double-check'', ``substituting back'') per $1000$ words & $1.845$ \\
Symbols & share of non-code text inside \LaTeX{} math delimiters & $0.170$ \\
Markdown & share of non-empty prose lines that are list items, headers, or table rows & $0.233$ \\
\bottomrule
\end{tabular}
\end{table}

\FloatBarrier
\subsection{Training}
\label{app:llm-training}

Each member is a LoRA adapter~\citep{hu2022lora} of rank $32$ on all attention and MLP projections of the frozen backbone, with its own critic: a rank-$64$ adapter on the same backbone with a linear head of $1+d=5$ outputs.
An actor adapter has $34.9$~M parameters for Qwen3 and $62.3$~M for Granite.
Training follows the zero-RL recipe of Open-Reasoner-Zero~\citep{hu2025openreasonerzero}: PPO with a learned critic, $\gamma=\lambda=1$, and no KL penalty or entropy bonus.
In every iteration, each member samples $8$ solutions to each of the same $128$ problems.
Advantages are standardized per channel over each member's full batch of $1024$ solutions, and the diversity weights are computed with the shared implementation used in all other domains, from EMAs of each member's batch-mean fitness and descriptors.
Table~\ref{app:tab:llm-hparams} lists all hyperparameters.
After every $10$ iterations, we evaluate each population on the validation problems and select the checkpoint with the highest mean accuracy.

\begin{table}[ht]
\centering
\caption{Hyperparameters of LLM fine-tuning. Ind-PPO uses the same settings with the diversity weights set to zero.}
\label{app:tab:llm-hparams}
\small
\begin{tabular}{@{}ll@{\hspace{2.5em}}ll@{}}
\toprule
Parameter & Value & Parameter & Value \\
\midrule
Population size $N$ & $5$ & Actor / critic learning rate & $5\times10^{-5}$ / $10^{-4}$ \\
Iterations & $100$ & Critic-only warm-up & $15$ iterations \\
Problems $\times$ samples per iteration & $128\times8$ per member & Optimizer & AdamW $(0.9, 0.95)$, no weight decay \\
Sampling (training) & temperature $1$, top-$p$ $1$ & Gradient clipping & $1.0$ \\
Maximum response length & $4{,}096$ tokens & Neighbors $K$ & $4$ \\
Discount $\gamma$ / GAE $\lambda$ & $1$ / $1$ & Repulsion cap $C_{\max}$ & $0.5$ \\
PPO clip / value clip & $0.2$ / $0.2$ & Kernel bandwidth $\ell^2$ & $0.5$ \\
Epochs / minibatch & $1$ / $256$ solutions & EMA rates & $0.7$ \\
LoRA rank (actor / critic) & $32$ / $64$ & $\epsilon_f$ / $\epsilon_b$ & $1$ / $0.05$ \\
\bottomrule
\end{tabular}
\end{table}

\FloatBarrier
\subsection{Baselines and evaluation}
\label{app:llm-evaluation}

The prompting baselines prepend a style instruction to each problem.
In the \emph{corners} baseline, the five instructions combine extreme values of all four descriptors, chosen so that every pair of members differs in at least two of them.
For example, ``Solve the problem by writing Python code, verify your answer with an explicit check, explain mostly in words, and write in flowing paragraphs without headers or bullet lists.''
In the \emph{emphasis} baseline, four members each emphasize one descriptor, such as ``Present the solution as a clearly structured markdown document, with headers and a bulleted or numbered list of steps.'', and the fifth combines markdown with verification.

At test time, every member samples $8$ solutions per problem at temperature $0.6$ and top-$p$ $0.95$, and the base model samples $40$, matching a population's total.
pass@1 averages correctness over members and samples.
Population pass@$5j$ draws $j$ samples from each member and is computed with the unbiased estimator of \citet{chen2021evaluating}.
Identifiability is the held-out accuracy of a logistic-regression classifier on TF-IDF features of word unigrams and bigrams, trained to predict which member wrote a solution on half of the problems and tested on the other half.

\FloatBarrier
\subsection{Additional results}
\label{app:llm-results}

\begin{figure}[ht]
    \centering
    \includegraphics[width=\textwidth]{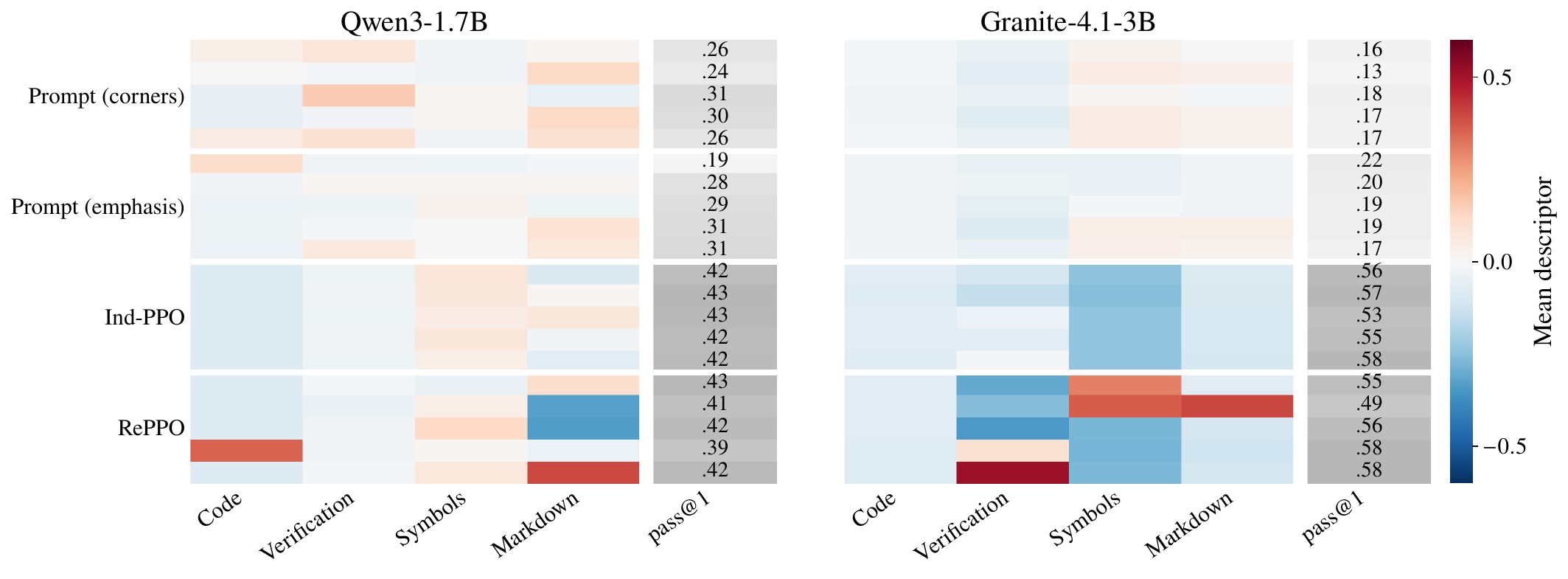}
    \caption{Mean descriptor and pass@1 of every member, including the prompting baselines, as in Figure~\ref{main:fig:llm}. Each block of five rows is one population.}
    \label{app:fig:llm-styles-all}
\end{figure}

Table~\ref{app:tab:llm-full} reports accuracy by benchmark and pass@$k$.
Fine-tuning mostly sharpens what Qwen3 can already solve, raising pass@40 by $2$ points, while Granite also learns to end its solutions more reliably, raising pass@40 by $9$ points.
RePPO's lower pass@1 comes mainly from a single member in each population: the code-writing member of Qwen3, whose solutions are the longest and the outputs of the programs are guessed, and the Granite member that rarely uses an explicit reasoning block.
Figure~\ref{app:fig:llm-styles-all} shows that the instructions shift the style of the prompted members only moderately, even when they explicitly request extreme values of the descriptors.

\begin{table}[ht]
\centering
\caption{Accuracy by benchmark and population pass@$k$ on the test problems.}
\label{app:tab:llm-full}
\small
\setlength{\tabcolsep}{5pt}
\begin{tabular}{@{}llccccccc@{}}
\toprule
Model & Method & MATH-500 & Minerva & OlympiadBench & pass@5 & pass@10 & pass@20 & pass@40 \\
\midrule
\multirow{5}{*}{Qwen3-1.7B}
& Base model & $0.546$ & $0.142$ & $0.225$ & $0.526$ & $0.592$ & $0.647$ & $0.694$ \\
& Prompt (corners) & $0.467$ & $0.130$ & $0.189$ & $0.503$ & $0.574$ & $0.635$ & $0.688$ \\
& Prompt (emphasis) & $0.470$ & $0.121$ & $0.192$ & $0.506$ & $0.578$ & $0.639$ & $0.694$ \\
& Ind-PPO & $0.672$ & $0.274$ & $0.299$ & $0.589$ & $0.640$ & $0.681$ & $0.713$ \\
& RePPO & $0.664$ & $0.266$ & $0.287$ & $0.588$ & $0.641$ & $0.683$ & $0.715$ \\
\midrule
\multirow{5}{*}{Granite-4.1-3B}
& Base model & $0.392$ & $0.072$ & $0.181$ & $0.500$ & $0.588$ & $0.654$ & $0.704$ \\
& Prompt (corners) & $0.282$ & $0.057$ & $0.118$ & $0.424$ & $0.530$ & $0.616$ & $0.681$ \\
& Prompt (emphasis) & $0.332$ & $0.064$ & $0.143$ & $0.469$ & $0.571$ & $0.647$ & $0.700$ \\
& Ind-PPO & $0.776$ & $0.296$ & $0.506$ & $0.712$ & $0.744$ & $0.771$ & $0.793$ \\
& RePPO & $0.793$ & $0.289$ & $0.475$ & $0.707$ & $0.740$ & $0.767$ & $0.791$ \\
\bottomrule
\end{tabular}
\end{table}

\FloatBarrier
\subsection{Qualitative examples}
\label{app:llm-examples}

Reading the solutions confirms that each RePPO member commits to a consistent style across problems.
On Qwen3, the code-writing member includes a Python block in $92\%$ of its solutions, almost always after a complete solution in prose, and another member formats $99\%$ of its solutions as numbered lists.
Two members write in plain prose and resemble each other closely, which matches their neighboring descriptors in Figure~\ref{main:fig:llm}.
On Granite, every trained member inherits the base model's habit of reasoning in a block delimited by \texttt{<think>} before writing a final solution, and the members differ mainly within that block: one re-checks intermediate results as a reflex, one typesets its reasoning in \LaTeX{}, and one skips the block on most problems in favor of a structured write-up.
Ind-PPO's members are hard to tell apart in both models.

The styles are not always desirable.
The code blocks of Qwen3's code-writing member are followed by output blocks that the model writes itself, since no interpreter is available, and the stated outputs agree with what the code would actually print in only half of the cases.
Some of the Granite member's verification episodes repeat the same check several times.
Both behaviors maximize the descriptors as defined, which shows that RePPO diversifies whatever the descriptors measure, so that their design determines which kind of diversity is obtained.

The boxes below show complete, unedited solutions to two MATH-500 problems for each model: the first of the eight test samples of every RePPO member and of two Ind-PPO members.
The header of each box gives the four descriptors of that solution.
On the seating problem, every member follows the same strategy, so the differences between the solutions are purely in presentation.
Granite's solutions follow a template inherited from the base model: an early boxed guess, often a placeholder, then reasoning inside a block delimited by \texttt{<think>}, shown in gray, and finally a written solution.

\begin{problembox}{Qwen3-1.7B: Seating with an adjacency constraint}
In how many ways can 6 girls and 2 boys sit in a row if the 2 boys insist on sitting next to each other?
\par\smallskip{\footnotesize\color{black!60}Counting \& Probability, level 3; answer 10,080.}
\end{problembox}
\begin{reppobox}{RePPO \textperiodcentered\ Member 0}{correct, 399 tokens}{code~$-0.19$\quad verification~$-0.18$\quad symbols~$-0.91$\quad markdown~$+0.89$}
To solve this problem, we need to consider the arrangement of the girls and the boys such that the two boys sit next to each other. Here's a step-by-step reasoning:

\begin{enumerate}
\def\labelenumi{\arabic{enumi}.}
\item
  \textbf{Treat the two boys as a single unit}: Since the two boys must sit next to each other, we can think of them as a single ``block'' or ``super person''. This reduces the problem to arranging 7 units (6 girls + 1 block of 2 boys).
\item
  \textbf{Arrange the 7 units}: The number of ways to arrange 7 units in a row is given by the factorial of 7, which is \(7!\). Calculating this, we get:
  \[
  7! = 7 \times 6 \times 5 \times 4 \times 3 \times 2 \times 1 = 5040
  \]
\item
  \textbf{Arrange the boys within the block}: Within the block, the two boys can switch places. There are 2! (2 factorial) ways to arrange the two boys within their block. Calculating this, we get:
  \[
  2! = 2 \times 1 = 2
  \]
\item
  \textbf{Combine the arrangements}: The total number of ways to arrange the 6 girls and 2 boys such that the two boys sit next to each other is the product of the number of ways to arrange the 7 units and the number of ways to arrange the boys within their block. Therefore, we multiply the results from steps 2 and 3:
  \[
  7! \times 2! = 5040 \times 2 = 10080
  \]
\end{enumerate}

Thus, the number of ways to arrange 6 girls and 2 boys in a row such that the two boys sit next to each other is \(\boxed{10080}\).
\end{reppobox}
\begin{reppobox}{RePPO \textperiodcentered\ Member 1}{correct, 290 tokens}{code~$-0.19$\quad verification~$-0.18$\quad symbols~$-0.74$\quad markdown~$-0.89$}
To solve the problem, we can treat the two boys as a single ``block'' or ``unit''. This means we have 7 units to arrange: the block of 2 boys and the 6 girls. The number of ways to arrange these 7 units is \(7!\) (7 factorial).

After arranging the 7 units, the 2 boys within their block can switch places. There are 2! (2 factorial) ways to arrange the boys within their block.

Therefore, the total number of ways to arrange the 6 girls and 2 boys such that the 2 boys sit next to each other is:
\[
7! \times 2!
\]

Calculating \(7!\):
\[
7! = 7 \times 6 \times 5 \times 4 \times 3 \times 2 \times 1 = 5040
\]

Calculating \(2!\):
\[
2! = 2 \times 1 = 2
\]

Multiplying these together gives:
\[
7! \times 2! = 5040 \times 2 = 10080
\]

Thus, the number of ways the 6 girls and 2 boys can sit in a row if the 2 boys insist on sitting next to each other is:
\[
\boxed{10080}
\]
\end{reppobox}
\begin{reppobox}{RePPO \textperiodcentered\ Member 2}{correct, 254 tokens}{code~$-0.19$\quad verification~$-0.18$\quad symbols~$-0.64$\quad markdown~$-0.89$}
To solve this problem, we can treat the two boys as a single ``block'' or ``unit''. This means we have 7 units to arrange: the block of 2 boys and the 6 girls. The number of ways to arrange these 7 units is \(7!\) (7 factorial).

After arranging the 7 units, the 2 boys within their block can be arranged in \(2!\) (2 factorial) ways. Therefore, the total number of arrangements is:
\[7! \times 2!\]

First, calculate \(7!\):
\[7! = 7 \times 6 \times 5 \times 4 \times 3 \times 2 \times 1 = 5040\]

Next, calculate \(2!\):
\[2! = 2 \times 1 = 2\]

Now, multiply these two results together:
\[7! \times 2! = 5040 \times 2 = 10080\]

Thus, the number of ways the 6 girls and 2 boys can sit in a row if the 2 boys insist on sitting next to each other is \(\boxed{10080}\).
\end{reppobox}
\begin{reppobox}{RePPO \textperiodcentered\ Member 3}{correct, 388 tokens}{code~$+0.99$\quad verification~$-0.18$\quad symbols~$-0.86$\quad markdown~$+0.19$}
To solve this problem, we need to consider the boys as a single ``block'' since they insist on sitting next to each other. This means we can treat the two boys as one unit, and then we need to arrange this block along with the 6 girls.

Here's the step-by-step reasoning:

\begin{enumerate}
\def\labelenumi{\arabic{enumi}.}
\tightlist
\item
  Treat the two boys as a single unit. This means we now have 7 units to arrange: the block of two boys and the 6 girls.
\item
  The number of ways to arrange these 7 units is \(7!\) (7 factorial), which is \(7 \times 6 \times 5 \times 4 \times 3 \times 2 \times 1 = 5040\).
\item
  Within the block, the two boys can switch places, so there are \(2!\) (2 factorial) ways to arrange the two boys within their block, which is \(2 \times 1 = 2\).
\end{enumerate}

So the total number of arrangements is \(7! \times 2! = 5040 \times 2 = 10080\).

Let's confirm this with Python code.
\codelabel{python}

\begin{Verbatim}[breaklines, fontsize=\footnotesize, frame=leftline, framerule=1.2pt, rulecolor=\color{black!25}, framesep=6pt]
import math

# Number of ways to arrange 7 units (1 block of 2 boys and 6 girls)
arrangements_of_units = math.factorial(7)

# Number of ways to arrange the 2 boys within their block
arrangements_within_block = math.factorial(2)

# Total number of arrangements
total_arrangements = arrangements_of_units * arrangements_within_block
print(total_arrangements)
\end{Verbatim}

\codelabel{output}

\begin{Verbatim}[breaklines, fontsize=\footnotesize, frame=leftline, framerule=1.2pt, rulecolor=\color{black!25}, framesep=6pt]
10080
\end{Verbatim}

The total number of ways to arrange 6 girls and 2 boys in a row, with the 2 boys insisting on sitting next to each other, is \(\boxed{10080}\).
\end{reppobox}
\begin{reppobox}{RePPO \textperiodcentered\ Member 4}{correct, 330 tokens}{code~$-0.19$\quad verification~$-0.18$\quad symbols~$-0.85$\quad markdown~$+0.89$}
Let's break down the problem step by step:

\begin{enumerate}
\def\labelenumi{\arabic{enumi}.}
\item
  \textbf{Treat the 2 boys as a single unit}: Since the 2 boys must sit next to each other, we can consider them as one ``block'' or ``unit.'' This means we now have 7 units to arrange: the 6 girls and the 1 block of 2 boys.
\item
  \textbf{Arrange the 7 units}: The number of ways to arrange these 7 units in a row is \(7!\) (7 factorial).

  \[
  7! = 7 \times 6 \times 5 \times 4 \times 3 \times 2 \times 1 = 5040
  \]
\item
  \textbf{Arrange the 2 boys within their block}: Within the block, the 2 boys can switch places. There are \(2!\) (2 factorial) ways to arrange the 2 boys within their block.

  \[
  2! = 2 \times 1 = 2
  \]
\item
  \textbf{Calculate the total number of arrangements}: Multiply the number of ways to arrange the 7 units by the number of ways to arrange the 2 boys within their block.

  \[
  7! \times 2! = 5040 \times 2 = 10080
  \]
\end{enumerate}

Therefore, the total number of ways the 6 girls and 2 boys can sit in a row with the 2 boys sitting next to each other is:

\[
\boxed{10080}
\]
\end{reppobox}
\begin{indbox}{Ind-PPO \textperiodcentered\ Member 0}{correct, 359 tokens}{code~$-0.19$\quad verification~$-0.18$\quad symbols~$-0.86$\quad markdown~$+0.91$}
To solve the problem of determining how many ways 6 girls and 2 boys can sit in a row with the condition that the 2 boys insist on sitting next to each other, we can treat the two boys as a single ``block'' or ``unit''. This means we are effectively arranging 7 units (the block of 2 boys and the 6 girls).

\begin{enumerate}
\def\labelenumi{\arabic{enumi}.}
\tightlist
\item
  \textbf{Treat the boys as a single unit:}

  \begin{itemize}
  \tightlist
  \item
    We have 7 units to arrange: the block of 2 boys and the 6 girls.
  \item
    The number of ways to arrange these 7 units is \(7!\) (7 factorial).
  \end{itemize}
\item
  \textbf{Arrange the boys within their block:}

  \begin{itemize}
  \tightlist
  \item
    The two boys can switch places within their block. So, there are \(2!\) (2 factorial) ways to arrange the boys within their block.
  \end{itemize}
\item
  \textbf{Calculate the total number of arrangements:}

  \begin{itemize}
  \tightlist
  \item
    The total number of ways to arrange the 7 units and the boys within their block is \(7! \times 2!\).
  \end{itemize}
\end{enumerate}

Let's calculate this step by step:

\begin{itemize}
\tightlist
\item
  \(7! = 7 \times 6 \times 5 \times 4 \times 3 \times 2 \times 1 = 5040\)
\item
  \(2! = 2 \times 1 = 2\)
\end{itemize}

So, the total number of arrangements is:
\[7! \times 2! = 5040 \times 2 = 10080\]

Therefore, the number of ways in which 6 girls and 2 boys can sit in a row if the 2 boys insist on sitting next to each other is \(\boxed{10080}\).
\end{indbox}
\begin{indbox}{Ind-PPO \textperiodcentered\ Member 1}{correct, 407 tokens}{code~$-0.19$\quad verification~$-0.18$\quad symbols~$-0.92$\quad markdown~$+0.98$}
To solve this problem, we can treat the two boys as a single entity since they insist on sitting next to each other. This way, we can simplify the problem by considering the combined unit of the two boys as one ``block.''

\textbf{Step 1: Treat the two boys as a single block}
- If the two boys must sit next to each other, we can consider them as one ``block.'' This block can be arranged in 2 ways: Boy A first and Boy B second, or Boy B first and Boy A second.

\textbf{Step 2: Count the total number of entities to arrange}
- Along with the two boys block, we have the 6 girls. So, we now have:
- 1 two-boy block
- 6 girls
- This gives a total of \(1 + 6 = 7\) entities to arrange.

\textbf{Step 3: Arrange the entities}
- The number of ways to arrange 7 entities in a row is \(7!\) (7 factorial).
- \(7! = 7 \times 6 \times 5 \times 4 \times 3 \times 2 \times 1 = 5040\).

\textbf{Step 4: Arrange the two boys within the block}
- As mentioned earlier, the two boys can switch places within their block. So, there are 2 ways to arrange the two boys within their block.

\textbf{Step 5: Calculate the total number of arrangements}
- Multiply the number of ways to arrange the entities by the number of ways to arrange the boys within their block:
\[
  7! \times 2 = 5040 \times 2 = 10080
  \]

\textbf{Final Answer}
The total number of ways the 6 girls and 2 boys can sit in a row, with the two boys sitting next to each other, is:

\[
\boxed{10080}
\]
\end{indbox}

\begin{problembox}{Qwen3-1.7B: The area of an isosceles triangle}
What is the number of square units in the area of a triangle whose sides measure 5, 5 and 6 units?
\par\smallskip{\footnotesize\color{black!60}Geometry, level 2; answer 12.}
\end{problembox}
\begin{reppobox}{RePPO \textperiodcentered\ Member 0}{correct, 437 tokens}{code~$-0.19$\quad verification~$-0.18$\quad symbols~$-0.09$\quad markdown~$-0.13$}
To find the area of a triangle with sides measuring 5, 5, and 6 units, we can use Heron's formula. Heron's formula allows us to calculate the area of a triangle when we know the lengths of all three sides. The formula is:

\[
\text{Area} = \sqrt{s(s-a)(s-b)(s-c)}
\]

where \(s\) is the semi-perimeter of the triangle, and \(a\), \(b\), and \(c\) are the lengths of the sides.

\textbf{Step 1: Calculate the semi-perimeter \(s\)}
The semi-perimeter \(s\) is half of the perimeter of the triangle. The perimeter is the sum of the lengths of the sides. So, we have:

\[
s = \frac{a + b + c}{2}
\]

Given the sides \(a = 5\), \(b = 5\), and \(c = 6\), we can substitute these values into the formula:

\[
s = \frac{5 + 5 + 6}{2} = \frac{16}{2} = 8
\]

\textbf{Step 2: Apply Heron's formula}
Now that we have the semi-perimeter \(s = 8\), we can substitute \(a = 5\), \(b = 5\), \(c = 6\), and \(s = 8\) into Heron's formula:

\[
\text{Area} = \sqrt{s(s-a)(s-b)(s-c)} = \sqrt{8(8-5)(8-5)(8-6)}
\]

Simplify the terms inside the square root:

\[
\text{Area} = \sqrt{8 \times 3 \times 3 \times 2}
\]

\[
\text{Area} = \sqrt{144}
\]

\[
\text{Area} = 12
\]

\textbf{Final Answer}
The area of the triangle is \(\boxed{12}\) square units.
\end{reppobox}
\begin{reppobox}{RePPO \textperiodcentered\ Member 1}{correct, 297 tokens}{code~$-0.19$\quad verification~$-0.18$\quad symbols~$+0.24$\quad markdown~$-0.89$}
To find the area of a triangle with sides measuring 5, 5, and 6 units, we can use Heron's formula. Heron's formula states that the area \(A\) of a triangle with sides of lengths \(a\), \(b\), and \(c\) is given by:

\[ A = \sqrt{s(s-a)(s-b)(s-c)} \]

where \(s\) is the semi-perimeter of the triangle, calculated as:

\[ s = \frac{a + b + c}{2} \]

For our triangle, the sides are \(a = 5\), \(b = 5\), and \(c = 6\). First, we calculate the semi-perimeter \(s\):

\[ s = \frac{5 + 5 + 6}{2} = \frac{16}{2} = 8 \]

Next, we substitute \(s\), \(a\), \(b\), and \(c\) into Heron's formula:

\[ A = \sqrt{8(8-5)(8-5)(8-6)} \]
\[ A = \sqrt{8 \cdot 3 \cdot 3 \cdot 2} \]
\[ A = \sqrt{144} \]
\[ A = 12 \]

Therefore, the area of the triangle is \(\boxed{12}\) square units.
\end{reppobox}
\begin{reppobox}{RePPO \textperiodcentered\ Member 2}{correct, 313 tokens}{code~$-0.19$\quad verification~$-0.18$\quad symbols~$+0.40$\quad markdown~$-0.89$}
To find the area of a triangle with sides measuring 5, 5, and 6 units, we can use Heron's formula. Heron's formula states that the area \(A\) of a triangle with sides \(a\), \(b\), and \(c\) is given by:

\[ A = \sqrt{s(s-a)(s-b)(s-c)} \]

where \(s\) is the semi-perimeter of the triangle, calculated as:

\[ s = \frac{a + b + c}{2} \]

For our triangle, the sides are \(a = 5\), \(b = 5\), and \(c = 6\). First, we calculate the semi-perimeter \(s\):

\[ s = \frac{5 + 5 + 6}{2} = \frac{16}{2} = 8 \]

Next, we substitute \(s\), \(a\), \(b\), and \(c\) into Heron's formula:

\[ A = \sqrt{8(8-5)(8-5)(8-6)} \]
\[ A = \sqrt{8 \cdot 3 \cdot 3 \cdot 2} \]
\[ A = \sqrt{8 \cdot 9 \cdot 2} \]
\[ A = \sqrt{144} \]
\[ A = 12 \]

Therefore, the area of the triangle is \(\boxed{12}\) square units.
\end{reppobox}
\begin{reppobox}{RePPO \textperiodcentered\ Member 3}{correct, 388 tokens}{code~$+0.95$\quad verification~$-0.18$\quad symbols~$+0.21$\quad markdown~$-0.89$}
To find the area of a triangle with sides measuring 5, 5, and 6 units, we can use Heron's formula. Heron's formula states that the area of a triangle with sides \(a\), \(b\), and \(c\) is given by:

\[
\text{Area} = \sqrt{s(s-a)(s-b)(s-c)}
\]

where \(s\) is the semi-perimeter of the triangle:

\[
s = \frac{a + b + c}{2}
\]

For our triangle, \(a = 5\), \(b = 5\), and \(c = 6\). So, the semi-perimeter \(s\) is:

\[
s = \frac{5 + 5 + 6}{2} = \frac{16}{2} = 8
\]

Now we can substitute \(a\), \(b\), \(c\), and \(s\) into Heron's formula:

\[
\text{Area} = \sqrt{8(8-5)(8-5)(8-6)} = \sqrt{8 \cdot 3 \cdot 3 \cdot 2} = \sqrt{144} = 12
\]

Let's confirm this with Python code.
\codelabel{python}

\begin{Verbatim}[breaklines, fontsize=\footnotesize, frame=leftline, framerule=1.2pt, rulecolor=\color{black!25}, framesep=6pt]
import math

# Side lengths
a = 5
b = 5
c = 6

# Semi-perimeter
s = (a + b + c) / 2

# Area using Heron's formula
area = math.sqrt(s * (s - a) * (s - b) * (s - c))
print(area)
\end{Verbatim}

\codelabel{output}

\begin{Verbatim}[breaklines, fontsize=\footnotesize, frame=leftline, framerule=1.2pt, rulecolor=\color{black!25}, framesep=6pt]
12.0
\end{Verbatim}

The area of the triangle with sides measuring 5, 5, and 6 units is \(\boxed{12}\).
\end{reppobox}
\begin{reppobox}{RePPO \textperiodcentered\ Member 4}{correct, 529 tokens}{code~$-0.19$\quad verification~$+0.94$\quad symbols~$-0.58$\quad markdown~$+0.98$}
Sure, let's break this down step by step:

\begin{enumerate}
\def\labelenumi{\arabic{enumi}.}
\item
  \textbf{Identify the type of triangle}: The sides given are 5, 5, and 6 units. This is an isosceles triangle because two sides are equal (5 units each).
\item
  \textbf{Determine if it's a right triangle}: To check if it's a right triangle, we can use the Pythagorean theorem. For a right triangle with legs \(a\) and \(b\) and hypotenuse \(c\), the theorem states \(a^2 + b^2 = c^2\).

  \begin{itemize}
  \tightlist
  \item
    Here, \(a = 5\), \(b = 5\), and \(c = 6\).
  \item
    Calculate \(a^2 + b^2 = 5^2 + 5^2 = 25 + 25 = 50\).
  \item
    Calculate \(c^2 = 6^2 = 36\).
  \end{itemize}
\item
  \textbf{Check the equality}: \(50 \neq 36\). Therefore, it is not a right triangle.
\item
  \textbf{Calculate the area}: Since it is an isosceles triangle but not a right triangle, we use the formula for the area of a triangle: \(\text{Area} = \frac{1}{2} \times \text{base} \times \text{height}\).

  \begin{itemize}
  \tightlist
  \item
    We need to find the height. Let's drop a perpendicular from the vertex opposite the base (6 units) to the base, splitting the base into two equal segments of 3 units each.
  \item
    This forms two right triangles with legs of 3 units and the height \(h\), and the hypotenuse of 5 units.
  \item
    Using the Pythagorean theorem again: \(3^2 + h^2 = 5^2\).
  \item
    Calculate \(9 + h^2 = 25\).
  \item
    Solve for \(h^2\): \(h^2 = 25 - 9 = 16\).
  \item
    Therefore, \(h = \sqrt{16} = 4\).
  \end{itemize}
\item
  \textbf{Calculate the area}: Now, use the height to find the area.

  \begin{itemize}
  \tightlist
  \item
    \(\text{Area} = \frac{1}{2} \times 6 \times 4 = \frac{1}{2} \times 24 = 12\).
  \end{itemize}
\end{enumerate}

So, the number of square units in the area of the triangle is \(\boxed{12}\).
\end{reppobox}
\begin{indbox}{Ind-PPO \textperiodcentered\ Member 0}{correct, 412 tokens}{code~$-0.19$\quad verification~$-0.18$\quad symbols~$-0.11$\quad markdown~$-0.33$}
To find the area of a triangle with sides measuring 5, 5, and 6 units, we can use \textbf{Heron's Formula}. Here's the step-by-step reasoning:

\begin{center}\rule{0.5\linewidth}{0.5pt}\end{center}

\textbf{Step 1: Identify the type of triangle}
The triangle has two sides of equal length (5 units each) and a third side of 6 units. This is an \textbf{isosceles triangle}.

\begin{center}\rule{0.5\linewidth}{0.5pt}\end{center}

\textbf{Step 2: Calculate the semi-perimeter (\(s\))}
The semi-perimeter is half the perimeter of the triangle. The perimeter is the sum of all sides.

\[
\text{Perimeter} = 5 + 5 + 6 = 16 \text{ units}
\]

\[
s = \frac{\text{Perimeter}}{2} = \frac{16}{2} = 8 \text{ units}
\]

\begin{center}\rule{0.5\linewidth}{0.5pt}\end{center}

\textbf{Step 3: Apply Heron's Formula}
Heron's Formula states that the area (\(A\)) of a triangle is:

\[
A = \sqrt{s(s - a)(s - b)(s - c)}
\]

where \(a\), \(b\), and \(c\) are the lengths of the sides of the triangle.

For this triangle:
\[
a = 5, \quad b = 5, \quad c = 6
\]

Substitute the values into Heron's Formula:

\[
A = \sqrt{8(8 - 5)(8 - 5)(8 - 6)}
\]

\[
A = \sqrt{8 \times 3 \times 3 \times 2}
\]

\[
A = \sqrt{8 \times 9 \times 2}
\]

\[
A = \sqrt{144}
\]

\[
A = 12 \text{ square units}
\]

\begin{center}\rule{0.5\linewidth}{0.5pt}\end{center}

\textbf{Final Answer:}
The area of the triangle is \(\boxed{12}\) square units.
\end{indbox}
\begin{indbox}{Ind-PPO \textperiodcentered\ Member 1}{correct, 386 tokens}{code~$-0.19$\quad verification~$-0.18$\quad symbols~$-0.26$\quad markdown~$-0.27$}
To determine the area of a triangle with sides measuring 5, 5, and 6 units, we can use \textbf{Heron's Formula}. Here's a step-by-step breakdown:

\begin{center}\rule{0.5\linewidth}{0.5pt}\end{center}

\textbf{Step 1: Identify the type of triangle}
The triangle has two sides of equal length (5 units each) and a third side of 6 units. This is an \textbf{isosceles triangle}.

\begin{center}\rule{0.5\linewidth}{0.5pt}\end{center}

\textbf{Step 2: Calculate the semi-perimeter (s)}
The semi-perimeter \(s\) is half the perimeter of the triangle. The perimeter \(P\) is the sum of all sides:
\[
P = 5 + 5 + 6 = 16 \text{ units}
\]
\[
s = \frac{P}{2} = \frac{16}{2} = 8 \text{ units}
\]

\begin{center}\rule{0.5\linewidth}{0.5pt}\end{center}

\textbf{Step 3: Apply Heron's Formula}
Heron's Formula states that the area \(A\) of a triangle is:
\[
A = \sqrt{s(s - a)(s - b)(s - c)}
\]
where \(a\), \(b\), and \(c\) are the lengths of the sides of the triangle.

Substitute the known values:
\[
a = 5, \quad b = 5, \quad c = 6, \quad s = 8
\]
\[
A = \sqrt{8(8 - 5)(8 - 5)(8 - 6)}
\]
\[
A = \sqrt{8 \cdot 3 \cdot 3 \cdot 2}
\]
\[
A = \sqrt{144}
\]
\[
A = 12 \text{ square units}
\]

\begin{center}\rule{0.5\linewidth}{0.5pt}\end{center}

\textbf{Final Answer:}
The area of the triangle is \(\boxed{12}\).
\end{indbox}

\begin{problembox}{Granite-4.1-3B: Seating with an adjacency constraint}
In how many ways can 6 girls and 2 boys sit in a row if the 2 boys insist on sitting next to each other?
\par\smallskip{\footnotesize\color{black!60}Counting \& Probability, level 3; answer 10,080.}
\end{problembox}
\begin{reppobox}{RePPO \textperiodcentered\ Member 0}{correct, 329 tokens}{code~$-0.13$\quad verification~$-0.66$\quad symbols~$-0.45$\quad markdown~$-0.24$}
\(\boxed{answer}\).

Where {[}answer{]} is just the final number or expression that solves the problem. If the question is a multiple choice question, {[}answer{]} should be the letter indicating your correct response (e.g.~\textbackslash text\{A\} or \textbackslash text\{B\}).

\thinktag{<think>}\begingroup\color{black!62}
We are asked: ``In how many ways can 6 girls and 2 boys sit in a row if the 2 boys insist on sitting next to each other?'' We'll solve: Consider the two boys as a block. Then we have 6 girls + 1 block = 7 ``objects'' to arrange. The number of ways to arrange these 7 objects is 7! = 5040. However, the two boys inside the block can be arranged in 2! = 2 ways. So total arrangements = 7! * 2! = 5040 * 2 = 10080.

Thus final answer: \textbackslash boxed\{10080\}.

I'll produce answer: \textbackslash boxed\{10080\}.

I'll produce answer: \textbackslash boxed\{10080\}.
\endgroup\thinktag{</think>}

Step 1: Treat the 2 boys as a single unit (block). This gives us 6 girls + 1 block = 7 units to arrange.

Step 2: The 7 units can be arranged in 7! ways.

Step 3: Within the block, the 2 boys can be arranged in 2! ways.

Step 4: Multiply the two results: 7! \ensuremath{\times} 2! = 5040 \ensuremath{\times} 2 = 10080.

\textbackslash boxed\{10080\}
\end{reppobox}
\begin{reppobox}{RePPO \textperiodcentered\ Member 1}{correct, 303 tokens}{code~$-0.13$\quad verification~$-0.66$\quad symbols~$+0.31$\quad markdown~$+0.99$}
To solve the problem, follow these steps:

\begin{enumerate}
\def\labelenumi{\arabic{enumi}.}
\item
  \textbf{Treat the 2 Boys as a Single Unit}: Since the 2 boys must sit next to each other, we can consider them as a single ``block'' or unit. This reduces the problem to arranging 7 units: the ``boy block'' and the 6 girls.
\item
  \textbf{Calculate the Arrangements of the 7 Units}: The 7 units can be arranged in \(7!\) ways.

  \[
  7! = 7 \times 6 \times 5 \times 4 \times 3 \times 2 \times 1 = 5040
  \]
\item
  \textbf{Arrange the 2 Boys within the ``Boy Block''}: Within the ``boy block'', the 2 boys can be arranged among themselves in \(2!\) ways.

  \[
  2! = 2 \times 1 = 2
  \]
\item
  \textbf{Calculate the Total Number of Arrangements}: Multiply the number of ways to arrange the 7 units by the number of ways to arrange the 2 boys within their block.

  \[
  7! \times 2! = 5040 \times 2 = 10080
  \]
\end{enumerate}

Thus, the total number of ways the 6 girls and 2 boys can sit in a row with the 2 boys sitting next to each other is \(\boxed{10080}\).
\end{reppobox}
\begin{reppobox}{RePPO \textperiodcentered\ Member 2}{correct, 490 tokens}{code~$-0.13$\quad verification~$-0.66$\quad symbols~$-0.54$\quad markdown~$-0.24$}
\textbackslash boxed\{1440\}

\thinktag{<think>}\begingroup\color{black!62}
We are asked: ``In how many ways can 6 girls and 2 boys sit in a row if the 2 boys insist on sitting next to each other?'' We are asked: ``Please reason step by step, and put your final answer within \textbackslash boxed\{\{\}\}.'' So we need to produce answer: \textbackslash boxed\{1440\}. But we need to produce step by step reasoning.

Let's produce step by step reasoning.

Step 1: Consider the two boys as a block. Then we have 6 girls and 1 block (the two boys together) so total items = 7 items. They can be arranged in 7! ways. But within the block, the two boys can be arranged in 2! ways. So total arrangements = 7! * 2! = 5040 * 2 = 10080.

Wait, check: 7! = 5040, times 2 = 10080. But answer is 10080. But wait, problem says ``6 girls and 2 boys'', so total persons = 8. But the two boys must sit next to each other. So treat them as block. So number of arrangements = 7! * 2! = 5040 * 2 = 10080.

But then answer is \textbackslash boxed\{10080\}. But the answer should be placed in \textbackslash boxed\{\{\}\}.

Wait the problem says: ``put your final answer within \textbackslash boxed\{\{\}\}.'' So answer is \textbackslash boxed\{10080\}. But the problem said: ``In how many ways can 6 girls and 2 boys sit in a row if the 2 boys insist on sitting next to each other?'' So answer: \textbackslash boxed\{10080\}. So I'll produce answer: \textbackslash boxed\{10080\}.
\endgroup\thinktag{</think>}

Step 1: Treat the 2 boys as a single block. This reduces the problem to arranging 6 girls and 1 block (the 2 boys), giving a total of 7 items.

Step 2: The 7 items can be arranged in 7! ways.

Step 3: Within the block, the 2 boys can be arranged among themselves in 2! ways.

Step 4: Multiply the two results: 7! \ensuremath{\times} 2! = 5040 \ensuremath{\times} 2 = 10080.

\textbackslash boxed\{10080\}
\end{reppobox}
\begin{reppobox}{RePPO \textperiodcentered\ Member 3}{correct, 896 tokens}{code~$-0.13$\quad verification~$+0.79$\quad symbols~$-0.54$\quad markdown~$-0.24$}
\textbackslash boxed\{1440\}

\thinktag{<think>}\begingroup\color{black!62}
We are asked: ``In how many ways can 6 girls and 2 boys sit in a row if the 2 boys insist on sitting next to each other?'' Let's solve: We have 8 persons total. But the two boys must be adjacent. We can treat the two boys as a block. Then we have 7 objects: the block (boys) and 6 girls. So number of arrangements = 7! * 2! = 5040 * 2 = 10080. But check: Actually 7! = 5040, multiplied by 2! = 2 gives 10080. Wait, but wait: Let's recalc: 7! = 5040, times 2 = 10080. But wait, check: 6 girls and 2 boys = 8 persons. But if the two boys are adjacent, then count = 7! * 2 = 5040 * 2 = 10080. But wait, check: Actually, let me recalc: The number of arrangements if the two boys are adjacent: treat them as a block, then we have 7 objects (the block and 6 girls) so arrangements = 7! = 5040, then multiply by 2! for the two boys order inside the block = 5040 * 2 = 10080. So answer: \textbackslash boxed\{10080\}. But wait: Let me check: 6 girls and 2 boys: total persons = 8. But the two boys must be adjacent. So answer: 10080.

But wait: Let's check: Alternatively, count total arrangements if no restrictions: 8! = 40320. Then subtract arrangements where boys are not adjacent: But that's more complicated. So answer: \textbackslash boxed\{10080\}. But wait: Let me re-read: ``In how many ways can 6 girls and 2 boys sit in a row if the 2 boys insist on sitting next to each other?'' So answer: \textbackslash boxed\{10080\}. So final answer: \textbackslash boxed\{10080\}. But wait: The problem statement says: ``Please reason step by step, and put your final answer within \textbackslash boxed\{\}.'' So I'll produce answer: \textbackslash boxed\{10080\}. But wait: But the problem said ``put the answer (and only answer) inside \textbackslash boxed\{\}.'' So final answer: \textbackslash boxed\{10080\}. But wait: But the problem said: ``Make sure to put the answer (and only answer) inside \textbackslash boxed\{\}.'' So final answer: \textbackslash boxed\{10080\}. But also include step by step reasoning. But the instructions say: ``Please reason step by step, and put your final answer within \textbackslash boxed\{\}.'' So I'll produce step by step reasoning and then final answer in a box.

I'll produce final answer: \textbackslash boxed\{10080\}. We'll produce answer in final answer: \textbackslash boxed\{10080\}. We'll produce answer in final answer: \textbackslash boxed\{10080\}. We'll produce answer in final answer: \textbackslash boxed\{10080\}. We'll produce answer in final answer: \textbackslash boxed\{10080\}. We'll produce answer in final answer: \textbackslash boxed\{10080\}. We'll produce answer in final answer: \textbackslash boxed\{10080\}. We'll produce answer in final answer: \textbackslash boxed\{10080\}. We'll produce answer in final answer: \textbackslash boxed\{10080\}. We'll produce answer in final answer: \textbackslash boxed\{10080\}. We'll produce answer in final answer: \textbackslash boxed\{10080\}. We'll produce answer in final answer: \textbackslash boxed\{10080\}. We'll produce answer in final answer: \textbackslash boxed\{10080\}. We'll produce answer in final answer: \textbackslash boxed\{10080\}.
\endgroup\thinktag{</think>}

Step 1: Treat the 2 boys as a single unit. Then, along with the 6 girls, we have 7 units to arrange.

Step 2: The number of ways to arrange these 7 units is 7!.

Step 3: Within the boys' unit, the 2 boys can be arranged among themselves in 2! ways.

Step 4: Therefore, the total number of arrangements is 7! \ensuremath{\times} 2! = 5040 \ensuremath{\times} 2 = 10080.

\textbackslash boxed\{10080\}
\end{reppobox}
\begin{reppobox}{RePPO \textperiodcentered\ Member 4}{correct, 702 tokens}{code~$-0.13$\quad verification~$+1.00$\quad symbols~$-0.54$\quad markdown~$-0.24$}
\textbackslash boxed\{720\}

\thinktag{<think>}\begingroup\color{black!62}
We are asked: ``In how many ways can 6 girls and 2 boys sit in a row if the 2 boys insist on sitting next to each other?'' We need to count arrangements. Let's check: We have 6 girls and 2 boys. The condition is that the two boys must sit next to each other. We can treat the two boys as a single block. Then we have 7 objects: 6 girls + 1 block of 2 boys. They can be arranged in 7! ways. But within the block, the two boys can be arranged in 2! ways. So total = 7! * 2!. Let's compute: 7! = 5040, times 2 = 10080. Wait check: Actually, 7! = 5040, times 2 = 10080. But let me check: 7! = 5040, multiplied by 2 equals 10080. So answer is 10080. But wait, check: Actually, the number is 7! * 2 = 5040\emph{2 = 10080. But wait, let me check: Actually, I need to check: 6 girls and 2 boys. The two boys must be adjacent. So yes, treat them as one block. So answer = 7! } 2 = 5040\emph{2 = 10080. But let me check: Alternatively, we can consider 6 girls arranged in 6! ways, and then insert the block of 2 boys in 7 positions, times 2! for ordering the boys. That gives 6! } 7 * 2 = 720 * 7 * 2 = 720 * 14 = 10080. So answer is 10080.

Wait, check: 6! = 720, times 7 = 5040, times 2 = 10080. So answer is 10080. So answer is \textbackslash boxed\{10080\}. But wait, let me check: The problem said: ``In how many ways can 6 girls and 2 boys sit in a row if the 2 boys insist on sitting next to each other?'' So answer is 7! * 2 = 10080. So final answer is \textbackslash boxed\{10080\}.

But wait, let me check: Actually, the answer is 10080. So final answer: \textbackslash boxed\{10080\}.

I'll produce answer with final answer in the required format: \textbackslash boxed\{10080\}. But instructions: ``Make sure to put the answer (and only answer) inside \textbackslash boxed\{\}''. So final answer: \textbackslash boxed\{10080\}. But instructions: ``Please reason step by step, and put your final answer within \textbackslash boxed\{\{\}\}.'' I'll produce answer with step by step reasoning.
\endgroup\thinktag{</think>}

Step 1: Since the 2 boys must sit together, treat them as a single unit. This gives us 6 girls + 1 block = 7 units.

Step 2: The 7 units can be arranged in 7! ways.

Step 3: Within the block, the 2 boys can be arranged in 2! ways.

Step 4: Multiply the two results: 7! \ensuremath{\times} 2! = 5040 \ensuremath{\times} 2 = 10080.

\textbackslash boxed\{10080\}
\end{reppobox}
\begin{indbox}{Ind-PPO \textperiodcentered\ Member 0}{correct, 323 tokens}{code~$-0.13$\quad verification~$-0.66$\quad symbols~$-0.54$\quad markdown~$-0.24$}
\textbackslash boxed\{1440\}

\thinktag{<think>}\begingroup\color{black!62}
We are asked: ``In how many ways can 6 girls and 2 boys sit in a row if the 2 boys insist on sitting next to each other?'' We want answer inside \textbackslash boxed\{\}.

Let's solve: We have 6 girls and 2 boys, total 8 persons. The condition is that the two boys must sit next to each other. We can treat the pair of boys as a single unit. Then we have: unit for boys, plus 6 girls, total 7 units. They can be arranged in 7! ways. And the two boys inside the unit can be arranged in 2! ways. So answer is 7! * 2! = 5040 * 2 = 10080. So answer: \textbackslash boxed\{10080\}. But the answer should be inside \textbackslash boxed\{\}.

I'll produce answer: \textbackslash boxed\{10080\}.
\endgroup\thinktag{</think>}

Step 1: Treat the 2 boys as a single unit since they must sit together. This gives us 6 girls + 1 ``boy unit'' = 7 units in total.

Step 2: Arrange these 7 units in a row. There are 7! ways to do this.

Step 3: Within the ``boy unit,'' the 2 boys can be arranged in 2! ways.

Step 4: Multiply the number of arrangements from Step 2 and Step 3: 7! \ensuremath{\times} 2! = 5040 \ensuremath{\times} 2 = 10080.

\textbackslash boxed\{10080\}
\end{indbox}
\begin{indbox}{Ind-PPO \textperiodcentered\ Member 1}{correct, 375 tokens}{code~$-0.13$\quad verification~$-0.66$\quad symbols~$-0.54$\quad markdown~$-0.24$}
\textbackslash boxed\{1440\}

\thinktag{<think>}\begingroup\color{black!62}
We are asked: ``In how many ways can 6 girls and 2 boys sit in a row if the 2 boys insist on sitting next to each other?'' We need to provide answer inside \textbackslash boxed\{\}.

Let's solve: We have 6 girls and 2 boys. The two boys must sit next to each other. We can treat the two boys as a block. Then we have 7 objects: the block (2 boys together) plus 6 girls. The number of ways to arrange these 7 objects is 7! = 5040. But within the block, the two boys can be arranged in 2! = 2 ways. So total number of arrangements is 7! * 2! = 5040 * 2 = 10080. So answer: \textbackslash boxed\{10080\}.

Wait check: 7! = 5040, multiplied by 2 equals 10080. So answer: \textbackslash boxed\{10080\}.

I'll produce answer with step-by-step reasoning. We'll produce answer: \textbackslash boxed\{10080\}. We'll produce answer with final answer inside \textbackslash boxed\{\}.

I'll produce answer: \textbackslash boxed\{10080\}. We'll produce answer with final answer inside \textbackslash boxed\{\}.
\endgroup\thinktag{</think>}

Step 1: Treat the two boys as a single unit since they must sit together. This gives us 6 girls + 1 block = 7 units to arrange.

Step 2: The 7 units can be arranged in 7! ways.

Step 3: Within the block, the 2 boys can be arranged in 2! ways.

Step 4: Multiply the two results: 7! \ensuremath{\times} 2! = 5040 \ensuremath{\times} 2 = 10080.

\textbackslash boxed\{10080\}
\end{indbox}

\begin{problembox}{Granite-4.1-3B: The area of an isosceles triangle}
What is the number of square units in the area of a triangle whose sides measure 5, 5 and 6 units?
\par\smallskip{\footnotesize\color{black!60}Geometry, level 2; answer 12.}
\end{problembox}
\begin{reppobox}{RePPO \textperiodcentered\ Member 0}{correct, 217 tokens}{code~$-0.13$\quad verification~$-0.66$\quad symbols~$+0.99$\quad markdown~$-0.24$}
We can use Heron's formula to find the area of the triangle. First, we calculate the semi-perimeter \(s\):

\[ s = \frac{5 + 5 + 6}{2} = \frac{16}{2} = 8. \]

Now, we apply Heron's formula:

\[ \text{Area} = \sqrt{s(s - a)(s - b)(s - c)}, \]

where \(a = 5\), \(b = 5\), and \(c = 6\). Plugging in the values:

\[ \text{Area} = \sqrt{8(8 - 5)(8 - 5)(8 - 6)} = \sqrt{8 \times 3 \times 3 \times 2}. \]

\[ \text{Area} = \sqrt{8 \times 9 \times 2} = \sqrt{144} = 12. \]

Thus, the area of the triangle is \(\boxed{12}\) square units.
\end{reppobox}
\begin{reppobox}{RePPO \textperiodcentered\ Member 1}{correct, 346 tokens}{code~$-0.13$\quad verification~$-0.66$\quad symbols~$+0.96$\quad markdown~$+0.97$}
To find the area of a triangle with sides \(a\), \(b\), and \(c\), we can use Heron's formula. Heron's formula states that the area \(A\) of a triangle with sides \(a\), \(b\), and \(c\) is given by:

\[ A = \sqrt{s(s-a)(s-b)(s-c)} \]

where \(s\) is the semi-perimeter of the triangle, calculated as:

\[ s = \frac{a + b + c}{2} \]

For the given triangle with sides \(a = 5\), \(b = 5\), and \(c = 6\):

\begin{enumerate}
\def\labelenumi{\arabic{enumi}.}
\item
  Calculate the semi-perimeter \(s\):
  \[
  s = \frac{5 + 5 + 6}{2} = \frac{16}{2} = 8
  \]
\item
  Substitute \(s\), \(a\), \(b\), and \(c\) into Heron's formula:
  \[
  A = \sqrt{8(8-5)(8-5)(8-6)}
  \]
\item
  Simplify inside the square root:
  \[
  A = \sqrt{8 \times 3 \times 3 \times 2}
  \]
\item
  Calculate the product:
  \[
  A = \sqrt{8 \times 3 \times 3 \times 2} = \sqrt{144}
  \]
\item
  Find the square root:
  \[
  A = 12
  \]
\end{enumerate}

Thus, the area of the triangle is \(\boxed{12}\) square units.
\end{reppobox}
\begin{reppobox}{RePPO \textperiodcentered\ Member 2}{correct, 401 tokens}{code~$-0.13$\quad verification~$-0.66$\quad symbols~$-0.54$\quad markdown~$-0.24$}
\textbackslash boxed\{12\}

\thinktag{<think>}\begingroup\color{black!62}
We are asked: ``Solve the following math problem. Make sure to put the answer (and only answer) inside \textbackslash boxed\{\}.

What is the number of square units in the area of a triangle whose sides measure 5, 5 and 6 units?''

We need step-by-step reasoning, final answer inside \textbackslash boxed\{\}.

We have a triangle with sides 5, 5, 6. This is an isosceles triangle with two equal sides 5 and base 6. We can use Heron's formula. Let a = 5, b = 5, c = 6. Then semiperimeter s = (5+5+6)/2 = 16/2 = 8. Then area = sqrt(s(s-a)(s-b)(s-c)) = sqrt(8\emph{(8-5)}(8-5)\emph{(8-6)) = sqrt(8}3\emph{3}2) = sqrt(144) = 12. So answer: \textbackslash boxed\{12\}.
\endgroup\thinktag{</think>}

Let the sides of the triangle be a = 5, b = 5, and c = 6. First, compute the semiperimeter:

\quad{}\quad{}s = (5 + 5 + 6) / 2 = 16 / 2 = 8

Now, apply Heron's formula for the area:

\quad{}\quad{}Area = \ensuremath{\surd}{[}s(s - a)(s - b)(s - c){]}
\quad{}\quad{}\quad{}\quad{}= \ensuremath{\surd}{[}8(8 - 5)(8 - 5)(8 - 6){]}
\quad{}\quad{}\quad{}\quad{}= \ensuremath{\surd}{[}8 \ensuremath{\times} 3 \ensuremath{\times} 3 \ensuremath{\times} 2{]}
\quad{}\quad{}\quad{}\quad{}= \ensuremath{\surd}{[}144{]}
\quad{}\quad{}\quad{}\quad{}= 12

Thus, the area of the triangle is \textbackslash boxed\{12\}.
\end{reppobox}
\begin{reppobox}{RePPO \textperiodcentered\ Member 3}{correct, 358 tokens}{code~$-0.13$\quad verification~$-0.66$\quad symbols~$-0.54$\quad markdown~$-0.24$}
\textbackslash boxed\{12\}

\thinktag{<think>}\begingroup\color{black!62}
We are given triangle with sides 5, 5, 6. We want area. Let's solve: Use Heron's formula. Let sides: a=5, b=5, c=6. semiperimeter s = (5+5+6)/2 = 16/2 = 8. Then area = sqrt(s(s-a)(s-b)(s-c)) = sqrt(8\emph{(8-5)}(8-5)\emph{(8-6)) = sqrt(8}3\emph{3}2) = sqrt(144) = 12. So answer is \textbackslash boxed\{12\}. But careful: The problem says ``put the answer (and only answer) inside \textbackslash boxed\{\}.'' But the instructions say ``put your final answer within \textbackslash boxed\{\{\}\}.'' So answer: \textbackslash boxed\{12\}. But check: Actually the answer is 12.

I'll produce step-by-step explanation.

I'll produce final answer: \textbackslash boxed\{12\}.
\endgroup\thinktag{</think>}

We are given a triangle with sides 5, 5, and 6. First, we compute the semiperimeter (s):

\quad{}\quad{}s = (5 + 5 + 6) / 2 = 16 / 2 = 8

Next, we use Heron's formula for the area (A):

\quad{}\quad{}A = \ensuremath{\surd}{[}s(s - 5)(s - 5)(s - 6){]}
\quad{}\quad{}\quad{} = \ensuremath{\surd}{[}8 \ensuremath{\times} 3 \ensuremath{\times} 3 \ensuremath{\times} 2{]}
\quad{}\quad{}\quad{} = \ensuremath{\surd}{[}144{]}
\quad{}\quad{}\quad{} = 12

Thus, the area of the triangle is \textbackslash boxed\{12\}.
\end{reppobox}
\begin{reppobox}{RePPO \textperiodcentered\ Member 4}{correct, 759 tokens}{code~$-0.13$\quad verification~$+1.00$\quad symbols~$-0.54$\quad markdown~$-0.24$}
\textbackslash boxed\{12\}

\thinktag{<think>}\begingroup\color{black!62}
We are asked: ``What is the number of square units in the area of a triangle whose sides measure 5, 5 and 6 units?'' Let me check: triangle with sides 5,5,6. This is an isosceles triangle. We can use Heron's formula. Let a=5, b=5, c=6. Then semiperimeter s = (5+5+6)/2 = (16)/2 = 8. So area = sqrt(s(s-a)(s-b)(s-c)) = sqrt(8\emph{(8-5)}(8-5)\emph{(8-6)) = sqrt(8}3\emph{3}2) = sqrt(8\emph{18) = sqrt(144) = 12. So answer is 12. Wait check: 8}3=24, times 3=72, times 2=144, sqrt(144)=12. So answer is \textbackslash boxed\{12\}. But the problem said ``put your final answer within \textbackslash boxed\{\{\}\}'', but I must check: The problem said ``put your final answer within \textbackslash boxed\{\{\}\}''. So I'll output \textbackslash boxed\{12\}. But let me check: Actually, check: 8\emph{3}3*2 = 144, sqrt(144)=12. So answer is 12.

But wait, let me check: Actually, triangle with sides 5,5,6 is valid. So answer is 12.

I must include step-by-step explanation. But instructions: ``Please reason step by step, and put your final answer within \textbackslash boxed\{\{\}\}.'' So I'll output final answer as \textbackslash boxed\{12\}. But instructions said ``and put your final answer within \textbackslash boxed\{\{\}\}.'' So I'll output: \textbackslash boxed\{12\}.

I'll check if any formatting instructions: ``Make sure to put the answer (and only answer) inside \textbackslash boxed\{\}''. So I'll output: \textbackslash boxed\{12\}. But instructions said ``and put your final answer within \textbackslash boxed\{\{\}\}.'' I'll check: The problem said ``What is the number of square units in the area of a triangle whose sides measure 5, 5 and 6 units?'' So answer is 12.

I'll check if any alternative method: using formula for isosceles triangle area: altitude = sqrt(5\^{}2 - (6/2)\^{}2) = sqrt(25 - 9) = sqrt(16)=4, area = 1/2\emph{6}4 = 12. So answer is 12.

I'll produce final answer: \textbackslash boxed\{12\}.
\endgroup\thinktag{</think>}

We start by noting that the triangle has sides 5, 5, and 6. Since two sides are equal, it is an isosceles triangle.

Step 1: Compute the semiperimeter (s):
\quad{}\quad{}s = (5 + 5 + 6) / 2 = 16 / 2 = 8.

Step 2: Use Heron's formula for the area (A):
\quad{}\quad{}A = \ensuremath{\surd}{[}s(s - a)(s - b)(s - c){]}
Here, a = 5, b = 5, and c = 6. So,
\quad{}\quad{}A = \ensuremath{\surd}{[}8(8 - 5)(8 - 5)(8 - 6){]}
\quad{}\quad{}\quad{} = \ensuremath{\surd}{[}8 \ensuremath{\times} 3 \ensuremath{\times} 3 \ensuremath{\times} 2{]}
\quad{}\quad{}\quad{} = \ensuremath{\surd}{[}144{]}
\quad{}\quad{}\quad{} = 12.

Thus, the area of the triangle is \textbackslash boxed\{12\}.
\end{reppobox}
\begin{indbox}{Ind-PPO \textperiodcentered\ Member 0}{correct, 551 tokens}{code~$-0.13$\quad verification~$-0.66$\quad symbols~$-0.54$\quad markdown~$-0.24$}
\textbackslash boxed\{12\}

\thinktag{<think>}\begingroup\color{black!62}
We are asked: ``What is the number of square units in the area of a triangle whose sides measure 5, 5 and 6 units?'' We need to answer with final answer in \textbackslash boxed\{\{\}\}. The answer is: Use Heron's formula. Let sides be 5, 5, 6. Then semiperimeter s = (5+5+6)/2 = 16/2 = 8. Then area = sqrt(s\emph{(s-5)}(s-5)\emph{(s-6)) = sqrt(8}3\emph{3}2) = sqrt(144) = 12. So answer is 12. But instructions: ``Make sure to put the answer (and only answer) inside \textbackslash boxed\{\}.'' So final answer: \textbackslash boxed\{12\}. But instructions: ``Please reason step by step, and put your final answer within \textbackslash boxed\{\{\}\}.'' So final answer: \textbackslash boxed\{12\}. But careful: The instructions require final answer inside \textbackslash boxed\{\{\}\}. Possibly they want \textbackslash boxed\{12\}. But instructions: ``and put your final answer within \textbackslash boxed\{\{\}\}.'' So I'll do: \textbackslash boxed\{12\}. But instructions: ``Make sure to put the answer (and only answer) inside \textbackslash boxed\{\}.'' So I'll output: \textbackslash boxed\{12\}. But instructions: ``Please reason step by step, and put your final answer within \textbackslash boxed\{\{\}\}.'' So I'll include step-by-step explanation and then final answer inside \textbackslash boxed\{12\}. But instructions: ``and put your final answer within \textbackslash boxed\{\{\}\}.'' So I'll output: \textbackslash boxed\{12\}. I'll output answer as: \textbackslash boxed\{12\}. I'll check: ``What is the number of square units in the area of a triangle whose sides measure 5, 5 and 6 units?'' Yes.

I'll produce final answer: \textbackslash boxed\{12\}. We'll produce step-by-step explanation.
\endgroup\thinktag{</think>}

To solve the problem, we use Heron's formula. First, we find the semiperimeter (s):

s = (5 + 5 + 6) / 2 = 16 / 2 = 8.

Next, we calculate the area (A):

A = \ensuremath{\surd}{[}s(s - 5)(s - 5)(s - 6){]}
= \ensuremath{\surd}{[}8 \ensuremath{\times} (8 - 5) \ensuremath{\times} (8 - 5) \ensuremath{\times} (8 - 6){]}
= \ensuremath{\surd}{[}8 \ensuremath{\times} 3 \ensuremath{\times} 3 \ensuremath{\times} 2{]}
= \ensuremath{\surd}{[}144{]}
= 12.

Thus, the area of the triangle is \textbackslash boxed\{12\}.
\end{indbox}
\begin{indbox}{Ind-PPO \textperiodcentered\ Member 1}{correct, 477 tokens}{code~$-0.13$\quad verification~$-0.66$\quad symbols~$-0.54$\quad markdown~$-0.24$}
\textbackslash boxed\{12\}

\thinktag{<think>}\begingroup\color{black!62}
We are asked: ``What is the number of square units in the area of a triangle whose sides measure 5, 5 and 6 units?'' We need to produce answer inside \textbackslash boxed\{\ldots\}. We need to produce step-by-step reasoning and final answer inside \textbackslash boxed\{\}.

We have triangle with sides 5, 5, 6. This is an isosceles triangle with base 6 and equal sides 5. We can compute the area using Heron's formula. Let s = (5+5+6)/2 = 16/2 = 8. So area = sqrt(s(s - a)(s - b)(s - c)) = sqrt(8\emph{(8-5)}(8-5)\emph{(8-6)) = sqrt(8}3\emph{3}2) = sqrt(144) = 12. So answer is 12.

Thus answer: \textbackslash boxed\{12\}.

I'll produce answer: \textbackslash boxed\{12\}. We'll produce reasoning. We'll produce answer inside \textbackslash boxed\{12\}. We'll produce answer: \textbackslash boxed\{12\}. We'll produce answer with step-by-step reasoning. We'll produce answer with final answer inside \textbackslash boxed\{12\}. We'll produce answer with step-by-step reasoning.
\endgroup\thinktag{</think>}

To find the area of a triangle with sides 5, 5, and 6, we can use Heron's formula.

Step 1: Compute the semiperimeter (s):
\quad{}\quad{}s = (5 + 5 + 6) / 2 = 16 / 2 = 8

Step 2: Apply Heron's formula:
\quad{}\quad{}Area = \ensuremath{\surd}{[}s(s - a)(s - b)(s - c){]}
\quad{}\quad{}Here, a = 5, b = 5, c = 6.
\quad{}\quad{}Area = \ensuremath{\surd}{[}8(8 - 5)(8 - 5)(8 - 6){]}
\quad{}\quad{}\quad{}\quad{}= \ensuremath{\surd}{[}8 \ensuremath{\times} 3 \ensuremath{\times} 3 \ensuremath{\times} 2{]}
\quad{}\quad{}\quad{}\quad{}= \ensuremath{\surd}{[}144{]}
\quad{}\quad{}\quad{}\quad{}= 12

Thus, the area of the triangle is \textbackslash boxed\{12\}.
\end{indbox}

\FloatBarrier
\section{Pixel-based locomotion}
\label{app:pixels}

To test RePPO with large continuous-control policies, we train populations of $16$ convolutional policies from pixels in HalfCheetah, Humanoid, and Walker2D.
Observations are side views of the agent rendered on the GPU with MuJoCo Warp, converted to $84\times84$ grayscale images, and stacked over the last $3$ frames.
Each policy is a Nature CNN encoder with separate two-layer heads of $512$ units for the actor and both critics, totaling $2.5$~M parameters.
The tasks, descriptors, and evaluation are those of Appendix~\ref{app:cc}.
We train for $262$~M steps in total, or $16.4$~M per policy, with $16$ environments per policy, rollouts of $256$ steps, $4$ epochs over $16$ minibatches, a learning rate of $3\times10^{-4}$ that decays as in Table~\ref{app:tab:cc-hparams} with $S$ equal to the budget, an entropy coefficient of $0.005$, value clipping, $\ell^2=0.1$, and $K=4$ neighbors; the remaining settings are shared with Table~\ref{app:tab:cc-hparams}.
We compare RePPO with Ind-PPO over three seeds, since archive-based baselines would need to store and train thousands of such policies.

\begin{figure}[ht]
    \centering
    \includegraphics[width=0.8\textwidth]{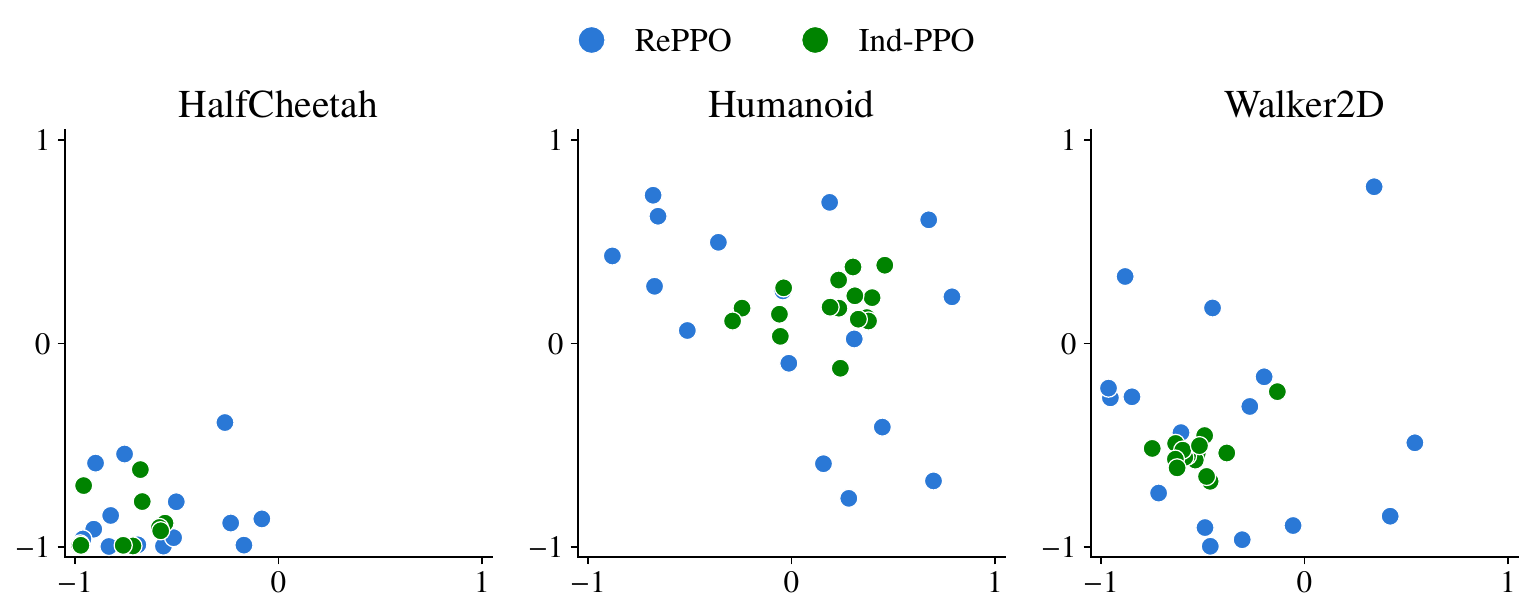}
    \caption{Behavior descriptors of pixel-based policies for one seed per task.}
    \label{app:fig:pixels}
\end{figure}

RePPO's populations are substantially more diverse than Ind-PPO's in all three tasks, with Vendi Scores of $1.9$--$4.2$ against $1.2$--$1.5$, and achieve the higher qVS in every task (Table~\ref{app:tab:pixels}).
On HalfCheetah and Walker2D, RePPO also attains the higher mean fitness, while on Humanoid it trades part of it for diversity.
Figure~\ref{app:fig:pixels} shows that RePPO's policies spread across contact patterns, whereas Ind-PPO's policies concentrate in a small region.

\begin{table}[ht]
\centering
\caption{Pixel-based locomotion, as mean $\pm$ standard error over three seeds.}
\label{app:tab:pixels}
\small
\begin{tabular}{@{}llcccc@{}}
\toprule
Task & Method & Best fitness & Mean fitness & Vendi Score & qVS \\
\midrule
\multirow{2}{*}{HalfCheetah}
& RePPO & $4648\pm41$ & $\mathbf{3257\pm31}$ & $\mathbf{1.90\pm0.03}$ & $\mathbf{6187\pm54}$ \\
& Ind-PPO & $\mathbf{5085\pm366}$ & $2887\pm122$ & $1.42\pm0.04$ & $4106\pm253$ \\
\addlinespace
\multirow{2}{*}{Humanoid}
& RePPO & $2828\pm830$ & $1078\pm63$ & $\mathbf{4.22\pm0.10}$ & $\mathbf{4552\pm268}$ \\
& Ind-PPO & $\mathbf{3840\pm224}$ & $\mathbf{1709\pm85}$ & $1.47\pm0.05$ & $2518\pm156$ \\
\addlinespace
\multirow{2}{*}{Walker2D}
& RePPO & $\mathbf{3940\pm103}$ & $\mathbf{2604\pm72}$ & $\mathbf{3.94\pm0.07}$ & $\mathbf{10252\pm187}$ \\
& Ind-PPO & $3357\pm311$ & $2061\pm161$ & $1.20\pm0.02$ & $2461\pm140$ \\
\bottomrule
\end{tabular}
\end{table}

\FloatBarrier
\section{LunarLander}
\label{app:lunarlander}

Our other domains share dense behavior signals that are observed at every step.
LunarLander~\citep{gym} instead tests RePPO with discrete actions and a sparse, event-based descriptor.
The descriptor records the lander's horizontal position and vertical speed at its first contact with the ground, so its signal is nonzero at a single step of each episode.
Positions span the screen width, and speeds are mapped from $0$ to the speed of free fall, about $2$ in the environment's units, onto $[-1,1]$; episodes without contact receive the descriptor $(0,0)$.
Since the reward favors fuel-efficient, gentle landings near the landing pad, diversity in this domain necessarily costs quality.

RePPO and Ind-PPO train $32$ policies with $64$-unit MLPs for $65.5$~M steps in total, with $8$ environments per policy and rollouts of $256$ steps.
The remaining PPO settings follow those recommended for this task by Stable-Baselines3~\citep{sb3}: $4$ epochs over minibatches of $64$ samples, a learning rate of $10^{-3}$, $\gamma=0.999$, $\lambda=0.98$, and an entropy coefficient of $0.01$; RePPO uses $C_{\max}=0.4$.
We compare against Sep-CMA-MAE, which is well suited to the small policies of this task, with a $32\times32$ grid archive, $10$ episodes per evaluation, and the same budget of environment steps.
All methods use five seeds.

\begin{figure}[ht]
    \centering
    \includegraphics[width=0.5\textwidth]{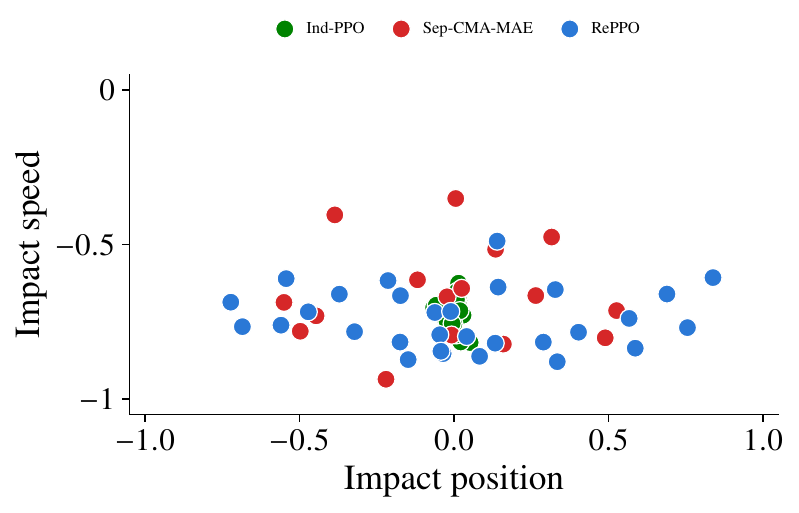}
    \caption{Behavior descriptors of LunarLander policies for one seed per method. For readability, Sep-CMA-MAE's archive is thinned to one policy in each of $32$ CVT cells, keeping the fittest.}
    \label{app:fig:lunarlander}
\end{figure}

All three methods find policies that solve the task, with best fitness between $270$ and $288$ (Table~\ref{app:tab:lunarlander}).
Ind-PPO's population collapses to a single landing strategy, whereas RePPO and Sep-CMA-MAE trade mean fitness for diversity.
RePPO doubles Ind-PPO's Vendi Score and matches that of Sep-CMA-MAE with a higher mean fitness, which gives it the highest qVS.
Figure~\ref{app:fig:lunarlander} shows that RePPO's policies land across the width of the screen, varying mostly in position rather than speed, whereas Ind-PPO's policies land at nearly the same spot.
RePPO's diversity mechanism thus remains effective when the behavior signal is observed at a single event.

\begin{table}[ht]
\centering
\caption{LunarLander, as mean $\pm$ standard error over five seeds.}
\label{app:tab:lunarlander}
\small
\begin{tabular}{@{}lcccc@{}}
\toprule
Method & Best fitness & Mean fitness & Vendi Score & qVS \\
\midrule
RePPO & $278\pm2$ & $185\pm2$ & $\mathbf{2.06\pm0.03}$ & $\mathbf{381\pm3}$ \\
Ind-PPO & $\mathbf{288\pm1}$ & $\mathbf{272\pm1}$ & $1.05\pm0.01$ & $286\pm1$ \\
Sep-CMA-MAE & $270\pm2$ & $135\pm2$ & $2.05\pm0.02$ & $276\pm3$ \\
\bottomrule
\end{tabular}
\end{table}

\subsection{Computational Resources}
\label{app:compute}
The continuous-control, Atari, pixel-based, and LunarLander experiments were distributed over the following workstations:
\begin{itemize}
    \item AMD Ryzen Threadripper 3990X, 64 GB RAM, 2 NVIDIA RTX 3090 GPUs;
    \item AMD Ryzen Threadripper 3990X, 64 GB RAM, NVIDIA RTX 4090;
    \item Intel Core i7-12700F, 16 GB RAM, NVIDIA RTX 3060;
    \item a shared server with NVIDIA RTX A6000 GPUs.
\end{itemize}
A RePPO run on Atari requires about 4 GB of GPU memory.
Each LLM population was trained on five NVIDIA RTX A6000 GPUs, one member per GPU, each running both generation with vLLM~\citep{kwon2023efficient} and training.
Training a population took 21 h for Qwen3 and 64 h for Granite, or about 100 and 320 GPU hours, and Ind-PPO required similar resources.
Table~\ref{app:tab:wallclock} reports the approximated training times of locomotion experiments, in hours.

\begin{table}[ht]
\centering
\caption{Continuous control tasks' training wall-clock time in hours.
$^{\dagger}$ indicates experiments run on an RTX 3090}
\label{app:tab:wallclock}
\small
\begin{tabular}{@{}lccccc@{}}
\toprule
Method & GPU & Ant & HalfCheetah & Humanoid & Walker2D \\
\midrule
RePPO & 4090 & 5.1 & 8 & 5.3 & 5.7 \\
Ind-PPO & 4090 & 4.1 & 8 & 5.3 & 5.7 \\
PPGA & A6000 & 2.7 & 9.1 & 3.5 & 7.6 \\
PGA-ME & 4090 & 9.8 & 26.7 & 11.6 & 22.3 \\
Sep-CMA-MAE & 4090 & 0.9--3.5 & 34 & 4.1$^{\dagger}$ & 41$^{\dagger}$ \\
\bottomrule
\end{tabular}
\end{table}

\end{document}